\documentclass[journal]{IEEEtran}
\usepackage{definitions}
\usepackage{amsfonts}
\usepackage[ruled,vlined,linesnumbered]{algorithm2e}
\SetNlSkip{0.4em}
\usepackage{array}
\usepackage{multirow}
\usepackage[caption=false,font=normalsize,labelfont=sf,textfont=sf]{subfig}
\usepackage{textcomp}
\usepackage{stfloats}
\usepackage{url}
\usepackage{verbatim}

\usepackage{graphicx}

\usepackage{cite}
\usepackage{hyperref}
\usepackage{xspace}
\usepackage{cleveref}
\usepackage{xcolor} % colored text
\usepackage{booktabs}
\usepackage{pifont}
\usepackage[cal=pxtx,scr=boondox]{mathalpha}
\usepackage{dsfont}
\usepackage{mathtools}
\usepackage{adjustbox}
\usepackage{enumitem}
\usepackage{siunitx}
\usepackage{tikz}

\usepackage{colortbl, xcolor}

\newcommand{\spx}{{\mathscr{x}}}
\newcommand{\sps}{{\mathscr{s}}}
\newcommand{\spr}{{\mathscr{r}}}

\newcommand{\ula}{{\bm{\lambda}}}
\newcommand{\bmp}{{\bm{\Phi}}}
\newcommand{\rg}[1]{{\mathring{#1}}}

\newcommand{\cpX}{{\mathcal{{X}}}}

\newcommand{\augx}{{\breve\ux}}
\newcommand{\augP}{{\breve\fP}}
\newcommand{\framelabel}[3][south west]{%
	\begin{tikzpicture}[baseline,inner sep=0pt,outer sep=0pt]
		\node[anchor=south west] (img) {#2};
		\node[anchor=#1,inner xsep=0pt,inner ysep=2pt,fill=white,fill opacity=0.7,text=black,text opacity=1] at (img.#1) {\footnotesize\bfseries #3};
	\end{tikzpicture}%
}

\newcommand{\fimu}{{\ttt{I}}}
\newcommand{\flid}{{\ttt{L}}}
\newcommand{\fwor}{{\ttt{W}}}
\newcommand{\fcam}{{\ttt{V}}}
\newcommand{\frob}{{\ttt{B}}}

\newcommand{\acc}{{\ttt{acc}}}
\newcommand{\gyro}{{\ttt{gyro}}}
\newcommand{\twi}{{\text{w}}}

\begin{document}

\title{RESPLE2: Inertial-Informed Recursive Non-Uniform Spline Estimation for Multi-Sensor Odometry with Global Refinement}

\author{Ziyu Cao, William Talbot, Song Li, Pakpong Chirarattananon, Fredrik Gustafsson, Marco Hutter, and Kailai Li%
\thanks{This work was supported in part by the Swiss National Science Foundation (SNSF) [20CH-1\_229464/1] under Grant CHIST-ERA-23-MultiGIS-07 (to W.T.), in part by the Natural Sciences and Engineering Research Council of Canada (NSERC) under Discovery Grant RGPIN-2026-05891 (to P.C.), and in part by the Canada Foundation for Innovation (CFI) John R. Evans Leaders Fund (JELF) under Award 47054 (to P.C.). \textit{(Corresponding author: Kailai Li.)}}%
\thanks{Ziyu Cao, Fredrik Gustafsson and Kailai Li are with the Department of Electrical Engineering, Linköping University, 58183 Linköping, Sweden (e-mail: ziyu.cao@liu.se; fredrik.gustafsson@liu.se; kailai.li@liu.se).}%
\thanks{William Talbot and Marco Hutter are with the Robotic Systems Lab (RSL), ETH Zürich, 8092 Zurich, Switzerland (e-mail: wtalbot@ethz.ch; mahutter@ethz.ch).}%
\thanks{Song Li and Pakpong Chirarattananon are with the Department of Mechanical and Industrial Engineering, University of Toronto, Toronto, ON M5S 3G8, Canada. (e-mail: lsong.li@mail.utoronto.ca; p.chirarattananon@utoronto.ca)}%
\thanks{Song Li is also with the Department of Biomedical Engineering, City University of Hong Kong, Hong Kong (e-mail: lsong22-c@my.cityu.edu.hk).}%
}

\maketitle

\begin{abstract} 
    We present RESPLE2, a recursive multi-sensor state estimation framework using non-uniform B-splines for 6-DoF motion representation. By exploiting the knot-clamping boundary property, we propose a theoretically grounded mechanism for exact control-point prediction directly informed by inertial kinematics along the trajectory, with knot placement adapted to motion and measurement availability. Within an iterated extended Kalman filter, we perform this prediction using a subset of inertial measurements. A subsequent reset step preserves trajectory consistency as the knots extend. The update step further incorporates heterogeneous sensor measurements at their native timestamps through continuous-time observation models, supporting flexible combinations of LiDARs, cameras, inertial measurement units, and leg kinematics. We further develop a lightweight global optimization module that directly refines non-uniform B-spline control points by incorporating loop-closure constraints derived from LiDAR-visual cues. We conduct extensive evaluations on public and experimental datasets across handheld, tracked, legged, and hybrid aerial-legged platforms, covering highly dynamic maneuvers with extreme accelerations. Compared with state-of-the-art multi-sensor odometry systems, RESPLE2 achieves competitive or superior estimation accuracy and robustness in the corresponding sensor configurations, particularly under aggressive motion, while maintaining real-time operation. We publicly release our code and experimental datasets at \ttt{\url{https://github.com/ASIG-X/RESPLE2}}. 
\end{abstract}

\section{Introduction}
Recent advances in high-performance sensing and actuation technologies have enabled mobile robots to operate in increasingly diverse and challenging environments. Emerging platforms, including legged and hybrid locomotion systems equipped with multimodal sensors~\cite{miki2022learning,li2025high}, have been developed for demanding applications spanning search and rescue, logistics, healthcare, entertainment, household assistance, in situ forestry, and environmental monitoring~\cite{matias2025building,lee2024learning,richter2026large,bai2026parallel,tranzatto2022cerberus}. 

\begin{figure*}[t!]
    \centering
    \includegraphics[width=0.98\linewidth]{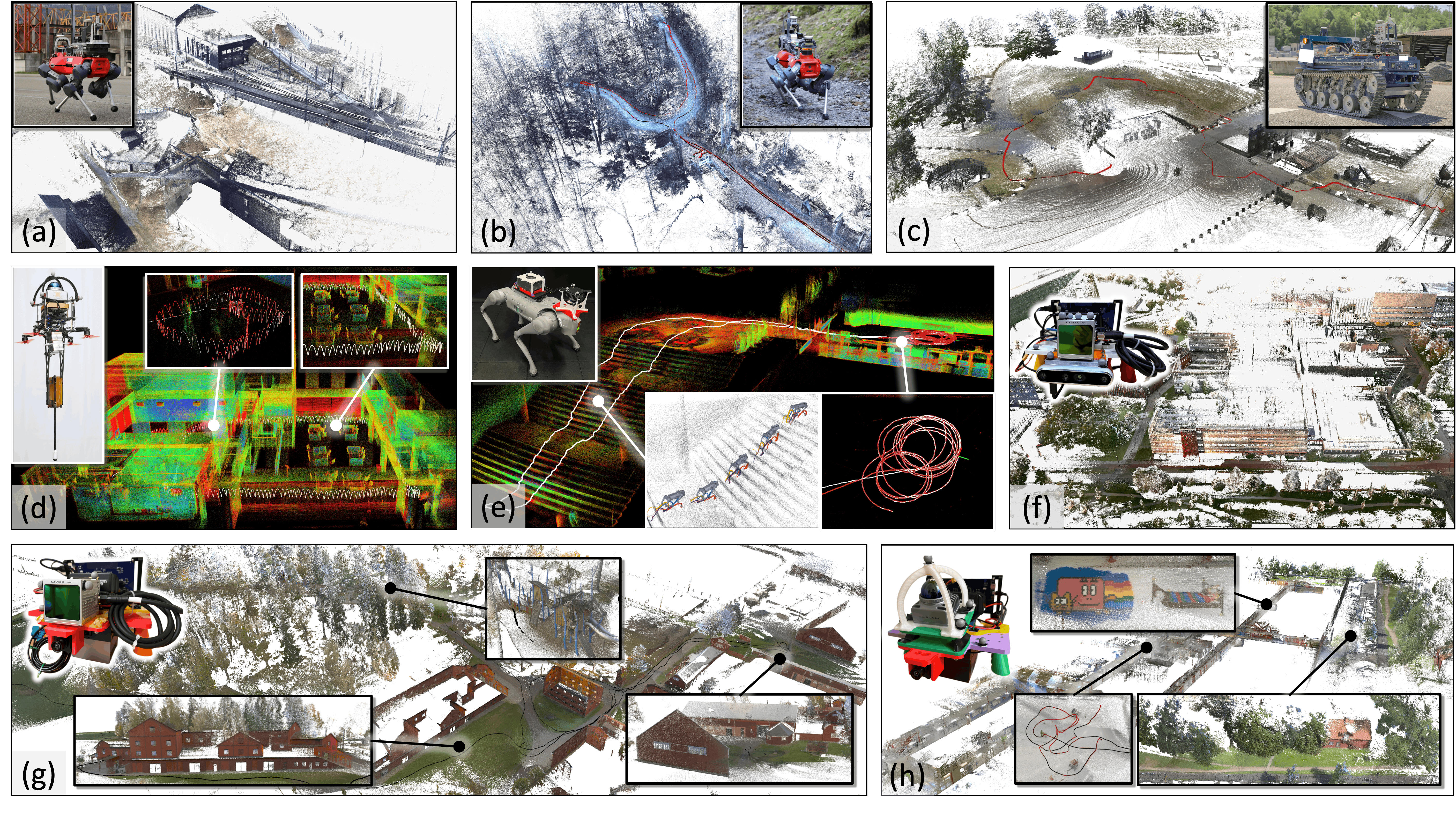}
    \vspace{-4mm}

    \caption{RESPLE2 enables multi-sensor odometry across diverse scenarios on ANYmal D (a, b) and Unitree Go2 (e) legged robots, a tracked vehicle (c), a hybrid legged quadcopter (d), and handheld platforms (f, g, h). The \textit{Boxi} sensor payload~\cite{frey2025boxi} was deployed for the ANYmal D and tracked vehicle.}
    \label{fig:front}
	\vspace{-4mm}
\end{figure*}

Accurate and robust six-degree-of-freedom (6-DoF) egomotion estimation is essential for high-performance autonomy in these emerging scenarios. Odometry systems using cameras and inertial measurement units (IMUs) are well established~\cite{vins2018}. Light detection and ranging (LiDAR) sensors complement cameras by providing accurate, illumination-invariant geometric observations over long ranges and across a wide field of view (FoV), motivating configurations such as LiDAR-inertial odometry (LIO), LiDAR-visual-inertial odometry (LVIO), and LIO augmented with leg kinematics (LKIO)~\cite{xu2022fast,li2021towards,bai2022faster,ou2024leg}. Fusing these complementary measurements mitigates modality-specific degradation, improves estimation accuracy and robustness, and enables more informative scene maps. Nevertheless, aggressive platform dynamics, unstructured environments, and limited onboard computational resources continue to challenge reliable egomotion estimation~\cite{frey2025boxi,nubert2026holistic}. Beyond these operational challenges, most existing odometry systems remain sensor-specific and standalone, with limited extensibility to new modalities. Furthermore, they have been validated primarily on conventional aerial and wheeled platforms rather than on emerging embodiments deployed in demanding real-world mobile scenarios, such as highly dynamic (loco)motion and complex terrain interaction~\cite{li2025high,yuan2025reasan}.

Conventional multi-sensor estimation methods represent dynamic motion in discrete time and use recursive filtering or nonlinear sliding-window optimization, with motion states associated with specific timestamps at fixed frame rates. During online estimation, asynchronous sensor measurements are typically approximated, e.g., through temporal interpolation~\cite{vins2018,niu2025esvo2}. This paradigm often scales poorly as sensor rates and the number of sensors vary, particularly when changing sensors without synchronized triggering. Moreover, measurements from many exteroceptive sensors, such as LiDARs and rolling-shutter cameras, exhibit motion distortion during agile maneuvers and require explicit modeling or compensation for undistortion in state estimation. These design choices can compromise the spatiotemporal fidelity of raw observations and alter underlying noise characteristics, potentially degrading estimation accuracy and robustness while increasing both methodological and engineering complexity~\cite{lang2022ctrl,xu2022fast}.

Continuous-time motion representations decouple the number and temporal placement of estimates from observations, facilitating the integration of asynchronous multi-sensor readings at their native timestamps without complicating system design. Piecewise-linear models show promise for LiDAR-only odometry, e.g., Traj-LO~\cite{zheng2024trajlo}, due to simplicity and computational efficiency, but may inadequately capture nonlinear motion during rapid or aggressive maneuvers. A more advanced, well-studied approach uses \textit{exactly sparse} temporal Gaussian processes (GPs), in which system dynamics are modeled as a Markovian LTV-SDE process, such as white-noise-on-acceleration (WNOA) or white-noise-on-jerk (WNOJ)~\cite{tang2019white,nguyen2026third,shen2025cte,johnson2024continuous,barfoot2024state}. B-splines, often cubic, represent $6$-DoF motion as piecewise polynomial functions, with each segment defined by a local set of \textit{control points} blended according to a sequence of chosen scalar times called \textit{knots} of typically uniform intervals~\cite{nguyen2024eigen,cao2025resple,lang2023coco}. 

Most existing online continuous-time estimators solve maximum likelihood estimation (MLE) or maximum a posteriori (MAP) estimation problems in a sliding-window fashion~\cite{hug2022con,nguyen2024eigen,lang2023coco,lang2022ctrl}. Although these optimization-based methods generally achieve high accuracy and improve robustness to sensor degeneracy, their real-time deployment is challenging and often requires carefully designed solvers, which complicates the integration of additional sensor modalities.

Recursive Bayesian estimators, such as the extended Kalman filter (EKF), offer a methodologically lightweight alternative; however, they remain scarce and underexplored in continuous-time multi-sensor odometry systems and are restricted to a narrow, fixed set of sensor configurations. For example, CTE-MLO performs recursive estimation using a GP-based continuous-time representation for LiDAR-only odometry~\cite{shen2025cte}. RESPLE is the first $6$-DoF B-spline-based recursive motion estimator, supporting both LiDAR-only and LiDAR-inertial odometry~\cite{cao2025resple}. But its uniform knot spacing imposes a globally fixed temporal knot interval, where accurate control-point estimation requires knot intervals to be short enough for the polynomial to capture aggressive motions, yet long enough to include sufficient sensor measurements to constrain all degrees of freedom at each update, increasing susceptibility to geometric degeneracy. It is therefore theorized that well-chosen, non-uniform knots can bring both accuracy and computational benefits~\cite{talbot2024continuous,lang2023coco}. Despite preliminary efforts to develop non-uniform B-spline LVIO using sliding-window optimization~\cite{lang2023coco}, the broader methodological and practical potential of this motion-adaptive representation remains underexplored.

More fundamentally, despite its growing adoption in practical systems, the theoretical formulation of B-spline-based odometry systems remains incomplete. Regardless of whether estimation uses filtering or sliding-window optimization, or whether knot spacing is uniform or non-uniform, B-spline control points do not lie on the motion trajectory in general. Consequently, existing theory has not established exact kinematic models for the states (i.e., the control points) analogous to those used in discrete-time or GP-based continuous-time approaches. As a result, new control points have not been predicted or initialized along the trajectory by directly integrating kinematic information, such as IMU measurements. Furthermore, no existing non-uniform B-spline-based odometry system natively supports loop closure for global refinement. 

To address these methodological and engineering limitations, we present RESPLE2, a novel recursive Bayesian estimation framework that represents continuous-time 6-DoF motion using non-uniform B-splines within an iterated extended Kalman filter (IEKF). Unlike existing B-spline-based estimators, our framework introduces a inertial-informed process model to guarantee exact prediction of control points located on the motion trajectory. Moreover, RESPLE2 enables a versatile estimation framework supporting a configurable fusion of LiDAR, camera, IMU, and leg joint measurements. Extensive real-world validation spans generic mobile platforms and, particularly, emerging robotic embodiments. More specifically, our key contributions are summarized below.

\subsection*{Contributions}
\begin{itemize}[leftmargin=*]
    \item We propose the first recursive continuous-time estimator using non-uniform B-splines for multi-sensor odometry. Using the knot-clamping boundary property, the framework enables exact prediction of control points on the motion trajectory directly informed by inertial measurements and supports efficient and versatile multi-sensor fusion.
    \item  A lightweight optimization module is additionally developed for direct global refinement of non-uniform control points for correcting drifts at loop closures detected by LiDAR-visual cues, while preserving the local kinematic consistency of the continuous-time trajectory.
    \item The framework enables diverse multi-sensor odometry systems, including LO, LIO, LVIO, LKIO, and LKVIO, within a unified design. We conduct extensive evaluations on legged, tracked vehicle, and hybrid legged-quadcopter platforms under challenging real-world conditions, including aggressive motion and exteroceptive degradation. Compared with state-of-the-art methods, RESPLE2 achieves comparable or superior odometry robustness and accuracy in real time and supports optional global refinement.
    \item We publicly release the full system implementation and the experimental datasets.
\end{itemize}

\section{Related Work}\label{sec:related}
\subsection{Multi-Sensor Odometry}
Popular multi-sensor odometry systems remain predominantly based on discrete-time formulations, in which tightly coupled filtering-based or optimization-based estimators can fuse disparate sensor modalities more consistently and accurately than their loosely coupled counterparts~\cite{zhao2021super,nubert2026holistic}. The most closely related system, FAST-LIVO2~\cite{zheng2024fast}, uses IMU strapdown integration within an error-state iterated Kalman filter (ESIKF) and \textit{VoxelMap} for local map management~\cite{yuan2022efficient}. The sequential updates adopt LiDAR point-to-plane residuals using planes extracted from the local map and photometric visual residuals computed by warping image patches stored in the map and efficiently retrieved through on-demand voxel raycasting. Point-LIO focuses on robust LIO under dynamic motion and IMU saturation, with point-wise updates for both inertial and LiDAR measurements~\cite{he2023point}. For legged robots, kinematic information from joint encoders and contact sensing provides an additional measurement modality to complement inertial and exteroceptive sensors. Leg-KILO~\cite{ou2024leg} fuses LiDAR, inertial, and leg joint measurements using an error-state Kalman filter (ESKF) for quadrupedal odometry, with point-wise updates for individual sensors as in Point-LIO~\cite{he2023point}.

The proposed RESPLE2 adopts \textit{VoxelMap}~\cite{yuan2022efficient} for local map management and uses feature-based visual residuals. These design choices build on the state of the art and place less emphasis on variations in map representation while prioritizing estimation robustness and versatile multi-sensor designs, enabling LVIO, LKIO, and beyond.

\subsection{Continuous-Time Motion Estimation}
We refer readers to~\cite{talbot2024continuous,johnson2024continuous} for comprehensive surveys and comparisons of continuous-time motion estimation methods and their relative advantages in robotics. GP-based methods provide the most theoretically complete alternative, enabling constant-time interpolation of the mean and covariance between state variables and supporting online MAP estimation through sliding-window optimization. Following this strategy, systems fusing inertial measurements with individual modalities, including camera, LiDAR, radar, ultra-wideband, and GNSS, have been reported~\cite{johnson2024continuous,burnett2024continuous,nguyen2026third,talbot2024continuous}. CTE-MLO~\cite{shen2025cte} further combines the GP representation with an EKF for LiDAR-only odometry. Its filtering-based design makes it a GP-based counterpart to RESPLE2. However, to the best of our knowledge, no LVIO or LKVIO systems using the GP representation have been proposed.

Recent improvements in interpolation efficiency have promoted the use of B-splines in multi-sensor odometry~\cite{sommer2020efficient,li2023continuous}. As with other continuous-time methods, almost all B-spline-based systems perform MAP estimation by optimizing control points over one or more spline segments~\cite{talbot2024continuous}. For example, CLINS, SLICT2, and CLIC~\cite{lv2021clins,nguyen2024eigen,lv2023continuous} use sliding-window optimization to estimate uniform B-spline trajectories for LIO or LVIO. RESPLE~\cite{cao2025resple} was the first to introduce recursive B-spline estimation using an IEKF, supporting LO and LIO within a unified design. This lightweight estimator achieves comparable or superior accuracy and greater robustness under dynamic motion than these optimization-based systems. Its LIO variant provides a direct baseline for assessing RESPLE2's exact control-point prediction within a non-uniform B-spline framework.

For online non-uniform B-spline estimation, the knot can be placed by using the maximum magnitudes of IMU-measured angular velocity and acceleration as heuristic indicators of motion complexity, selecting shorter segments when these values are high~\cite{lang2023coco}. As a non-uniform B-spline LVIO system, Coco-LIC~\cite{lang2023coco} uses sliding-window MAP estimation and determines knot locations by subdividing a uniform knot distribution based on IMU excitation. Alternatively, the number and local distribution of visual keyframes can guide knot density selection~\cite{wu2026nurbs}.  A further methodological challenge is that the blending matrix needed for interpolation is no longer constant but depends on future knots to be determined in advance. This is problematic for Coco-LIC~\cite{lang2023coco} as the approach must wait for future IMU measurements to determine knot placements, inducing optimization delays~\cite{quenzel2025lio}. LIO-Mars~\cite{quenzel2025lio} sets future knots according to an expected interval, correcting these knots when future LiDAR scans arrive. 

For both uniform and non-uniform B-splines, no existing online estimation scheme enables \textit{exact} initialization or prediction of new control points by directly integrating kinematic information, such as IMU measurements, along the motion trajectory. For example, RESPLE~\cite{cao2025resple} uses a process model with a constant transition matrix to predict new control points by carrying forward the velocities at previous knots. SLICT2~\cite{nguyen2024eigen} initializes new control points at the trajectory endpoint plus an offset obtained by integrating IMU measurements, causing the resulting B-spline to deviate from the predicted trajectory. Coco-LIC~\cite{lang2023coco} adopts a simple scheme that copies the most recent control points. This theoretical limitation can compromise estimation performance under challenging conditions, such as aggressive dynamics, particularly with exteroceptive degradation. In comparison, RESPLE2 exploits the knot-clamping boundary property of non-uniform B-splines to ensure that the latest control point always lies on the trajectory, thereby providing a theoretical guarantee for exact control-point prediction using kinematic models informed by inertial measurements, with knot placement guided by both kinematic and exteroceptive cues. This enables significantly more accurate control-point prediction without estimation delays. 

Furthermore, native support for loop closure and global refinement within B-spline estimation remains limited. For loop closure detection, BTC~\cite{yuan2024btc} uses a binary and triangle descriptor to encode global and local point-cloud structure for 3D place recognition. Its extension, iBTC~\cite{zou2024ibtc}, incorporates visual information, outperforming geometry-based BTC and bag of visual words in LVIO systems while remaining efficient for online operation. CLINS~\cite{lv2021clins} adopts a two-stage global correction scheme at loop closures for uniform B-splines in LIO: pose-graph optimization of discrete-time key-scan poses using loop-closure constraints, followed by control-point updates that align the continuous-time trajectory while preserving its local shape. RESPLE2 integrates iBTC~\cite{zou2024ibtc} for loop closure detection in the LVIO setting and is the first to provide native global trajectory refinement by directly optimizing non-uniform B-spline control points.

\section{Preliminaries} \label{sec:preliminaries}

\subsection{Notation Conventions}
Throughout this paper, scalar values are written as lowercase letters, such as $a$. Vectors are denoted by bold lowercase letters, such as $\ua$, and matrices by bold uppercase letters, such as $\fA$. The $d$-dimensional zero vector and the $d\times d$ zero and identity matrices are denoted by $\uzero_{d\times 1}$, $\fzero_{d}$, and $\fI_{d}$, respectively. Specifically, continuous-time functions represented by B-splines are denoted by script letters, such as $\sps(t)$. $\diag(\cdot)$ assembles a (block-)diagonal matrix from its arguments. Operators $\bullet$ and $\otimes$ denote the Hamilton and Kronecker products, respectively. Rigid-body orientations are represented by unit quaternions on the manifold as $\ur\in\Sbb^3$, $\llmat{\scQ}(\cdot)$ and $\lrmat{\scQ}(\cdot)$ denote its left and right matrix expressions, and $\scR(\cdot):\Sbb^3\to\SOT$ maps a quaternion $\ur$ to its corresponding rotation matrix $\fR$~\cite{sola2017quaternion}. $\ua_\times$ denotes the skew-symmetric cross-product matrix for vector $\ua\in\R^3$. $\fR_\flid^\fimu$ and $\us_\flid^\fimu$ denote the rotation matrix and translation vector from the LiDAR ($\flid$) to the IMU ($\fimu$) frame, respectively. $(\cdot)^\fwor$ denotes a vector in the world frame, $(\cdot)^\fcam$ in the camera frame, and $(\cdot)^\frob$ in the robot base frame.

\subsection{Non-Uniform B-Spline Motion Representation}\label{subsec:nbs}
We represent the 6-DoF egomotion trajectory in continuous time with cubic B-splines of non-uniformly spaced knots, decoupled into a translational and a rotational component as
\begin{equation}\label{eq:spl}
	\spx(t)=[\sps(t)^\top,\spr(t)^\top]^\top\in\R^3\times\Sbb^3\subset\R^7\,,
\end{equation}
where $\sps(t)$ and $\spr(t)$ are the position and quaternion-valued orientation trajectories, respectively. The trajectory in~\eqref{eq:spl} is defined over a sequence of $n$ knots $\{t_i\}_{i=1}^n$ with non-uniform intervals $\tau_i=t_{i+1}-t_i$, and is determined by translational control points $\{\us_i\}_{i=1}^n$ and rotational control points $\{\ur_i\}_{i=1}^n$.

Given a query time $t\in[\,t_{n-1},t_n]$, the position is obtained from the four local control points according to
\begin{equation}\label{eq:st}
	\sps(t)=[\us_{n-3}\,,\us_{n-2}\,,\us_{n-1}\,,\us_n]\,\bom\,\uu\,,\quad\text{where}\quad
\end{equation}
\begin{equation*}
	\bom=\Bigg[\sbbmat
	m_{0,0} & -3m_{0,0} & 3m_{0,0} & -m_{0,0} \\
	1-m_{0,0}-m_{2,0} & 3m_{0,0}-m_{2,1} & -3m_{0,0}-m_{2,2} & m_{0,0}-m_{2,3}-m_{3,3} \\
	m_{2,0} & m_{2,1} & m_{2,2} & m_{2,3} \\
	0 & 0 & 0 & m_{3,3}
	\sebmat\Bigg]
\end{equation*}	
is the basis matrix for non-uniform cubic B-splines. The entries $m_{i,j}$ depend on local knots $\{t_i\}_{i=n-3}^{n+2}$ as detailed in~\cite{qin1998general}. Unlike the uniform case, the non-uniform cubic B-spline is locally determined here by six knots while retaining four control points. The vector $\uu=[\,1,u,u^2,u^3]^\top$ collects the powers of the normalized time $u=(t-t_{n-1})/\tau_{n-1}\in[\,0,1]$.

The orientation B-spline has a cumulative expression~\cite{li2023continuous}
\begin{equation}\label{eq:rt}
	\spr(t)=\ur_{n-4}\bullet\!\prod_{i=n-3}^n\!\Exp(\lambda_i\ude_i)\,,\,\text{with}\,\ude_i=\Log(\ur_{i-1}^{-1}\bullet\ur_i)
\end{equation}
being the increment of adjacent control points measured in the tangent space. Here, $\Exp(\cdot):\R^3\to\Sbb^3$ and $\Log(\cdot):\Sbb^3\to\R^3$ denote the exponential and logarithm maps at the identity unit quaternion $[\,1,0,0,0\,]^\top$~\cite{sola2017quaternion}. $\{\lambda_i\}_{i=n-3}^n$ are the cumulative basis functions computed by
\begin{equation}\label{eq:lam}
	[\,\lambda_{n-3},\lambda_{n-2},\lambda_{n-1},\lambda_n\,]^\top=\bmp\,\uu\,,
\end{equation}    
where $\bmp$ is the cumulative basis matrix~\cite{sommer2020efficient,qin1998general}
\begin{equation*}
	\bmp=\Bigg[\sbbmat
	1 & 0 & 0 & 0 \\
	1-m_{0,0} & 3m_{0,0} & -3m_{0,0} & m_{0,0} \\
	m_{2,0} & m_{2,1} & m_{2,2} & m_{2,3}+m_{3,3} \\
	0 & 0 & 0 & m_{3,3}
	\sebmat\Bigg]\,.
\end{equation*}	

\subsection{Kinematic Interpolations}\label{subsec:kItp}
We now present the temporal derivatives of the position and orientation B-splines $\sps(t)$ and $\spr(t)$ defined in \eqref{eq:st} and \eqref{eq:rt}, yielding the linear velocity, acceleration, and angular velocity used in the proposed continuous-time multi-sensor IEKF.

\subsubsection{Positions}\label{subsec:transItp}
Since $\sps(t)$ is linear in the control points, its derivatives follow directly from those of $\uu$:
\begin{equation}\label{eq:stKino}
	\begin{aligned}
		\rg{\sps}(t)&=[\us_{n-3}\,,\us_{n-2}\,,\us_{n-1}\,,\us_n]\,\bom\,\rg{\uu}\,,\eqwith
	\end{aligned}
\end{equation}
$\rg{(\cdot)}$ denoting any of the zeroth- to second-order time derivatives of the underlying quantity, e.g., position $\sps(t)$,  velocity $\dot\sps(t)$, and acceleration $\ddot\sps(t)$~\cite{li2025embedding}.

\subsubsection{Orientations}\label{subsec:rotItp} 
The angular velocity $\uom(t)$ in the body frame satisfies $\dot\spr(t)=\tfrac12\spr(t)\bullet\uom(t)$. Differentiating $\spr(t)$ in \eqref{eq:rt} directly scales quadratically with the spline order. Following~\cite{li2023continuous}, we instead compute $\uom(t)$ recursively according to
\begin{equation}\label{eq:omrec}
	\begin{aligned}
		\uom_1(t)&=2\dot{\lambda}_{n-2}\,\ude_{n-2}\,,\\
		\uom_2(t)&=\ue_{n-1}^{-1}\bullet\uom_1(t)\bullet\ue_{n-1}+2\dot{\lambda}_{n-1}\,\ude_{n-1}\,,\\
		\uom(t)&=\ue_n^{-1}\bullet\uom_2(t)\bullet\ue_{n}+2\dot{\lambda}_{n}\,\ude_{n}\,,
	\end{aligned}
\end{equation}
with $\ue_i=\Exp(\lambda_i\ude_i)$ and $[\dot\lambda_{n-3},\dot\lambda_{n-2},\dot\lambda_{n-1},\dot\lambda_n]^\top=\bmp\,\dot\uu$.

The Jacobians of these kinematics w.r.t. the control points are required to linearize sensor measurement models for recursive state estimation. For position, this Jacobian follows directly from the linear relation in \eqref{eq:stKino}. The Jacobians of $\spr(t)$ and $\uom(t)$ w.r.t. the orientation-increment control points follow from the chain rule applied to \eqref{eq:rt}--\eqref{eq:omrec}, as derived in~\cite{cao2025resple}.

\subsection{Spline-State-Space Model for Recursive State Estimation}\label{subsec:rse} 

The spline state-space (TriS) model was originally developed for continuous-time estimation in Euclidean space using B-splines of uniform knot intervals~\cite{li2025embedding}. Given a measurement $\uz_k$ at timestamp $t_k\in[\,t_{n-1},t_n]$, the B-spline within this time interval is determined by the following state comprising a set of recurrent control points (RCPs):
\begin{equation*}
    \ux_k=[\,\us_{n-3}^\top,\us_{n-2}^\top,\us_{n-1}^\top,\us_n^\top\,]^\top\in\R^{12}\,.
\end{equation*}
Here, $k$ indexes the time steps within the TriS model, comprising the following process and measurement models
\begin{equation}\label{eq:trisLinear}
    \begin{aligned}
        \ux_{k+1}&=\fA_k\ux_k+\uw_k\,,\\
        \uz_k&=h(\sps(\ux_k;t_k))+\uv_k\,.
    \end{aligned}
\end{equation}
$\fA_k$ governs linear RCP propagation, and $\uw_k$ denotes process noise. $h(\sps(\ux_k;t_k))$ maps the sensor-relevant kinematic quantities interpolated at $t_k$ following \Cref{subsec:kItp} to the measurement domain, corrupted by measurement noise $\uv_k$. Continuous-time trajectory estimation then boils down to recursive estimation of the RCPs via an EKF~\cite{li2025embedding}. 

Building on this framework, the TriS model was extended in~\cite{cao2025resple} for $6$-DoF continuous-time LiDAR odometry, where an iterated update step was proposed to handle point cloud association during the measurement update. The TriS model, however, has not yet been extended to non-uniform B-splines for multi-sensor continuous-time motion estimation. More importantly, the transition matrix $\fA_k$ in \eqref{eq:trisLinear} typically propagates the RCPs by extrapolating motion from preceding knot intervals, without incorporating current kinematic information. Such propagation can yield inaccurate trajectory priors during rapid changes in motion, compromising data association particularly in perceptually degenerate scenes where weak observational constraints limit subsequent correction.

\section{System Overview} \label{sec:system}

\begin{figure*}[t]
	\centering
	\adjustbox{trim={0.\width} {0.03\height} {0.\width} {0.\height},clip}{\includegraphics[width=1\linewidth]{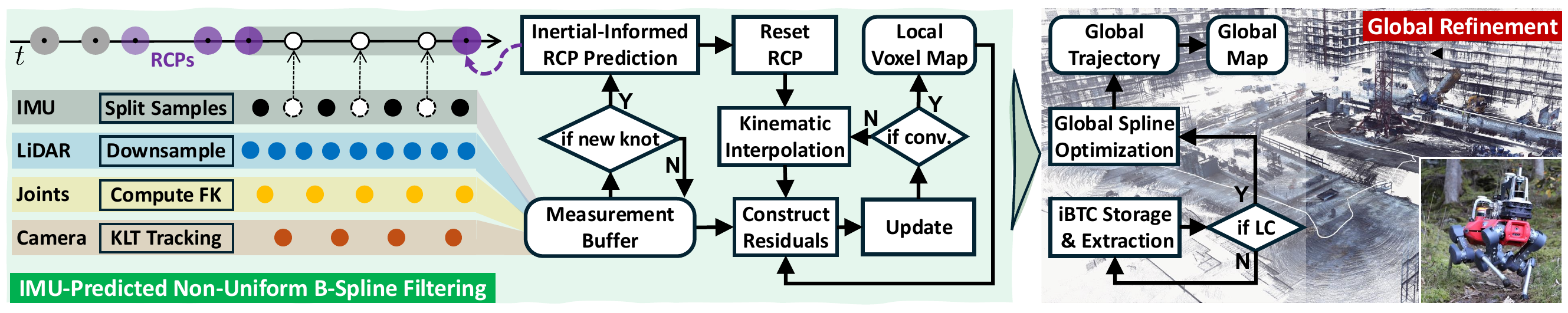}}
    \vspace{-5mm}
	\caption{System overview of the proposed RESPLE2 framework.}
	\label{fig:overview}
	\vspace{-4mm}
\end{figure*}

The proposed RESPLE2 framework is illustrated in \figref{fig:overview} in a multi-sensor setting for dynamic egomotion estimation. The basic configuration comprises an IMU and a LiDAR, with optional camera and legged-kinematic measurements (joint positions and velocities). Incoming LiDAR scans are downsampled using voxel-grid filtering, while IMU samples are split by alternating assignment into two \textit{disjoint} subsets for prediction and measurement update, respectively, over the current knot interval. When camera images are available, they are undistorted, and visual features are tracked across frames using a Kanade-Lucas-Tomasi (KLT) sparse optical-flow tracker~\cite{vins2018}. When available, leg joint measurements are preprocessed using forward kinematics to obtain foot positions and velocities with respect to the robot base frame.

We introduce a generalized spline-state-space model for the non-uniform B-splines, where a new process model is proposed based on the knot-clamping condition that guarantees control point initialization directly informed by inertial measurements on the motion trajectory (\Cref{sec:models}). Once the criteria for creating a new knot are met, we clamp a new knot to the spline endpoint and initialize its associated control point through exact RCP prediction using the IMU measurement subset (\Cref{subsec:prediction}). The knot at the previous endpoint is then unclamped to accommodate incoming measurements, for which a reset step is first performed to correct RCPs over the finalized knot spacing together with their covariance (\Cref{subsec:correction}). Afterwards, IMU measurements assigned to the update and other sensor measurements falling within the current knot interval are buffered and fused through the iterated update (\Cref{subsec:update}) via kinematic interpolation to construct residuals according to the modality-specific observation models (\Cref{subsec:meas}). This includes establishing point-to-plane correspondences between LiDAR points and the local voxel map~\cite{yuan2022efficient}. After convergence, the voxel map is updated with the undistorted LiDAR points, and new image features are initialized with depth from the co-registered LiDAR points.

The estimated control points from odometry are sent to the global refinement module that maintains the global spline trajectory and map. Loop closures are detected using iBTC~\cite{zou2024ibtc}, after which global refinement is performed directly on the non-uniform B-spline control points to correct accumulated drift while preserving local kinematic consistency (\Cref{sec:global}).

\section{Generalized Spline State-Space Modeling with Multi-Sensor Measurements} \label{sec:models}
\subsection{Generalized Spline-State-Space Model}
In our IMU-centric multi-sensor setting, we parameterize the IMU body trajectory in the world frame using the non-uniform B-splines introduced in \Cref{subsec:nbs}. The TriS model is extended accordingly. For a spline with $n$ control points, the TriS state at $t_k\in[\,t_{n-1},t_n]$ is given by
\begin{equation}\label{eq:state}
    \ux_k=[\,\uc_{n-3}^\top\,,\uc_{n-2}^\top\,,\uc_{n-1}^\top\,,\uc_{n}^\top\,,\ub_\acc^\top\,,\ub_\gyro^\top\,]^\top\in\R^{30}\,,
\end{equation}
where $\uc_i=[\us_i^\top,\ude_i^\top]^\top\in\R^6$ comprises the position RCP $\us_i\in\R^3$ and orientation RCP increment $\ude_i\in\R^3$, with $i\in\{n-3,\ldots,n\}$. As in~\cite{cao2025resple}, we adopt the tangent-space parameterization in \eqref{eq:rt} for the orientation RCPs to mitigate nonlinearity during state estimation. $\ub_\acc\in\R^3$ and $\ub_\gyro\in\R^3$ denote the accelerometer and gyroscope biases over $t_k\in[\,t_{n-1},t_n]$, respectively. In the proposed framework, we propagate RCPs by directly integrating IMU readings, thereby generalizing the TriS model to the following form
\begin{equation}\label{eq:tris}
    \begin{aligned}
        \ux_{k+1}&=f(\ux_k,\uu_k,\uw_k)\,,\\
        \uz_k&=h(\spx(\ux_k;t_k))+\uv_k\,,
    \end{aligned}
\end{equation}
where $f(\cdot)$ denotes the IMU-predicted transition function that will be derived in \Cref{subsec:pred}, and $\uu_k$ denotes the IMU samples assigned to prediction. $\uw_k$ and $\uv_k$ are zero-mean process and measurement noise with covariance $\fQ_k$ and $\bSigma_k$, respectively. In the remainder of this section, we formulate the process and measurement models within the generalized TriS framework, providing the basis for the recursive non-uniform B-spline estimator derived in the next section.

\subsection{Process Model} \label{subsec:pred}
To establish a process model for the RCPs that is directly informed by inertial measurements, we clamp each new knot at the non-uniform B-spline boundary~\cite{ahlberg2016theory,wang2026physicsinformed}, ensuring that the associated control point lies on the motion trajectory and enabling its direct prediction through IMU integration.

\subsubsection{Knot-Clamping Boundary Property}
To extend the current B-spline with $n-1$ control points and terminal knot $t_{n-1}$, we introduce a new boundary knot $t_n$ with multiplicity three, i.e., $t_n=t_{n+1}=t_{n+2}$. Evaluating the position and velocity expressions in \eqref{eq:st} and \eqref{eq:stKino} at $t_n$ (therefore $u=1$) yields
\begin{equation}\label{eq:sn}
\lsft{-12mu}
\begin{aligned}
    \sps(t_n)&=\fS\,\bom\,\uu=\us_n\,,\\
    \dot\sps(t_n)&=\fS\,\bom\,\dot\uu=3(\us_n-\us_{n-1})/\tau_{n-1}\,,\\
    \ddot\sps(t_n)&=\fS\,\bom\,\ddot\uu=6(\mu\us_{n-2}-(1+\mu)\us_{n-1}+\us_n)/\tau_{n-1}^2\,,
\end{aligned}
\lsft{-10mu}
\end{equation}
with $\fS=[\us_{n-3}\,,\us_{n-2}\,,\us_{n-1}\,,\us_n]$ being the matrix of local control points and $\mu$ a coefficient derived from knots. The orientation kinematics follow analogously:
\begin{equation}\label{eq:rn}
\begin{aligned}
\spr(t_n)&=\ur_n\eqand
\uom(t_n)=6\,\ude_n/\tau_{n-1}\,.
\end{aligned}
\end{equation}
The full derivations of \eqref{eq:sn} and \eqref{eq:rn} are given in Appendix~\ref{subsec:appClamp}. Under this clamping property for non-uniform cubic B-splines, the $6$-DoF motion trajectory $\spx(t)$ passes exactly through the new control points $\us_n$ and $\ur_n$ at $t_n$, with translational and angular velocities proportional to the final position and orientation control point increments, respectively. As the spline extends, the clamped boundary advances to each new knot, ensuring that the latest control point always lies exactly at the trajectory endpoint. This property further enables control-point prediction by directly integrating IMU measurements along the trajectory, as described next.

\subsubsection{Exact Inertial-Informed RCP Propagation}
Given the incoming IMU measurement subset $\{\uz_i^\acc,\uz_i^\gyro\}_{i=0}^{M-1}$ within $[\,t_{n-1},t_n]$, we propagate the position and orientation RCPs progressively, yielding the intermediate states $\us_{n,j}$ and $\ude_{n,j}$ across substeps $j$. We start this procedure off by expressing the continuous-time inertial kinematics as
\begin{equation}\label{eq:ct}
    \bbmat
    \dot{\us}\\\dot{\uv}\\\dot{\ur}\\\dot{\ub}_\acc\\\dot{\ub}_\gyro\ebmat=
    \bbmat
    \uv\\\fR(\uz_\acc-\ub_\acc-\un_\acc)-\ug\\\frac{1}{2}\,\ur\bullet(\uz_\gyro-\ub_\gyro-\un_\gyro)\\\un_{b_\acc}\\\un_{b_\gyro}
    \ebmat\in\R^{16}\,,
\end{equation}
with $\un_\acc\,,\,\un_\gyro\,,\,\un_{b_\acc}$, and $\un_{b_\gyro}$ being the zero-mean accelerometer, gyroscope, and bias noise terms of covariances $\fQ_\acc\,,\,\fQ_\gyro\,,\,\fQ_{b_\acc}$, and $\fQ_{b_\gyro}$, respectively. $\ug$ denotes gravity. We augment the state vector in \eqref{eq:state} with the substep velocity $\uv_{j}$, leading to the substep state
\begin{equation}\label{eq:augx}
    \lsft{-5mu}\augx_{j}=[\,\uc_{n-3:n-1}^\top,\uc_{n,j}^\top,(\ub_{j}^\acc)^\top,(\ub_{j}^\gyro)^\top,\uv_{j}^\top]^\top\lsft{-5mu}\in\R^{33}\,,
\end{equation}
where $\uc_{n-3:n-1}=[\,\uc_{n-3}^\top,\,\uc_{n-2}^\top,\,\uc_{n-1}^\top]^\top$ collects the first three RCPs. $\uc_{n,j}=[\,\us_{n,j}^\top,\ude_{n,j}^\top]^\top$ denotes the new control point propagated to substep $j$, with $j\in\{1,\ldots,M\}$.

Discretizing \eqref{eq:ct} over $\Delta{t}=t_j-t_{j-1}$ between consecutive IMU measurements within the current knot interval $[\,t_{n-1},t_n]$ yields the process model for the substep state
\begin{equation}\label{eq:pm}
    \begin{aligned}
        \augx_{j}&=f(\augx_{j-1},\uu_{j-1},\uw_{j-1})\,,\eqwith\\
        \bbmat
            \uc_{n-3:n-1}\\
            % \uc_{n-2}\\\uc_{n-1}\\
            \us_{n,j}\\\ude_{n,j}\\\ub_j^\acc\\\ub_j^\gyro\\\uv_{j}
        \ebmat
        &=\bbmat
            \uc_{n-3:n-1}\\
            % \uc_{n-2}\\\uc_{n-1}\\
            f_s(\augx_{j-1},\uu_{j-1},\uw_{j-1})\\f_\delta(\augx_{j-1},\uu_{j-1},\uw_{j-1})\\\ub_{j-1}^\acc+\un_{b_\acc}\Delta{t}\\\ub_{j-1}^\gyro+\un_{b_\gyro}\Delta{t}\\f_v(\augx_{j-1},\uu_{j-1},\uw_{j-1})
        \ebmat\,.
    \end{aligned}
\end{equation}
Here, the IMU input and process noise are specified as 
\begin{equation*}
    \begin{aligned}
    &\uu_{j-1}=[(\uz_{j-1}^\acc)^\top,(\uz_{j-1}^\gyro)^\top]^\top\eqand\\
    &\uw_{j-1}=[\,\un_\acc^\top,\,\un_\gyro^\top,\,\un_{b_\acc}^\top,\,\un_{b_\gyro}^\top]^\top\,,
    \end{aligned}
\end{equation*}
respectively. Functions $f_s$, $f_v$, and $f_\delta$ in \eqref{eq:pm} follow the standard IMU strapdown model according to
\begin{equation*}
\begin{aligned}
f_s(\augx_{j-1},\uu_{j-1},\uw_{j-1})&=\us_{n,j-1}+\uv_{j-1}\Delta{t}+0.5\ua^\ttt{W}\Delta{t}^2\,,\\
f_v(\augx_{j-1},\uu_{j-1},\uw_{j-1})&=\uv_{j-1}+\ua^\ttt{W}\Delta{t}\,,\\      
f_\delta(\augx_{j-1},\uu_{j-1},\uw_{j-1})&=\Log(\Delta\ur_{n,j})\,, \eqwith\\
\Delta\ur_{n,j}&=\Exp(\ude_{n,j-1})\bullet\Exp(\Delta{t}\uom/2)\,, 
\end{aligned}
\end{equation*}
denoting the accumulated relative rotation obtained by composing the previous accumulated rotation with the rotation increment derived from the bias-compensated angular velocity $\uom=\uz_{j-1}^\gyro-\ub_{j-1}^\gyro-\un_\gyro$ expressed in the body frame. $\ua^\ttt{W}=\fR_{n,j-1}(\uz_{j-1}^\acc-\ub_{j-1}^\acc-\un_\acc)-\ug$ is the gravity- and bias-compensated acceleration in the world frame. Here, $\fR_{n,j-1}=\fR_{n,0}\scR(\Exp(\ude_{n,j-1}))$ is the $(j-1)$-th orientation estimate obtained by composing the initial rotation $\fR_{n,0}$ with the rotation increment induced by  $\ude_{n,j-1}$.

\subsection{Multi-Sensor Measurement Model}\label{subsec:meas}
We introduce the following measurement models for LiDAR, IMU, camera, and joint measurements within the proposed RESPLE2 framework. Each model instantiates the observation function $h$ in \eqref{eq:tris} with a sensor-specific metric.

\subsubsection{LiDAR Point-to-Plane Metric}\label{subsubsec:lidar} 
Given a raw LiDAR point $\up_i^\flid$ at $t_i$, the point-to-plane distance follows
\begin{equation}\label{eq:zL}
    {h}_i^\ttt{L}(\ux_k,t_i)=\un_i^\top\big(\scR(\spr(t_i))\,\up_i^\fimu+\sps(t_i)-\ual_i\big)\,.
\end{equation}
Here, $\up_i^\fimu=\fR_\flid^\fimu\,\up_i^\flid+\us_\flid^\fimu$ transforms the LiDAR point into the IMU frame via the extrinsics, and $\spr(t_i)$, $\sps(t_i)$ interpolate the body orientation and position at $t_i$ from \eqref{eq:rt} and \eqref{eq:st}. $\ual_i$ and $\un_i$ are the nearest-neighbor point and normal of the local-map plane associated with the transformed point, obtained via VoxelMap~\cite{yuan2022efficient}. Assuming that the transformed LiDAR point lies exactly on this plane, the residual is $e_i^\ttt{L}=-{h}_i^\ttt{L}(\ux_k,t_i)$.

The measurement noise variance $\sigma_i^2$ of each point-to-plane residual accounts for both plane and point uncertainties. Following VoxelMap, each map plane carries a covariance $\bSigma_{\un,\ual}\in\R^{6\times6}$ over its normal and center, and each raw LiDAR point carries a covariance $\bSigma_{\up}\in\R^{3\times3}$ in the LiDAR frame derived from its range and bearing noise. With the point $\up_i^\fwor=\scR(\spr(t_i))\,\up_i^\fimu+\sps(t_i)$ in the world frame, a first-order propagation through \eqref{eq:zL} yields
\begin{equation*}
    \sigma_i^2=\sigma_\text{min}^2+\fJ_i^\top\bSigma_{\un,\ual}\,\fJ_i+(\un_i^\flid)^\top\bSigma_{\up}\,\un_i^\flid\,,\,\text{with}\,\,\fJ_i=\big[\sbbmat\ual_i-\up_i^\fwor\\\un_i\sebmat\big]
\end{equation*}
being the Jacobian of the residual \wrt the plane normal and center point. The normal $\un_i^\flid=(\scR(\spr(t_i))\fR_\flid^\fimu)^\top\un_i$ is expressed in the LiDAR frame, and $\sigma_\text{min}^2$ is a constant variance floor that guards against overconfident residuals.

\subsubsection{IMU Metric}\label{subsubsec:imu}
Given an IMU measurement $\uz_i^\ttt{I}=[(\uz_i^\acc)^\top,(\uz_i^\gyro)^\top]^\top$ at timestamp $t_i$, the corresponding measurement model is
\begin{equation*}
    {h}_i^\ttt{I}(\ux_k,t_i)=\bbmat\scR(\spr(t_i))^\top(\ddot{\sps}(t_i)+\ug)+\ub_{\acc}\\\uom(t_i)+\ub_{\gyro}\ebmat\,,
\end{equation*}
with $\ddot\sps(t_i)$ and $\uom(t_i)$ being the acceleration and angular velocity interpolated as in \Cref{sec:preliminaries}. The former is combined with gravity and rotated into the body frame, and both are offset by the current bias estimates. The corresponding residual is $\ue_i^\ttt{I}=\uz_i^\ttt{I}-{h}_i^\ttt{I}(\ux_k,t_i)$.

For accelerometer residuals, we opt for an adaptive outlier suppression mechanism in handling sensor noise and potential saturation. Thus, a block-Huber weight $\beta_i$ is applied to each three-dimensional acceleration residual $[\ue_i^\ttt{I}]_{1:3}$. Let $\bSigma_\acc$ denote the accelerometer measurement covariance. We compute
\begin{equation}\label{eq:huber}
    \beta_i=\begin{cases}
        1, & s_i\leq\delta_\acc\,,\\
        \delta_\acc/s_i, & s_i>\delta_\acc\,,
    \end{cases}
\end{equation}
where $s_i$ is the Mahalanobis norm of the residual with respect to $\bSigma_\acc$, and $\delta_\acc>0$ is the Huber threshold applied to $s_i$. The same scalar weight scales the precision of all three accelerometer components, yielding the reweighted covariance $\tilde{\bSigma}_{\acc,i}^{-1}=\beta_i\bSigma_\acc^{-1}$. 

\subsubsection{Visual Projection Metric}\label{subsubsec:camera}  
For each visual feature tracked from the previous image via KLT optical flow~\cite{vins2018} and observed at pixel coordinates $\uz_i^\ttt{V}$ at time $t_i$, the observation model is
\begin{equation*}
    {h}_i^\ttt{V}(\ux_k,t_i)=\pi\big(\fR_\fimu^\fcam\scR(\spr(t_i))^\top(\up_i^\fwor-\sps(t_i))+\us_\fimu^\fcam\big)\,.
\end{equation*}
Here, we transform the feature position $\up_i^\fwor$, obtained using depth from co-registered LiDAR points in previous frames, into the camera frame using the pose interpolated at $t_i$ and the IMU-camera extrinsic, and then apply the camera model $\pi(\cdot)$ to obtain the predicted pixel coordinates. The same Huber weighting as in \eqref{eq:huber} is applied to the resulting visual residual $\ue_i^\ttt{V}=\uz_i^\ttt{V}-h_i^\ttt{V}(\ux_k,t_i)$ to suppress association outliers. 

\subsubsection{Legged Kinematics Metric}\label{subsubsec:leg} 
Given joint position and velocity measurements from a legged robot at $t_i$, we compute the foot position $\up_i^\frob$ using forward kinematics~\cite{bloesch2013state} and its velocity $\uv_i^\frob$ using the corresponding kinematic Jacobian, both expressed in the robot base frame. This yields 
\begin{equation}\label{eq:zKin}
    {h}_i^\ttt{K}(\ux_k,t_i)=\bbmat\dot\sps(t_i)+\scR(\spr(t_i))(\uom(t_i)_\times\up_i^\fimu+\fR_\frob^\fimu\uv_i^\frob)\\
    \un_i^\top(\fR_\frob^\fwor\,\up_i^\frob+\us_\frob^\fwor-\ual_i)
    \ebmat\,.
\end{equation}
The first row is a foot-velocity metric that predicts the velocity of the foot in the world frame. Expressing the foot position in the IMU frame as $\up_i^\fimu=\us_\frob^\fimu+\fR_\frob^\fimu\,\up_i^\frob$, its world-frame counterpart is $\up_i^\fwor=\sps(t_i)+\scR(\spr(t_i))\,\up_i^\fimu$. Differentiating $\up_i^\fwor$ with respect to time yields the first row, which combines the velocity of the IMU body and the foot velocity in the IMU frame rotated into the world frame. The term $\uom(t_i)_\times\up_i^\fimu+\fR_\frob^\fimu\uv_i^\frob$ consists of two contributions: the velocity induced by the angular velocity $\uom(t_i)$ acting on the lever arm $\up_i^\fimu$, and the foot velocity $\uv_i^\frob$ from forward kinematics expressed in the IMU frame. The second element is a point-to-plane metric analogous to \eqref{eq:zL}. It treats the foot-ground contact as a point observed against the nearest-neighbor plane $(\ual_i,\un_i)$ of the local map. The robot base pose in the world frame is obtained by composing the interpolated IMU pose at $t_i$ with the fixed robot-to-IMU extrinsics, namely,
\begin{equation*}
    \fR_\frob^\fwor=\scR(\spr(t_i))\fR_\frob^\fimu\eqand\us_\frob^\fwor=\scR(\spr(t_i))\us_\frob^\fimu+\sps(t_i)\,.
\end{equation*}
Under the assumption of no slippage during contact, the legged kinematics residual is $\ue_i^\ttt{K}=-h_i^\ttt{K}(\ux_k,t_i)$.

\section{Inertial-Informed Iterated Extended Kalman Filter on Non-Uniform B-Splines}
Based on the process and measurement models introduced in \Cref{sec:models}, we now derive an iterated extended Kalman filter to recursively estimate the state \eqref{eq:state}, which comprises the non-uniform B-spline RCPs and IMU biases.  

\subsection{Inertial-Informed RCP Prediction}\label{subsec:prediction}
At time step $k$, given the incoming IMU subset for prediction $\{\uz_i^\acc,\uz_i^\gyro\}_{i=0}^{M-1}$ with timestamps after knot $t_{n-1}$, we first initialize the augmented state \eqref{eq:augx} and its covariance at knot $t_{n-1}$ using the clamped boundary property established in~\secref{subsec:pred}. The initial state is
\begin{equation*}
    \augx_0=[\,\uc_{n-3}^\top,\,\uc_{n-2}^\top,\,\uc_{n-1}^\top,\,\us_{n-1}^\top,\,\uzero_{3\times1}^\top,\,\ub_0^\top,\,\uv_{0}^\top\,]^\top\,,
\end{equation*}
where $\ub_0\in\R^6$ collects the accelerometer and gyroscope bias estimates from the previous posterior $\hux_{k-1|k-1}$, and $\uv_{0}=\dot\sps(t_{n-1})=3(\us_{n-1}-\us_{n-2})/\tau_{n-2}$. When introducing knot $t_n$, we initialize its position and orientation control points to those at $t_{n-1}$, giving ${\us}_{n,0}=\us_{n-1}$ and ${\ude}_{n,0}=\uzero_{3\times1}$. The augmented initial state can be thus expressed as a linear transformation of the previous posterior, namely, $\augx_0=\fT_k\hux_{k-1|k-1}$, where  
\begin{equation*}
    \fT_k=
    \begin{bmatrix}
        \fzero_{18\times6}&\multicolumn{3}{c}{\fI_{18}}&\fzero_{18\times6}\\
        \fzero_6&\fzero_6&\fzero_6&\fD^\top\fD&\fzero_6\\
        \fzero_6&\fzero_6&\fzero_6&\fzero_6&\fI_6\\
        \fzero_{3\times6}&\fzero_{3\times6}
        &-\alpha\fD&\alpha\fD&\fzero_{3\times6}
    \end{bmatrix}
    \in\R^{33\times30}\,,
\end{equation*}
with $\alpha=3/\tau_{n-2}$ and $\fD=[\,\fI_3,\fzero_{3}\,]\in\R^{3\times6}$. The augmented covariance is initialized as
\begin{equation}\label{eq:p0}
    \augP_0=\fT_k\fP_{k-1|k-1}\fT_k^\top\,.
\end{equation}
The augmented state mean and covariance are propagated recursively via the process model \eqref{eq:pm} following
\begin{equation}\label{eq:subProp}
    \begin{aligned}
    \augx_{j}&=f(\augx_{j-1},\uu_{j-1},\uzero)\eqand\\
    \augP_{j}&=\fA_{j-1}\augP_{j-1}\fA_{j-1}^\top+\fB_{j-1}\fQ_\ttt{imu}\fB_{j-1}^\top\,,
    \end{aligned}
\end{equation}
with $j$ indexing the $M$ IMU substeps. The Jacobians of the process model can be derived as
\begin{equation}\label{eq:jac}
    \setlength{\arraycolsep}{1.5pt}
    \lsft{-8mu}
    \begin{aligned}
    &\fA_{j-1}=\pad{f(\ux,\uu,\uw)}{\ux}\Bigg\vert_{\ux=\augx_{j-1}\,,\uu=\uu_{j-1}\,,\uw=\uzero}\\
    &=\bbmat
    \fI_{18}&\fzero_{18\times3}&\fzero_{18\times3}&\fzero_{18\times3}&\fzero_{18\times3}&\fzero_{18\times3}\\
    \frac{\Delta{t}^2}{2}\fF_{a,1}&\fI_3&\frac{\Delta{t}^2}{2}\fF_{a,n}&-\frac{\Delta{t}^2}{2}\fR_{n,j-1}&\fzero_3&\Delta{t}\fI_3\\
    \fzero_{3\times18}&\fzero_3&\fF_{\delta,n}&\fzero_3&\fF_{b_\gyro}&\fzero_3\\
    \fzero_{3\times18}&\fzero_3&\fzero_3&\fI_3&\fzero_3&\fzero_3\\
    \fzero_{3\times18}&\fzero_3&\fzero_3&\fzero_3&\fI_3&\fzero_3\\    
    \Delta{t}\fF_{a,1}&\fzero_3&\Delta{t}\fF_{a,n}&-\Delta{t}\fR_{n,j-1}&\fzero_3&\fI_3
    \ebmat\,,    
    \lsft{-8mu}\\
    &\fB_{j-1}=\pad{f(\ux,\uu,\uw)}{\uw}\Bigg\vert_{\ux=\augx_{j-1}\,,\uu=\uu_{j-1}\,,\uw=\uzero}\\
    &\qquad={\setlength{\arraycolsep}{2pt}\begin{bmatrix}\fzero_{18\times3}&\fzero_{18\times9}\\
    -\frac{\Delta{t}^2}{2}\fR_{n,j-1}&\fzero_{3\times9}\\
    \fzero_{9\times3}&{\diag(\fF_{b_\gyro},\Delta{t}\fI_6)}\\
    -\Delta{t}\fR_{n,j-1}&\fzero_{3\times9}
    \end{bmatrix}}\,.\\
    \end{aligned}
\end{equation}
The process noise comprises IMU measurement and bias noise, with $\fQ_\ttt{imu}=\diag(\fQ_\acc,\fQ_\gyro,\fQ_{b_\acc},\fQ_{b_\gyro})$ being the covariance. The non-zero blocks in $\fA_{j-1}$ are obtained by applying the chain rule to functions $f_s$, $f_v$, and $f_\delta$ w.r.t.\ the augmented state components:
\begin{equation*}
    \begin{aligned}
    &\fF_{a,1}=[\,\fzero_3\,,\,\fF_{a,n-3}\,,\fzero_3\,,\,\fF_{a,n-2}\,,\fzero_3\,,\,\fF_{a,n-1}]\,,\\
    &\fF_{a,n-3+\ell}=\pad{\fR_{n,j-1}(\uz_{j-1}^\acc-\ub_{j-1}^\acc)}{\ude_{n-3+\ell}}\,,\,\ell=0\,,\ldots\,,3\,,\\
    &\fF_{\delta,n}=\pad{\Log(\Delta\ur_{n,j})}{\Delta\ur_{n,j}}\lrmat{\scQ}(\Exp(\Delta{t}\uom/2))\pad{\Exp(\ude_{n,j-1})}{\ude_{n,j-1}}\,,\\
    &\fF_{b_\gyro}=-\pad{\Log(\Delta\ur_{n,j})}{\Delta\ur_{n,j}}\llmat{\scQ}(\Exp(\ude_{n,j-1}))\pad{\Exp(\Delta{t}\uom/2)}{\uom}\,.
    \end{aligned}
\end{equation*}

\subsubsection*{Knot Extension}
We introduce a new knot $t_n$ once any of the following conditions is satisfied after the current knot $t_{n-1}$:
\begin{itemize}
    \item The LiDAR batch passes the geometric sufficiency check, with point count $>N_1$ and IMU sample count $>N_2$;    
    \item The knot interval exceeds $\tau_\text{max}$;
    \item The position change $\Vert\us_{n,j}-\us_{n-1}\Vert>d_{\max}$;
    \item The orientation increment $\Vert\ude_{n,j}\Vert>\theta_\text{max}$; or
    \item The velocity change $\Vert\uv_{j}-\uv_{0}\Vert>v_{\max}$.
\end{itemize}
The first criterion aims to reduce the risk of geometric degeneracy as LiDAR points accumulate in the current batch by checking both the strength and directional balance of its geometric constraints. We first compute the following weighted normal matrix~\cite{zhang2016degeneracy}:
\begin{equation*}
    \fG=\sum_{i\in\mC}\frac{\un_i\un_i^\top}{\sigma_i^2}\,,
\end{equation*}
where $\mC$ contains preliminary point-to-plane correspondences obtained by transforming LiDAR points into the world frame using the IMU-predicted poses from substep propagation \eqref{eq:subProp} and matching the transformed points to planes in the voxel map. For each correspondence, $\un_i$ is the matched plane's normal, and $\sigma_i^2$ is the residual variance computed from the plane and point uncertainties as in \Cref{subsubsec:lidar}. Following the eigenvalue-based degeneracy criterion, we require $\lambda_\text{min}(\fG)\geq\gamma>0$ to ensure sufficient information in the weakest translational direction. For directional balance, we also require $\lambda_\text{min}(\fG)/\operatorname{tr}(\fG)\geq\eta$, where the trace measures total information and $0<\eta\leq1/3$ sets the weakest direction's minimum relative contribution. The batch passes the check only when both conditions are satisfied.

To evaluate the last three criteria, we recursively propagate the changes in state $\us_{n,j}$, $\uv_{j}$, and $\ude_{n,j}$ at each IMU substep $j$ propagated via \eqref{eq:subProp}. In a LiDAR-inertial configuration, these criteria balance the accumulation of geometric and inertial information for the next RCP update with limits on knot interval duration and accumulated rotation. For additional sensor modalities, they can be adapted or supplemented with sensor-specific heuristics. 

\subsection{RCP Reset for Unclamping}\label{subsec:correction}

\subsubsection*{Unclamping}
Introducing a new knot at $t_n$ during prediction shifts the clamped boundary from $t_{n-1}$ to $t_n$, where \eqref{eq:sn} applies. The resulting \textit{unclamping} at $t_{n-1}$ changes the local spline basis, so the trajectory no longer necessarily interpolates $\us_{n-1}$ and $\ur_{n-1}$ at that time. Without correction, this alters the trajectory prior used in the subsequent measurement update over $[t_{n-1},t_n]$, potentially compromising data association. To compensate for unclamping at the previous boundary $t_{n-1}$, we reset the RCPs after prediction. This restores the position, velocity, acceleration, orientation, and angular velocity at $t_{n-1}$ to their values before unclamping. The joint covariance is transformed accordingly.

\subsubsection*{Position Reset}
We denote quantities before and after the reset using $(\cdot)^-$ and $(\cdot)^+$, respectively. The reset preserves position, velocity, and acceleration at $t_{n-1}$ with 
\begin{equation}\label{eq:reset}
    \rg{\sps}(t_{n-1})^-=\rg{\sps}(t_{n-1})^+\,.
\end{equation}
This yields a closed-form linear transformation
\begin{equation}\label{eq:pReset}
    \us_{n-3:n}^+=\fJ_s\us_{n-3:n}^-\in\R^{12}\,,
\end{equation}
with $\us_{n-3:n}^-$ and $\us_{n-3:n}^+$ being the position RCPs before and after reset, respectively. The derivation of $\fJ_s$ is provided in Appendix~\ref{subsubsec:posReset}.

\subsubsection*{Orientation Reset}
We aim to preserve both the orientation $\spr(t_{n-1})$ and the angular velocity $\uom(t_{n-1})$ across the reset
\begin{equation*}
    \lsft{-6mu}\spr(t_{n-1})^-=\spr(t_{n-1})^+\eqand\uom(t_{n-1})^-=\uom(t_{n-1})^+\,.
\end{equation*}
To enforce this, we define a residual $\uz_{\ude}=[\uz_r^\top,\uz_\omega^\top]^\top\in\R^6$, with the orientation and angular velocity components 
\begin{equation}\label{eq:ortResetResidual}
    \begin{aligned}
    \uz_r&=[(\spr(t_{n-1})^-)^{-1}\bullet\spr(t_{n-1})^+]_{2:4}\,, \eqand\\
    \uz_\omega&=\uom(t_{n-1})^+-\uom(t_{n-1})^-\,,
    \end{aligned}
\end{equation}
respectively. Minimizing the residual $\uz_{\ude}$ in the sense of least squares does not admit a closed-form solution. We therefore solve it iteratively using Newton's method. Only $\ude_{n-2}$ and $\ude_{n-1}$ need to be reset, because the remaining control points do not appear in the residual. To solve this, we derive the residual term and its Jacobian $\fH_{\ude}\in\R^{6\times6}$ with respect to the reset increments $\ude^+=[(\ude_{n-2}^+)^\top,(\ude_{n-1}^+)^\top]^\top$, as detailed in Appendix~\ref{subsubsec:ortReset}. Starting at $\ude_0^+=\ude^-$, we obtain the increment at iteration $m$ as 
\begin{equation}\label{eq:rReset}
\Delta\ude_m^+=-(\fH_{\ude,m})^{-1}\uz_{\ude,m}
\end{equation}
and update the solution iteratively. In practice, the cost typically converges within five iterations. 

Furthermore, to preserve the originally predicted endpoint orientation $\ur_n$, the latest RCP increment $\ude_n$ must also be reset after resetting $\ude_{n-2}$ and $\ude_{n-1}$. According to $\ur_n=\ur_{n-4}\bullet\prod_{i=n-3}^n\Exp(\ude_i^-)$, this is straightforward as
\begin{equation}\label{eq:udenreset}
\begin{aligned}
\ude_n^+&=\Log\big((\ur_{n-1}^+)^{-1}\bullet\ur_n\big)\\
&=\Log\bigg(\big(\prod_{i=n-2}^{n-1}\Exp(\ude_i^+)\big)^{-1}\bullet\prod_{i=n-2}^{n}\Exp(\ude_i^-)\bigg)\,.
\end{aligned}
\end{equation}

\subsubsection*{Covariance Reset}
The RCP covariance must be transformed according to the reset map \eqref{eq:pReset} and \eqref{eq:rReset}. The bias term in the state remains unchanged during reset, but its cross-covariances with the RCPs must be transformed. The full covariance reset follows $\fP^+=\fJ\fP^-\fJ^\top$, with
\begin{equation*}
    \fP^-=\begin{bmatrix}
        \fP_{ss}^- &\fP_{sr}^- &\fP_{sb}^-\\
        \fP_{rs}^- &\fP_{rr}^- &\fP_{rb}^-\\
        \fP_{bs}^- &\fP_{br}^- &\fP_{bb}^-
    \end{bmatrix}\in\R^{30\times30}
\end{equation*}
comprising the blocks extracted from the predicted covariance $\fP_{k\vert k-1}$. The reset transformation $\fJ=\diag(\fJ_s,\fJ_r,\fI_6)$ is the Jacobian of the reset map and is block diagonal, because the position and orientation reset maps depend only on their respective pre-reset variables and biases remain unchanged. We provide complete derivations of $\fJ_s$ and $\fJ_r$ in Appendix~\ref{subsec:appReset}. 

\subsection{Iterated Update}\label{subsec:update}
We apply an iterated update for incorporating the multi-sensor measurements introduced in \Cref{subsec:meas}. Starting from $\hux_0=\hux_{k|k-1}$, the estimate is refined over iterations $i=0,1,\ldots,N_\mathrm{max}-1$ via
\begin{equation*}
	\begin{aligned}
		&\delta\ux_i=\fK_i(\uz_k-h(\hux_i))-(\fI-\fK_i\fH_i)(\hux_i-\hux_{k|k-1})\,,\\
		&\text{with}\quad\fK_i=\fP_{k|k-1}\fH_i^\top(\fH_i\fP_{k|k-1}\fH_i^\top+\bSigma_k)^{-1}\,.
	\end{aligned}
\end{equation*}
We collect all the multi-modal measurements in the current buffer into $\uz_k$, with associated covariance $\bSigma_k$. Here, $h(\hux_i)$ is the observation model evaluated at the current iterate $\hux_i$. $\fH_i=\pad{h}{\ux}\big\vert_{\hux_i}$ is the corresponding observation Jacobian. The iteration terminates once $\nm{\delta\ux_i}$ falls below a threshold or a maximum number of iterations is reached, yielding the posterior $\hux_{k|k}=\hux_i+\delta\ux_i$ and $\fP_{k|k}=(\fI-\fK_i\fH_i)\fP_{k|k-1}$. 

\subsection{Filter Setup} \label{subsec:filter}
We summarize the full recursive non-uniform spline filtering in Algorithm~\ref{alg:resple2}, including inertial-informed RCP prediction, reset, and the iterated update. In the update step, the Jacobians for the LiDAR, IMU, and camera measurement models in \Cref{subsec:meas} are straightforward to derive~\cite{cao2025resple,nguyen2026third}. For legged joints, we hereby derive the Jacobian of the zero-velocity part of the legged kinematic measurement model in \eqref{eq:zKin} w.r.t.\ the position and orientation RCPs as
\begin{equation*}
    \begin{aligned}
        &\pad{{h}_i^{\ttt{K},v}}{\us_{n-3:n}}=(\bom\,\dot\uu\otimes\fI_3)^\top\,,\\
        &\pad{{h}_i^{\ttt{K},v}}{\ude_{n-3:n}}=\pad{{h}_i^{\ttt{K},v}}{\spr(t_i)}\pad{\spr(t_i)}{\ude_{n-3:n}}+\pad{{h}_i^{\ttt{K},v}}{\uom(t_i)}\pad{\uom(t_i)}{\ude_{n-3:n}}\,,\eqwith\\
        &\pad{{h}_i^{\ttt{K},v}}{\spr(t_i)}=\pad{(\scR(\spr(t_i))\,\uv_i^\fimu)}{\spr(t_i)}\,,\,\pad{{h}_i^{\ttt{K},v}}{\uom(t_i)}=-\scR(\spr(t_i))(\up_i^\fimu)_\times\,,
    \end{aligned}
\end{equation*}
with $\uv_i^\fimu=\uom(t_i)_\times\up_i^\fimu+\fR_\frob^\fimu\uv_i^\frob$. The Jacobians $\pad{\spr(t_i)}{\ude_{n-3:n}}$ and $\pad{\uom(t_i)}{\ude_{n-3:n}}$ are available in~\cite{cao2025resple} as in \Cref{subsec:rotItp}. The Jacobian blocks of IMU biases are zero. The Jacobian of the foot position observation in the second row of \eqref{eq:zKin} resembles that of the LiDAR residual~\cite{cao2025resple}.

\begin{algorithm}[t!]
    \setlength{\hsize}{\dimexpr\algowidth-\algomargin\relax}
    \caption{Inertial-Informed Recursive Non-Uniform B-Spline Estimation}
    \label{alg:resple2}
    \KwIn{posterior $\hux_{k-1|k-1}$, $\fP_{k-1|k-1}$, IMU subset $\{\uz_i^\ttt{imu}\}_{i=0}^{M-1}$ for prediction, measurement batch $\uz_k$} 
    \KwOut{posterior $\hux_{k|k}$, $\fP_{k|k}$}
    {\color{blue}\tcc{IMU-Informed RCP Prediction}}
    $\augx_0\gets[\,\uc_{n-3}^\top,\uc_{n-2}^\top,\uc_{n-1}^\top,\us_{n-1}^\top,\uzero_3^\top,\ub_0^\top,\uv_{0}^\top]^\top$\,\;
    $\augP_0\gets\fT_k\fP_{k-1\vert{k-1}}\fT_k^\top$\,\tcp*[r]{\eqref{eq:p0}}
    $\fQ_\ttt{imu}\gets\diag(\fQ_\acc\,,\fQ_\gyro\,,\fQ_{b_\acc}\,,\fQ_{b_\gyro})$\,\;
    \For{$j\gets$ $1$ \KwTo $M$}{
        $\fA_{j-1},\fB_{j-1}\gets\texttt{Jacobian}(\augx_{j-1},\uz_{j-1}^{\ttt{imu}})$\,\tcp*[r]{\eqref{eq:jac}}
        $\augx_{j}\gets f(\augx_{j-1},\uu_{j-1},\uzero)$\,\;
        $\augP_{j}\gets\fA_{j-1}\augP_{j-1}\fA_{j-1}^\top+\fB_{j-1}\fQ_\ttt{imu}\fB_{j-1}^\top$\,\;
    }
    $\hux_{k|k-1}\gets[\,\augx_{M}\,]_{1:30}$\,; $\fP_{k|k-1}\gets[\,\augP_{M}\,]_{1:30,1:30}$\,\;
    {\color{blue}\tcc{RCP Reset}}
    $\hux_{k|k-1},\fP_{k|k-1}\gets\ttt{reset}(\hux_{k|k-1},\fP_{k\vert{k-1}})$\,\;
    {\color{blue}\tcc{Iterated Update}}
    $i\gets0$\,; $\hux_i\gets\hux_{k|k-1}$\,\;
    \While{\ttt{true}}{
        $\bm{\gamma}_i\gets\uz_k-h(\hux_i;t_k^\ttt{z})$\,\;
        $\fH_i\gets\texttt{computeJacobian}(\hux_i,t_k^\ttt{z})$\,\;
        $\fK_i\gets\fP_{k|k-1}\fH_i^\top(\fH_i\fP_{k|k-1}\fH_i^\top+\bSigma_k)^{-1}$\,\;
        $\delta\ux_i\gets\fK_i\bm{\gamma}_i-(\fI-\fK_i\fH_i)(\hux_i-\hux_{k|k-1})$\,\;
        $\hux_{i+1}\gets\hux_i+\delta\ux_i$\,\;
        \If{$\Vert\delta\ux_i\Vert<\epsilon$ \bf{or} $i+1=N_\mathrm{max}$}{
            \ttt{break}\,\;
        }
        $i\gets{i}+1$\,\;
    }
    $\hux_{k|k}\gets\hux_{i+1}$\,; $\fP_{k|k}\gets(\fI-\fK_i\fH_i)\fP_{k|k-1}$\,\;
    \Return $\hux_{k|k}\,,\fP_{k|k}$
\end{algorithm}

\section{Global Refinement} \label{sec:global}
\subsection{Loop Closure Detection}
The image-assisted binary and triangle descriptor iBTC~\cite{zou2024ibtc} combines registered LiDAR point clouds, images, and odometry-estimated camera poses for place recognition. RESPLE2 employs iBTC to detect loop closures between the current LiDAR scan and historical keyframes in the LiDAR-visual-inertial configuration. The module aggregates consecutive scans into keyframe submaps, retrieves historical candidates using geometric and image-assisted descriptors, and verifies matches through geometric registration. Accepted matches provide paired timestamps and relative $6$-DoF transformations as constraints for global spline optimization.

\subsection{Global Spline Optimization}
Once a loop closure is detected by iBTC, we perform global optimization over the existing non-uniform B-spline trajectory to correct accumulated drift while preserving higher-order kinematics, such as angular and translational velocities, by directly refining the global state $\cpX_\twi=[\us_1^\top,\ur_1^\top,\cdots,\us_N^\top,\ur_N^\top]^\top$, comprising $N$ control points from odometry. We obtain the refined state $\cpX_\twi^\star$ by minimizing
\begin{equation}\label{eq:opt}
	\sum_{k=1}^{m}\big\Vert\scE_\ttt{lc}(\cpX_\twi,\uz_{\ttt{lc},k})\big\Vert^2_{\fC_\ttt{lc}}
	+\sum_{i=1}^{N-1}\big\Vert\scE_\ttt{od}(\cpX_\twi,\uz_{\ttt{od},i})\big\Vert^2_{\fC_\ttt{od}}\,,
\end{equation}
which combines $m$ loop-closure and $N-1$ control-point odometry residuals. The former follows
\begin{equation*}
	\scE_\ttt{lc}(\cpX_\twi,\uz_{\ttt{lc},k})=
    \bbmat
    \scR(\spr(t_i))^\top\big(\sps(t_j)-\sps(t_i)\big)-\uz_{ij}^s\\
    \Log\big((\uz_{ij}^r)^{-1}\bullet\spr(t_i)^{-1}\bullet\spr(t_j)\big)
	\ebmat\,,
\end{equation*}
with $\uz_{ij}^s$ and $\uz_{ij}^r$ denoting the relative translation and rotation, respectively, in the transformation $\uz_{\ttt{lc},k}$ between keyframes $i$ and $j$ provided by iBTC. To preserve the local kinematic consistency given by odometry during global optimization, we constrain the relative translation and rotation between adjacent control points according to 
\begin{equation*}
	\scE_\ttt{od}(\cpX_\twi,\uz_{\ttt{od},i})=\begin{bmatrix}
		\scR(\ur_i)^\top\big(\us_{i+1}-\us_i\big)-\uz_{i,i+1}^s\\
		\Log\big((\uz_{i,i+1}^r)^{-1}\bullet\ur_i^{-1}\bullet\ur_{i+1}\big)
	\end{bmatrix}\,,
\end{equation*}
with $\uz_{i,i+1}^s$ and $\uz_{i,i+1}^r$ denoting the relative translation and rotation in $\uz_{\ttt{od},i}$, respectively, between two consecutive control points estimated by odometry. Unlike existing approaches to global B-spline correction~\cite{lv2021clins}, our formulation in \eqref{eq:opt} applies odometry constraints directly to the non-uniform control points, requiring spline interpolation only for the infrequent loop-closure residuals. This limits the Jacobian support of each odometry residual to two control points, preserving sparsity and improving the efficiency of solving the optimization problem at scale via the standard Levenberg-Marquardt algorithm.
	
\section{Evaluation} \label{sec:eva}

\subsection{Implementation}\label{subsec:setting}
We implement RESPLE2 as a ROS2 package with separate nodes for multi-sensor odometry and global spline refinement. We use a minimal set of dependencies, including Eigen for numerical computations and OpenMP to parallelize B-spline kinematic interpolation, LiDAR point processing, nearest-neighbor association, and linearization during global optimization. Each node is typically allocated five OpenMP threads. We use Pinocchio for computing leg forward kinematics~\cite{pinocchio}. The global refinement solver is implemented from scratch and solves the sparse linear systems using Eigen's CHOLMOD interface, reusing symbolic factorization across iterations.

We combine the sensor models in \Cref{subsec:meas} to enable LiDAR-inertial (LIO), LiDAR-visual-inertial (LVIO), LiDAR-leg-inertial (LKIO), and LiDAR-leg-visual-inertial (LKVIO) odometry within the same filtering scheme, where multi-modal observations in each knot interval are stacked into a single measurement vector. Additionally, following the system design of RESPLE2, we provide a LiDAR-only variant (RESPLE2-LO) to support platforms without an IMU. Here, the original inertial-informed process model is replaced by a constant-velocity model with the same knot-clamping boundary property, predicting position using the linear velocity estimate and scaling the latest orientation increment by the ratio of successive knot intervals. Accordingly, the reset preserves the position, orientation and linear velocity at the previous endpoint while imposing zero acceleration.

\subsection{Benchmarking Setup}\label{subsec:bench}
We evaluate RESPLE2 under diverse real-world conditions and sensor configurations using the public datasets \ttt{GrandTour}~\cite{frey2026grandtour} and \ttt{JEPLO}~\cite{yuan2026jeplo}, together with our own experimental datasets \ttt{PPS Tracked Vehicle} (\ttt{PPS TV}), \ttt{HOPPER}, and \ttt{HANDI}. These datasets are summarized in Tab.~\ref{tab:dataset}, along with their sensor configurations and scenarios. 

\begin{table}[htbp]
	\vspace{-2mm}
    \scriptsize
    \centering
    \setlength{\tabcolsep}{2pt}
    \caption{Public and experimental datasets for benchmarking.}
    \begin{tabular}{
        >{\raggedright\arraybackslash}p{0.15\columnwidth}
        >{\raggedright\arraybackslash}p{0.43\columnwidth}
        >{\raggedright\arraybackslash}p{\dimexpr0.4\columnwidth-6\tabcolsep\relax}}
        \toprule 
        \textbf{Dataset} & \textbf{Deployment Scenario} & \textbf{Adopted Sensors}$^*$\\
        \midrule
        \ttt{GrandTour}\newline\cite{frey2026grandtour} &ANYmal D quadruped, indoor, outdoor, and wilderness environments &\ttt{Hesai} (\textbf{L}), \ttt{Alpha} (\textbf{V}),  \ttt{ICM40609} (\textbf{I}), leg joints (\textbf{K}) \\

        \rowcolor{gray!15} \ttt{JEPLO}~\cite{yuan2026jeplo} & Unitree Go2 quadruped, maneuvers over stairs and boxes &\ttt{Mid360} (\textbf{L}), \ttt{ICM40609} (\textbf{I}), leg joints (\textbf{K}) \\
     
        \ttt{PPS TV} (experiment) & Tracked vehicle, high-speed maneuvers in outdoor environments &  \ttt{Hesai} (\textbf{L}), \ttt{HDR} (\textbf{V}), \ttt{ICM40609} (\textbf{I})\\
     
        \rowcolor{gray!15}\ttt{HOPPER} (experiment) &Legged quadcopter~\cite{li2025high}, highly dynamic maneuvers in enclosed and open spaces  &\ttt{Mid360} (\textbf{L}), \ttt{ICM40609} (\textbf{I1}), \ttt{BMI088} (\textbf{I2})\\
     
        \ttt{HANDI} (experiment) &Handheld platform, long-distance walking in indoor and outdoor campus environments  &\ttt{Mid360} (\textbf{L1}), \ttt{ICM40609}
        (\textbf{I1}),  \ttt{Avia} (\textbf{L2}), \ttt{BMI088} (\textbf{I2}), \ttt{D455} (\textbf{V1}), \ttt{ZedX1} (\textbf{V2}) \\
        \bottomrule
    \end{tabular}
    \parbox{0.97\columnwidth}{
        \vspace{4pt}
        $^*$LiDARs: \ttt{Hesai} (Hesai XT32),  \ttt{Mid360} (Livox Mid360), and \ttt{Avia} (Livox Avia LiDAR). IMUs: \ttt{ICM40609} (built in \ttt{Mid360}), \ttt{BMI088} (Bosch Sensortec \ttt{BMI088} IMU, and \ttt{Avia} built-in IMU). Cameras: \ttt{Alpha} (Alphasense, front-center), \ttt{HDR} (TierIV C1 HDR, front),  \ttt{D455} (RealSense D455), and \ttt{ZedX1} (ZED X One).
        }
    \label{tab:dataset}
\end{table}

Our baselines include state-of-the-art LVIO systems Coco-LIC~\cite{lang2023coco} (CoLIC) and FAST-LIVO2~\cite{zheng2024fast} (FLIVO2). We also involve CTE-MLO~\cite{shen2025cte} (CMLO) for comparison with GP-based continuous-time estimation, Point-LIO~\cite{he2023point} (PLIO) for LIO under highly dynamic motion, and Leg-KILO~\cite{ou2024leg} (LeKILO) for LKIO on quadrupedal robots. Where applicable, we also evaluate Coco-LIC and FAST-LIVO2 for LIO. In this setting, FAST-LIVO2 has been reported to consistently outperform FAST-LIO2, with the gains attributed to its voxel-based local map and accurate noise modeling of LiDAR points~\cite{zheng2024fast}. Likewise, we include RESPLE~\cite{cao2025resple}, replacing its \textit{ikd-Tree} local map~\cite{cai2021ikd} with VoxelMap~\cite{yuan2022efficient} to match the local map and LiDAR point noise modeling in RESPLE2, thereby helping to isolate differences in estimation performance.

All evaluations are performed on a desktop running Ubuntu 24.04 with an AMD Ryzen 9 9950X CPU and 32\,GB of RAM. We use a single set of parameters across all sequences for each dataset and sensor setting. Unless otherwise stated, trajectory estimates are compared against the ground truth using the \textit{evo} package~\cite{grupp2017evo}, with RPEs computed over fixed \SI{0.5}{\meter} path-length increments using \textit{evo}'s point-distance pose relation. In the following tables of reported results, boldface and underlining indicate the best and second-best results, respectively, while \xmark\, denotes an error exceeding \SI{50}{\meter}.

\subsection{Public Datasets}

\subsubsection{\ttt{GrandTour}}

\begin{table*}[t!]
    \centering
    \scriptsize
    \setlength{\tabcolsep}{1.15pt}
    \caption{APE and RPE (RMSE, meters) on \ttt{GrandTour}.}
    \begin{tabular}{ll*{10}{cc}}
        \toprule
        & & \multicolumn{2}{c}{\ttt{ARC-2}$^*$} & \multicolumn{2}{c}{\ttt{ARC-7}$^*$} & \multicolumn{2}{c}{\ttt{CON-4}$^*$} & \multicolumn{2}{c}{\ttt{EIG-1}$^*$} & \multicolumn{2}{c}{\ttt{SNOW-2}$^*$} & \multicolumn{2}{c}{\ttt{SPX-2}$^*$} & \multicolumn{2}{c}{\ttt{ARC-3}$^\dagger$} & \multicolumn{2}{c}{\ttt{HIL-1}$^\dagger$} & \multicolumn{2}{c}{\ttt{LEE-1}$^\dagger$} & \multicolumn{2}{c}{\ttt{PIL-2}$^\dagger$}\\
        \cmidrule(lr){3-4} \cmidrule(lr){5-6} \cmidrule(lr){7-8} \cmidrule(lr){9-10} \cmidrule(lr){11-12} \cmidrule(lr){13-14} \cmidrule(lr){15-16} \cmidrule(lr){17-18} \cmidrule(lr){19-20} \cmidrule(lr){21-22}
        {Method} & {Variant} & APE & RPE & APE & RPE & APE & RPE & APE & RPE & APE & RPE & APE & RPE & APE & RPE & APE & RPE & APE & RPE & APE & RPE\\
        \midrule
        CMLO & LO & $3.090$ & $0.360$ & $0.182$ & $0.061$ & $0.024$ & $\underline{0.015}$ & $0.041$ & $0.023$ & $0.023$ & $0.016$ & $0.104$ & $0.030$ & \xmark & \xmark & $0.063$ & $0.025$ & \xmark & \xmark & \xmark & \xmark\\
        \rowcolor{gray!15} CoLIC & LVIO & $0.095$ & $0.028$ & $0.052$ & $\bf{0.005}$ & $0.027$ & $\bf{0.005}$ & $0.048$ & $\bf{0.006}$ & $\bf{0.012}$ & $\bf{0.005}$ & $0.040$ & $\bf{0.006}$ & $0.154$ & $\bf{0.019}$ & $0.031$ & $0.009$ & $0.051$ & $\underline{0.006}$ & ${0.031}$ & $0.013$\\
        FLIVO2 & LVIO & $\bf{0.012}$ & $0.009$ & $\bf{0.013}$ & $\underline{0.014}$ & $0.030$ & $\underline{0.015}$ & $\underline{0.033}$ & $0.014$ & $0.016$ & $0.016$ & $\underline{0.023}$ & $0.013$ & \xmark & \xmark & $\bf{0.018}$ & $\underline{0.008}$ & $0.037$ & $0.011$ & $0.071$ & $0.025$\\
        \rowcolor{gray!15} RESPLE & LIO & $0.109$ & $0.055$ & $0.347$ & $0.142$ & $0.054$ & $0.024$ & $0.107$ & $0.022$ & $0.026$ & $0.021$ & $0.042$ & $\underline{0.012}$ & \xmark & \xmark & $0.036$ & $0.009$ & $0.041$ & $0.020$ & \xmark & \xmark\\
        & LIO & $\bf{0.012}$ & $\bf{0.006}$ & $0.036$ & $0.016$ & $\underline{0.016}$ & $0.021$ & $0.034$ & $0.008$ & $0.014$ & $\underline{0.009}$ & $\bf{0.022}$ & $0.013$ & $0.078$ & $0.024$ & $0.022$ & $\bf{0.004}$ & $\underline{0.020}$ & $\bf{0.005}$ & $0.029$ & $0.010$\\
        \rowcolor{gray!15} & LVIO & $\bf{0.012}$ & $\underline{0.007}$ & $0.038$ & $0.016$ & $0.018$ & $0.020$ & $\bf{0.032}$ & $0.008$ & $0.014$ & $\underline{0.009}$ & $\underline{0.023}$ & $0.013$ & $0.072$ & $0.024$ & $\underline{0.020}$ & $\bf{0.004}$ & $\bf{0.017}$ & $\underline{0.006}$ & $\underline{0.028}$ & $0.009$\\
         & LKIO & $\underline{0.013}$ & $0.008$ & $0.032$ & $\underline{0.014}$ & $\bf{0.015}$ & $0.020$ & $0.034$ & $\underline{0.007}$ & $\underline{0.013}$ & $\underline{0.009}$ & $\bf{0.022}$ & $\underline{0.012}$ & $\bf{0.056}$ & $\underline{0.023}$ & $0.023$ & $\bf{0.004}$ & $0.027$ & $\underline{0.006}$ & $\underline{0.028}$ & $\bf{0.007}$\\
        \rowcolor{gray!15} \multirow{-4}{*}{RESPLE2} & LKVIO & $\bf{0.012}$ & $\underline{0.007}$ & $\underline{0.017}$ & $\underline{0.014}$ & $\bf{0.015}$ & $0.020$ & $\underline{0.033}$ & $\underline{0.007}$ & $\underline{0.013}$ & $\underline{0.009}$ & $\underline{0.023}$ & $\underline{0.012}$ & $\underline{0.060}$ & $\underline{0.023}$ & $0.021$ & $\bf{0.004}$ & ${0.023}$ & $\underline{0.006}$ & $\bf{0.026}$ & $\underline{0.008}$\\
        \bottomrule
    \end{tabular}
    \parbox{2\columnwidth}{%
        \vspace{4pt}
        $^*$Test sequences: RESPLE2 results are obtained by submitting pose estimates to the COMFORT Localization Benchmark\footnotemark. Results for the other systems are taken from~\cite{frey2026grandtour} and were obtained using the high-end \ttt{HG4930} MEMS IMU (unavailable to RESPLE2). $^\dagger$Training sequences: All systems are evaluated using the \ttt{Mid360} built-in IMU.
        }
    \vspace{-4mm}
    \label{tab:grandtour}
\end{table*}
\footnotetext{Available at \url{https://grand-tour.leggedrobotics.com/tasks/localization}}

The \ttt{GrandTour} dataset\footnote{Available at \url{https://grand-tour.leggedrobotics.com/dataset}} was collected using an ANYmal D quadruped equipped with the \textit{Boxi} multi-sensor rig~\cite{frey2025boxi}, including  LiDAR, camera, inertial, and leg-joint measurements for studying legged autonomy across diverse environments. We evaluate all six test sequences covering diverse egomotions. We additionally select four training sequences featuring challenging conditions, including stairs and hills, with potential LiDAR and visual degeneracy.

As summarized in Tab.~\ref{tab:grandtour}, despite using the lower-end \ttt{Mid360} built-in IMU, RESPLE2 variants achieve overall competitive or superior estimation accuracy. Noticeably, RESPLE2-LIO achieves competitive accuracy compared with Coco-LIC and FAST-LIVO2 in APE without visual input. Within the RESPLE2 framework, adding visual or kinematic measurements to the LIO configuration improves performance, with RESPLE2-LKVIO delivering the best overall performance among our variants. Compared with RESPLE, RESPLE2-LIO achieves lower APE and RPE on nearly all sequences, supporting the benefits of the proposed non-uniform B-spline estimation scheme. Coco-LIC exhibits low RPE, reflecting strong local consistency, which may be attributed to its sliding-window optimization of short-range relative motion. However, the corresponding APEs indicate that this does not translate into superior global accuracy. Examples of running RESPLE2 on \ttt{GrandTour} are shown in \figref{fig:front}-(a,b).
    
\begin{figure}[t!]
    \centering
    \begin{tabular}{@{}c@{\hspace{4pt}}c@{}}
        \framelabel[south west]{\adjustbox{trim={0.1\width} {0.12\height} {0.1\width} {0.2\height},clip}{\includegraphics[width=0.64\linewidth]{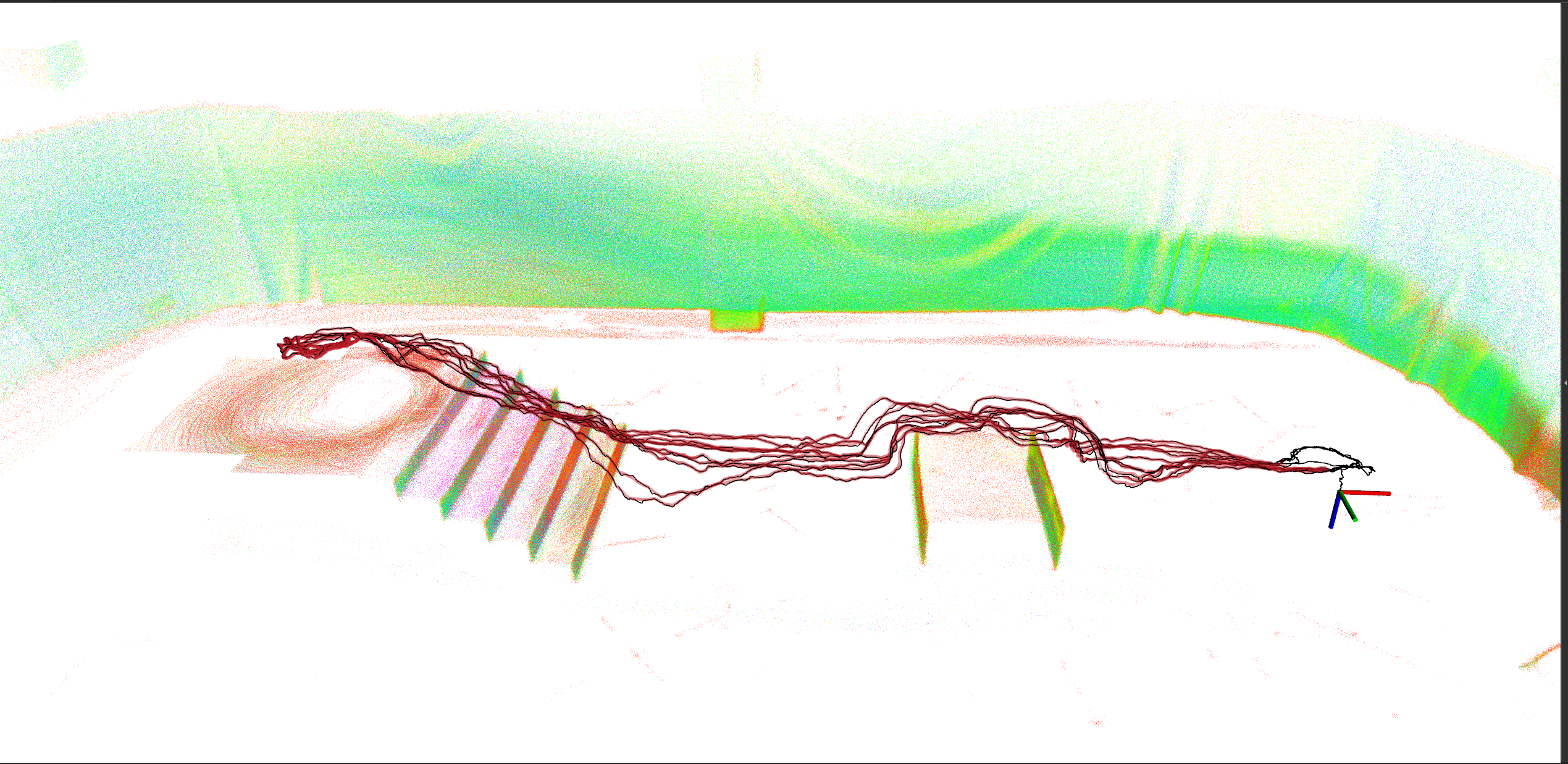}}}{\ttt{EIL\_Mix}}&
        \framelabel[south west]{\adjustbox{trim={0.2\width} {0.18\height} {0.15\width} {0.2\height},clip}{\includegraphics[width=0.7\linewidth]{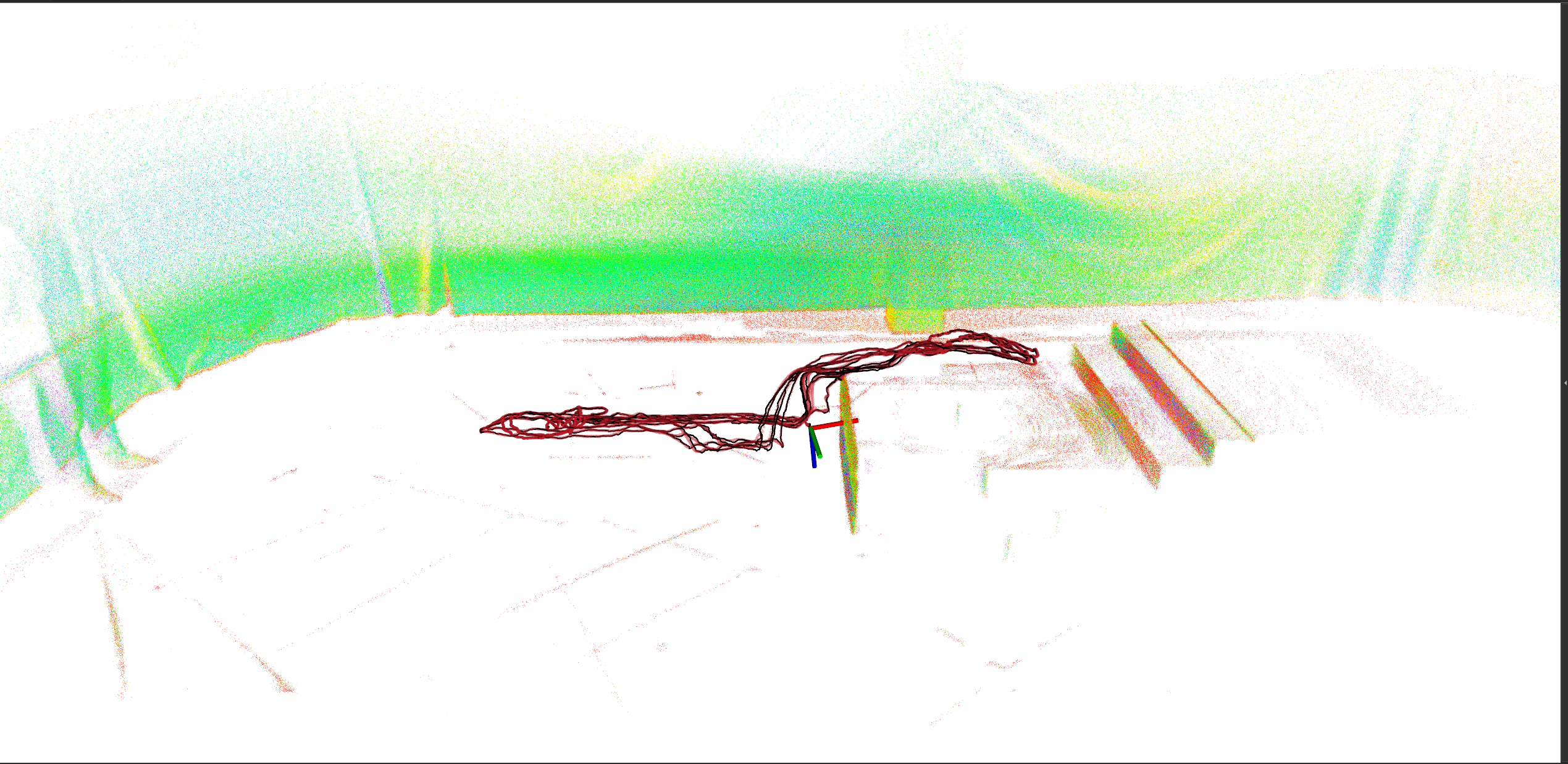}}}{\ttt{EIL\_Box}}\\
        \multicolumn{2}{@{}p{\linewidth}@{}}{\scriptsize\raggedright (a) RESPLE2 on two sequences of the \ttt{JEPLO} dataset. Black and red curves indicate estimates and ground truth, respectively.\par}\\[12pt]
        \multicolumn{2}{@{}c@{}}{\adjustbox{trim={0.05\width} {0.24\height} {0.05\width} {0.15\height},clip}{\includegraphics[width=1.1\linewidth]{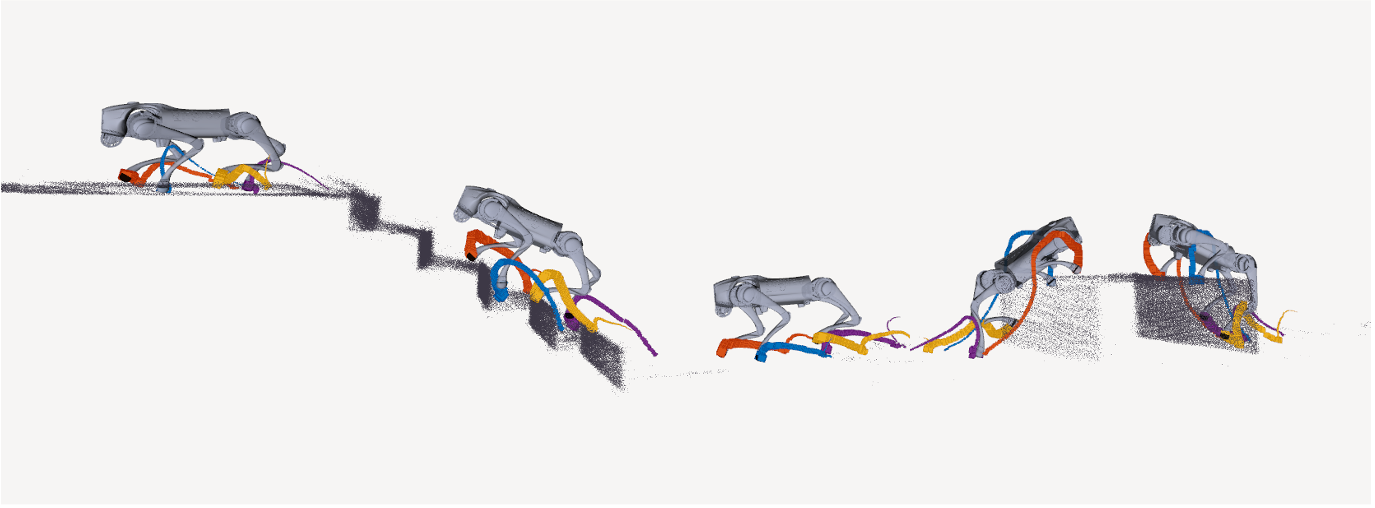}}}\\
        \multicolumn{2}{@{}p{\linewidth}@{}}{\scriptsize\raggedright (b) Foot position estimates from RESPLE2 during the \ttt{EIL\_Mix} sequence.\par}
    \end{tabular}
    \caption{Examples of RESPLE2 tested on \ttt{JEPLO}. }
    \label{fig:jeplo}
    \vspace{-4mm}
\end{figure}

\subsubsection{\ttt{JEPLO}} 
The \ttt{JEPLO} dataset is recorded onboard a Unitree Go2 quadruped driven by a perceptive locomotion policy to perform dynamic maneuvers at approximately \SI{1}{\meter/\second} across varying terrains, including long staircases and boxes, in both indoor and outdoor scenarios~\cite{yuan2026jeplo}, as shown in \figref{fig:front}-(e) for \ttt{Outdoor2} and \figref{fig:jeplo}-(a). A \ttt{Mid360} is mounted upside down on the robot's head, providing a limited FoV of the surroundings. In the \ttt{EIL\_Mask1} and \ttt{EIL\_Mask2} sequences, taped masks partially cover the sensor to deliberately restrict its FoV and induce perceptual degeneracy.

As shown in Tab.~\ref{tab:jeplo}, RESPLE2 achieves the best overall APE performance, with its LKIO variant attaining the lowest error on all eight sequences. Incorporating leg kinematic measurements improves or maintains estimation accuracy across all sequences.  FAST-LIVO2 performs substantially worse, possibly because its discrete-time formulation is sensitive to abrupt changes in the observed scene within the limited LiDAR FoV and to high-impact landings when the robot steps down from elevated platforms. Leg-KILO follows the filtering-based design of Point-LIO with point-wise measurement update. Like RESPLE2, it uses VoxelMap for local map management and incorporates a foot-velocity measurement model. Despite these shared design choices, RESPLE2-LKIO consistently achieves lower estimation errors than Leg-KILO, even without customization for legged odometry. This supports the benefits of our proposed continuous-time estimation method. Foot position estimation is demonstrated in \figref{fig:jeplo}-(b).

\begin{table}[htbp]
    \vspace{-2mm}
    \scriptsize               
    \centering
    \setlength{\tabcolsep}{2.3pt}
    \caption{APE (RMSE, meters) on \ttt{JEPLO}.}
    \begin{tabular}{r|cccc|cc}
        \toprule
        {Method} & {CoLIC} & {FLIVO2} & {RESPLE} & {LeKILO} & \multicolumn{2}{c}{RESPLE2} \\
        {Variant} & LIO & LIO & LIO & LKIO & LIO & LKIO \\
        \midrule
        \ttt{EIL\_Box}  &$0.054$ &$3.200$ &${0.011}$ &$0.032$ &$\underline{0.006}$ &$\bf{0.005}$ \\
        \rowcolor{gray!15} \ttt{EIL\_Mask1}  &$0.032$ &$0.551$ &$0.015$ &$0.032$ &$\underline{0.009}$ &$\bf{0.008}$ \\
        \ttt{EIL\_Mask2}  &$0.045$ &$0.830$ &${0.015}$ &$0.040$ &$\underline{0.011}$ &$\bf{0.010}$ \\
        \rowcolor{gray!15} \ttt{EIL\_Mix}  &$0.028$ &$1.628$ &$\underline{0.015}$ &$0.028$ &$\bf{0.010}$ &$\bf{0.010}$ \\
        \ttt{EIL\_Stairs}  &$0.047$ &\xmark &$0.013$ &$0.052$ &$\underline{0.010}$ &$\bf{0.007}$ \\
        \rowcolor{gray!15} \ttt{Long\_Stairs}  &$0.015$ &\xmark &$\underline{0.010}$ &$0.027$ &$\bf{0.007}$ &$\bf{0.007}$ \\
        \ttt{Outdoor1}  &$0.017$ &\xmark &${0.010}$ &$0.018$ &$\underline{0.008}$ &$\bf{0.007}$ \\
        \rowcolor{gray!15} \ttt{Outdoor2}  &$0.016$ &\xmark &$0.011$ &$0.026$ &$\underline{0.007}$ &$\bf{0.006}$ \\
        \bottomrule
    \end{tabular}
    \vspace{-2mm}
    \label{tab:jeplo}
\end{table}

\subsection{Experiments}

\subsubsection{\ttt{PPS TV}} 
\begin{figure}[t!]
    \centering
    \begin{tabular}{@{}c@{\hspace{4pt}}c@{\hspace{4pt}}c@{}}
    \framelabel[south west]{\adjustbox{trim={0.\width} {0.\height} {0.\width} {0.\height},clip}{\includegraphics[width=0.295\linewidth]{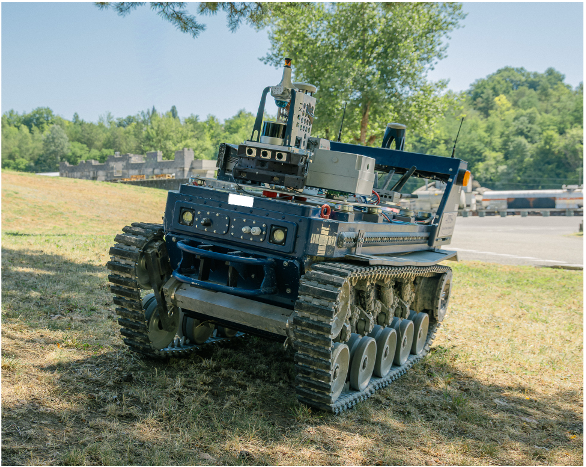}}}{\normalsize{\ttt{(a)}}}&
    \framelabel[south west]{\adjustbox{trim={0.\width} {0.\height} {0.\width} {0.\height},clip}{\includegraphics[width=0.68\linewidth]{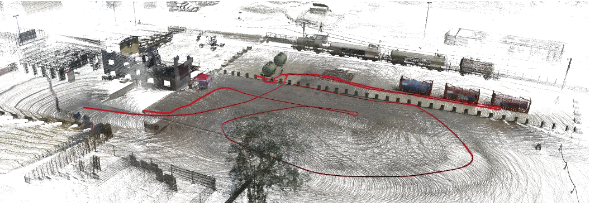}}}{\normalsize{\ttt{(b)}}}
    \end{tabular}
    \caption{(a) Tracked vehicle equipped with \textit{Boxi} for collecting the \ttt{PPS TV} dataset. (b) RESPLE2 results on the \ttt{day\_street} sequence.}
    \label{fig:tanki}
    \vspace{-4mm}
\end{figure}

The \ttt{PPS Tracked Vehicle} dataset\footnote{Available at \href{https://datasets.leggedrobotics.com/project/f44c68af-6827-4b16-bcd8-abf2d493eebb/missions}{https://datasets.leggedrobotics.com/project/f44c68af-6827-4b16-bcd8-abf2d493eebb/missions}} is recorded onboard a tracked robot equipped with the \textit{Boxi} multi-sensor rig~\cite{frey2025boxi} as shown in \figref{fig:tanki}-(a). It features high translational and angular velocities in unstructured outdoor environments under both daytime and nighttime conditions. The peak angular rate reaches \SI{4.5}{\radian/\second} in \ttt{night\_street}, and the accelerometer (\ttt{Mid360} built-in IMU) frequently saturates at its $\pm 4\,g$ full-scale limits during aggressive maneuvers. 

As summarized in Tab.~\ref{tab:tanki}, RESPLE2 achieves strong overall performance, with low APE and RPE across the dataset in both LIO and LVIO configurations. RESPLE-LO fails on \ttt{day\_open}, which becomes particularly challenging near the end as the robot climbs a hill and becomes stuck. In the LIO setting, RESPLE2 mostly achieves lower RPE and APE than FAST-LIVO2 and Coco-LIC. Incorporating visual measurements generally reduces odometry errors for RESPLE2 and FAST-LIVO2, whereas it does not improve the reported accuracy of Coco-LIC. In the LVIO setting, RESPLE2 achieves lower APE than Coco-LIC on every sequence, with equal or lower RPE on most sequences. We further show representative runs of RESPLE2 in \figref{fig:front}-(c) and \figref{fig:tanki}-(b).

\begin{table}[htbp]
    \vspace{-2mm}
    \scriptsize
    \centering
    \setlength{\tabcolsep}{2pt}
    \caption{APE and RPE (RMSE, meters) on \ttt{PPS Tracked Vehicle}.}
    \begin{tabular}{r|cccccccc|cc}
        \toprule
         & & CMLO & \multicolumn{2}{c}{CoLIC} & \multicolumn{2}{c}{FLIVO2} & \multicolumn{2}{c}{RESPLE} & \multicolumn{2}{|c}{RESPLE2}\\
        \cmidrule(lr){3-3} \cmidrule(lr){4-5} \cmidrule(lr){6-7} \cmidrule(lr){8-9} \cmidrule(lr){10-11}
         & & LO & LIO & LVIO & LIO & LVIO & LO & LIO & LIO & LVIO\\
        \midrule
        \multirow{2}{*}{\ttt{DO}} & APE & $0.082$ & $0.051$ & $0.051$ & $0.582$ & ${0.046}$ & \xmark & $0.066$ & $\underline{0.042}$ & $\bf{0.041}$\\
         & RPE & $0.079$ & $\underline{0.007}$ & $\underline{0.007}$ & $0.044$ & $\underline{0.007}$ & \xmark & $0.018$ & $\bf{0.006}$ & $\underline{0.007}$\\

         \rowcolor{gray!15} & APE & $0.063$ & ${0.026}$ & ${0.026}$ & ${0.026}$ & $0.028$ & $0.030$ & $\underline{0.024}$ & $\underline{0.024}$ & $\bf{0.020}$\\

         \rowcolor{gray!15} \multirow{-2}{*}{\ttt{DH}} & RPE & $0.060$ & $\underline{0.006}$ & $\underline{0.006}$ & $0.009$ & $0.009$ & $0.020$ & $0.012$ & $\bf{0.005}$ & $\bf{0.005}$\\
        % \cmidrule(lr){1-11}
        \multirow{2}{*}{\ttt{NH}} & APE & $0.071$ & $0.029$ & $0.029$ & ${0.024}$ & $\underline{0.023}$ & $0.034$ & $0.025$ & $0.025$ & $\bf{0.022}$\\
         & RPE & $0.070$ & ${0.006}$ & ${0.006}$ & $0.009$ & $0.008$ & $0.020$ & $0.008$ & $\underline{0.005}$ & $\bf{0.004}$\\
        % \cmidrule(lr){1-11}
        \rowcolor{gray!15} & APE & $0.077$ & $0.189$ & $0.202$ & $0.258$ & $0.058$ & $0.225$ & $\bf{0.039}$ & $\underline{0.045}$ & $0.059$\\
        \rowcolor{gray!15} \multirow{-2}{*}{\ttt{NL}} & RPE & $0.038$ & $\bf{0.006}$ & $\bf{0.006}$ & $0.010$ & $\underline{0.008}$ & $0.019$ & $0.012$ & $\bf{0.006}$ & $\bf{0.006}$\\
        % \cmidrule(lr){1-11}
        \multirow{2}{*}{\ttt{NS}} & APE & $0.057$ & $0.016$ & $0.016$ & ${0.015}$ & ${0.015}$ & $0.032$ & $0.017$ & $\underline{0.013}$ & $\bf{0.011}$\\
         & RPE & $0.043$ & $\bf{0.005}$ & $\bf{0.005}$ & $0.007$ & $0.008$ & $0.023$ & $0.011$ & $\underline{0.006}$ & $\bf{0.005}$\\
        % \cmidrule(lr){1-11}
        \rowcolor{gray!15} & APE & $0.058$ & ${0.023}$ & $0.024$ & $0.024$ & $0.024$ & $0.031$ & $0.025$ & $\underline{0.021}$ & $\bf{0.019}$\\
        \rowcolor{gray!15} \multirow{-2}{*}{\ttt{DS}} & RPE & $0.043$ & $\bf{0.005}$ & $\bf{0.005}$ & $\underline{0.006}$ & $0.007$ & $0.018$ & $0.011$ & $\bf{0.005}$ & $\bf{0.005}$\\
        \bottomrule
    \end{tabular}
    \parbox{0.97\columnwidth}{
        \vspace{4pt}
        Sequence abbreviations: \ttt{DO} (\ttt{day\_open}), \ttt{DH} (\ttt{day\_hill}), \ttt{NH} (\ttt{night\_hill}), \ttt{NL} (\ttt{night\_loop}), \ttt{NS} (\ttt{night\_street}), and \ttt{DS} (\ttt{day\_street}).
        }
    \label{tab:tanki}
\end{table}

\subsubsection{\ttt{HOPPER}}
\begin{figure}[t!]
    \centering
    \begin{tabular}{@{}c@{\hspace{4pt}}c@{\hspace{4pt}}c@{}}
        \framelabel[south west]{\adjustbox{trim={0.\width} {0.\height} {0.\width} {0.\height},clip}{\includegraphics[width=0.22\linewidth]{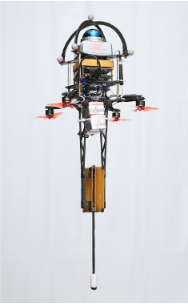}}}{\normalsize{\ttt{(a)}}}&
        \framelabel[south west]{\adjustbox{trim={0.\width} {0.\height} {0.\width} {0.\height},clip}{\includegraphics[width=0.355\linewidth]{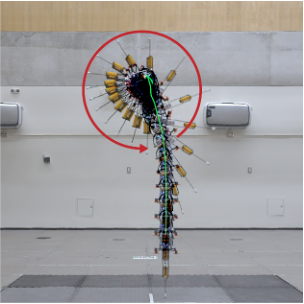}}}{\normalsize{\ttt{(b)}}}&
        \framelabel[south west]{\adjustbox{trim={0.\width} {0.\height} {0.\width} {0.\height},clip}{\includegraphics[width=0.355\linewidth]{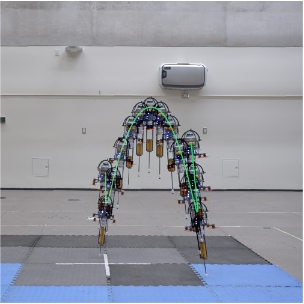}}}{\normalsize{\ttt{(c)}}}
    \end{tabular}
    \caption{Legged quadcopter (a) used to collect the \ttt{HOPPER} dataset, with examples of flip (b) and lateral jump (c) from \ttt{flip4}.}
    \label{fig:hopper}
    \vspace{-4mm}
\end{figure}

\begin{figure}[t!]
    \centering
    \adjustbox{trim={0.03\width} {0.13\height} {0.01\width} {0.01\height},clip}{\includegraphics[width=1\linewidth]{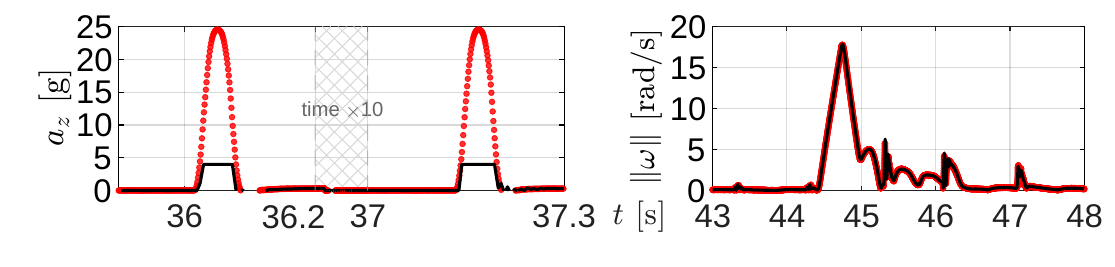}}
    \caption{Acceleration along the $z$-axis during \ttt{high2} (left) and angular velocity norms during \ttt{flip2} (right). Red and black curves correspond to the external and internal IMUs, respectively.}
    \label{fig:hoppermeas}
    \vspace{-5mm}
\end{figure}

The \ttt{HOPPER} dataset comprises 12 indoor sequences recorded onboard a legged quadcopter performing highly aggressive hybrid maneuvers, including vertical and lateral jumps, flight, and flips (\figref{fig:hopper}), in both enclosed and open spaces~\cite{li2025high,kledkruemas2026compliant}. The robot is equipped with a \ttt{Mid360}, with inertial measurements recorded by its built-in \ttt{ICM40609} (\SI{200}{\hertz}) and an external \ttt{BMI088} (range $\pm24$\,g), which operates at \SI{1000}{\hertz} and \SI{500}{\hertz} in enclosed and open spaces, respectively. The robot undergoes highly aggressive motion, with instantaneous accelerations reaching approximately $25$\,g and angular rates of up to \SI{18}{\radian/\second} (\figref{fig:hoppermeas}). These acceleration peaks exceed the $\pm4$\,g measurement range of the \ttt{Mid360} built-in IMU, highlighting a challenging test of robustness to severe inertial measurement saturation. We calibrate the extrinsic transformation between the external IMU and the LiDAR and compensate for their time offset. Both pose (\SI{300}{\hertz}) and velocity (\SI{100}{\hertz}) ground truth data are recorded using a Vicon motion capture system equipped with 10 Vantage cameras in a room of $\SI{10}{\meter}\times\SI{10}{\meter}$. The enclosed-space sequences take place within this room, while the open-space sequences (\ttt{long1} to \ttt{long3}) start and end there and cover longer distances through a larger area beyond the room.

\begin{figure}[t!]
    \centering
    \begin{tabular}{@{}c@{\hspace{4pt}}c@{}}
        \framelabel[south west]{\adjustbox{trim={0.2\width} {0.17\height} {0.2\width} {0.18\height},clip}{\includegraphics[width=0.79\linewidth]{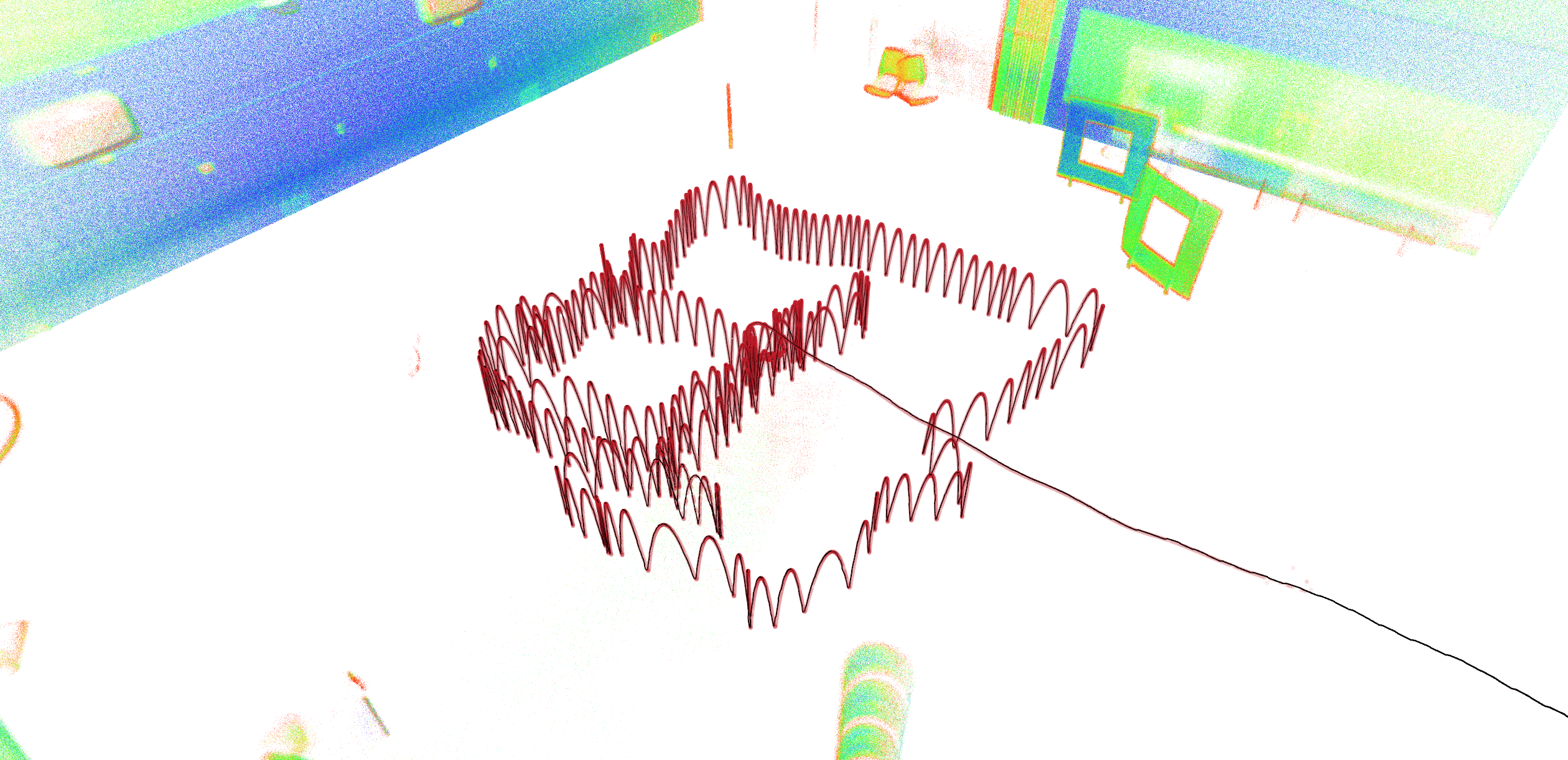}}}{{\ttt{base1}}}&
        \framelabel[south west]{\adjustbox{trim={0.2\width} {0.17\height} {0.2\width} {0.18\height},clip}{\includegraphics[width=0.79\linewidth]{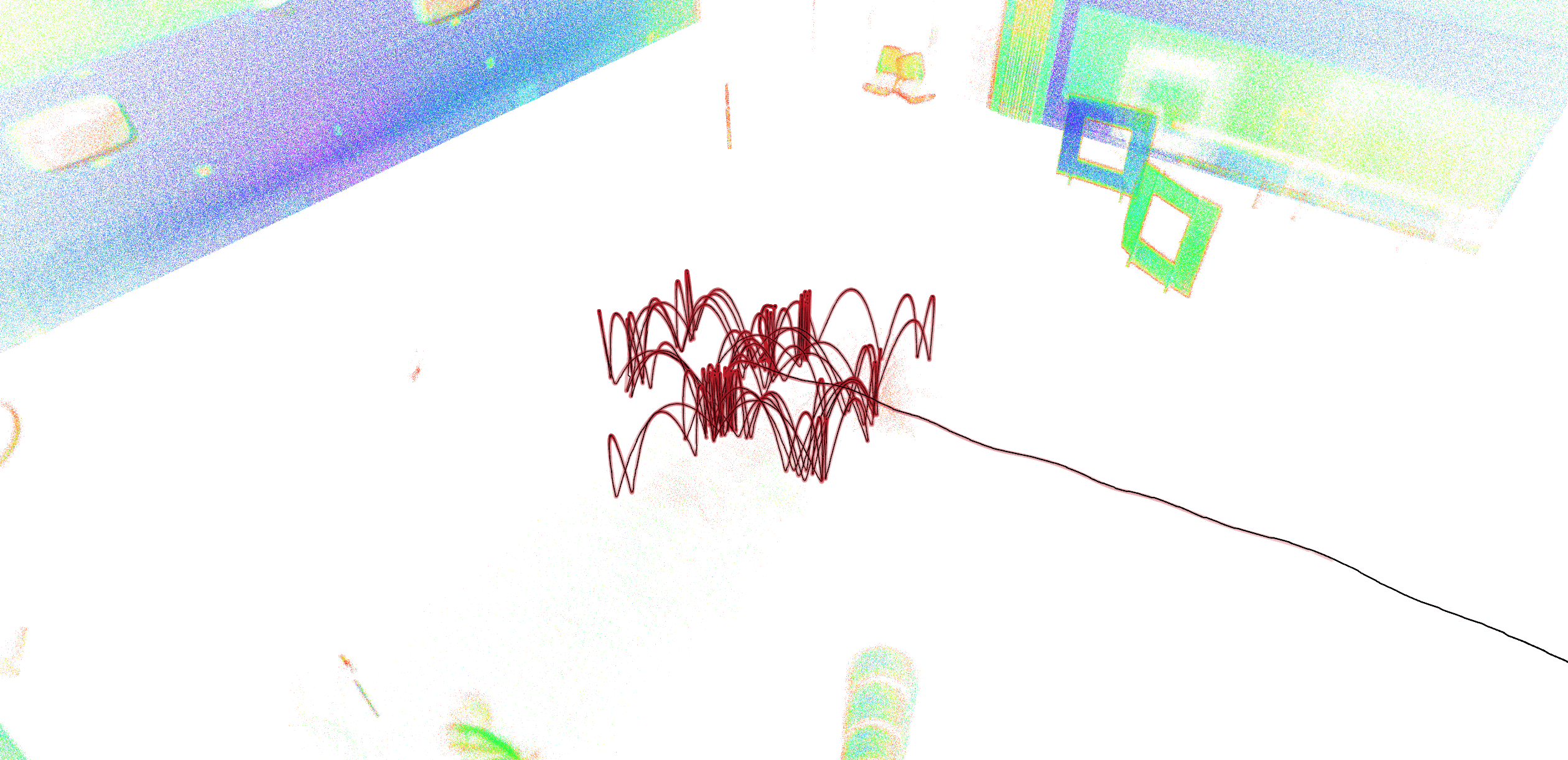}}}{{\ttt{side2}}}\\
        \framelabel[south west]{\adjustbox{trim={0.2\width} {0.17\height} {0.2\width} {0.18\height},clip}{\includegraphics[width=0.79\linewidth]{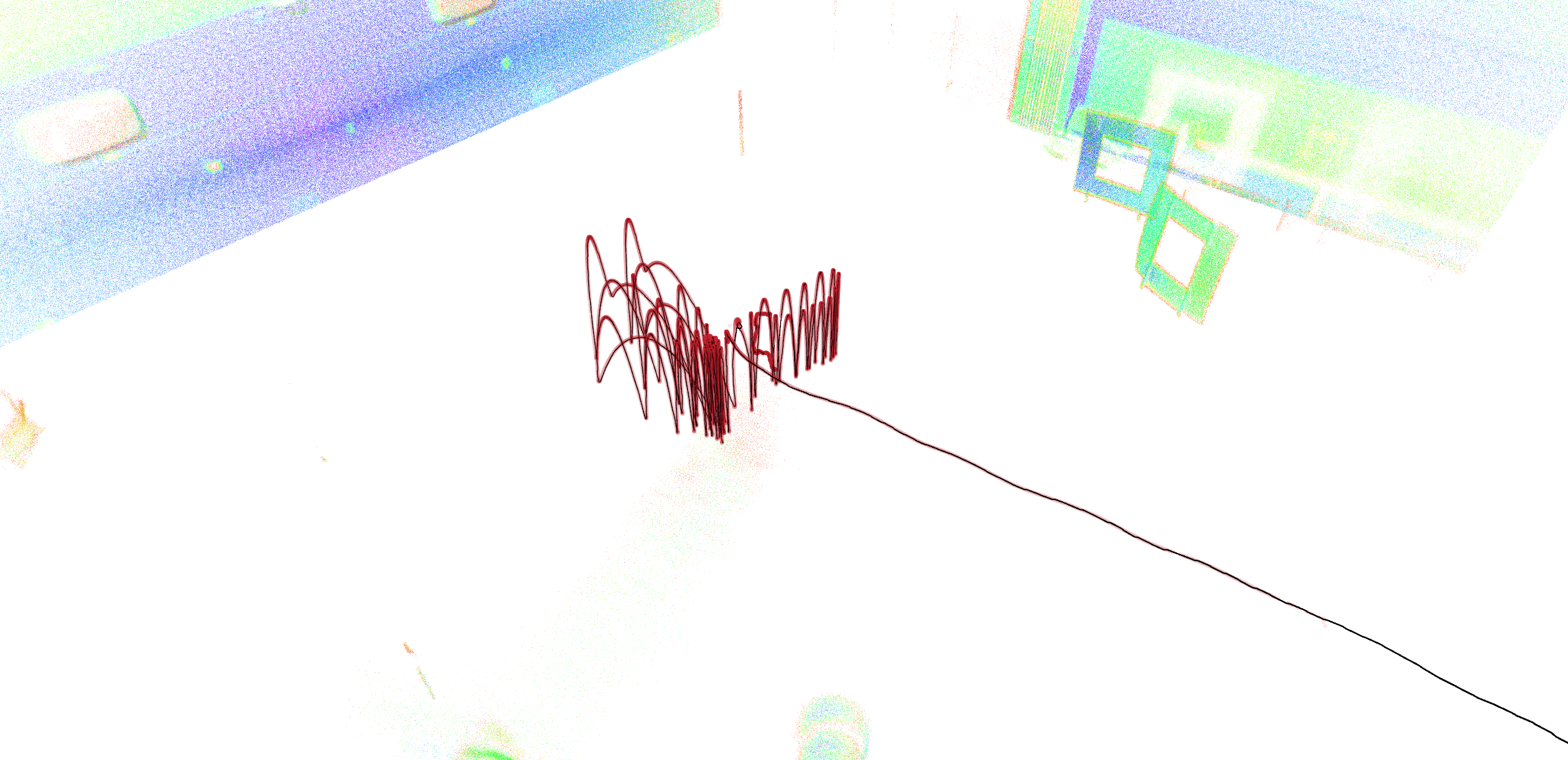}}}{{\ttt{flip2}}}&
        \framelabel[south west]{\adjustbox{trim={0.2\width} {0.17\height} {0.2\width} {0.18\height},clip}{\includegraphics[width=0.79\linewidth]{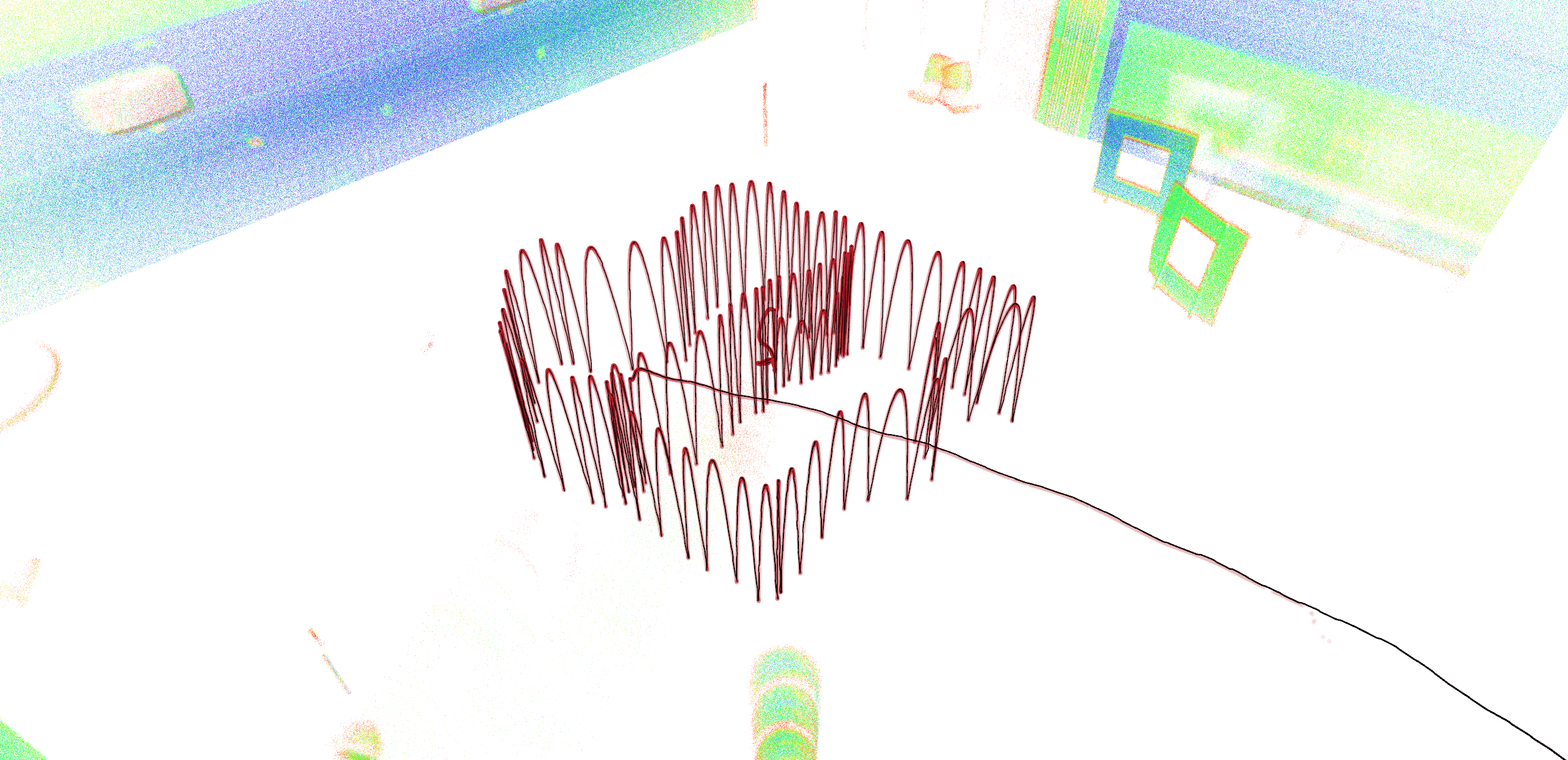}}}{{\ttt{high2}}}\\
    \end{tabular}
    \caption{RESPLE2 tested on \ttt{HOPPER}. Black and red curves represent the estimated and ground-truth trajectories, respectively.}
    \label{fig:hoppertraj}
    \vspace{-4mm}
\end{figure}

In the \ttt{flip} sequences, the robot performs flips, producing high angular velocities and abrupt changes in the LiDAR scans. In the \ttt{high} sequences, the robot jumps higher, generating extreme accelerations exceeding $24$\,g. In the \ttt{side} sequences, the robot performs large lateral jumps. The three \ttt{long} sequences evaluate performance over longer trajectories of jumps, with \ttt{long3} shown in \figref{fig:front}-(d). \figref{fig:hoppertraj} provides representative results from RESPLE2 on another four sequences.

Tab.~\ref{tab:hopperpos} reports position APE of LIO individually with the external and internal IMUs. RESPLE2 achieves the lowest APE on all sequences under both IMU configurations (except for \ttt{flip4} using internal IMU). Remarkably, its internal-IMU variant remains competitive with other methods using the external IMU, demonstrating consistent robustness to reduced IMU quality. Using the external IMU, RESPLE and Coco-LIC achieve comparable position accuracy and complete all sequences. FAST-LIVO2 is competitive on successful runs, although it fails on \ttt{flip1}. 

Coco-LIC and FAST-LIVO2 fail on all sequences with the internal IMU, highlighting the challenges posed by aggressive motion and severe inertial measurement saturation. Point-LIO completes all sequences with the internal IMU through its dedicated design for handling dynamic motion, using point-wise measurement updates and rejecting accelerometer measurements along saturated axes. However, its errors remain higher than those of RESPLE and RESPLE2, with severe deterioration on the \ttt{high} and \ttt{flip} sequences under more aggressive maneuvers when using the internal IMU. RESPLE2-LO also completes all sequences without IMU measurements, with estimation accuracy similar to or only slightly worse than that of RESPLE2 using the internal IMU. Specifically, it outperforms Point-LIO with the internal IMU on all sequences.

\begin{table}[htbp]
    \vspace{-2mm}
    \scriptsize
    \centering
    \setlength{\tabcolsep}{2.3pt} 
    \caption{APE (RMSE, meters) on \ttt{HOPPER}.}
    \begin{tabular}{r|cccccc|ccc}
        \toprule
        Method &\multicolumn{2}{c}{PLIO} &CoLIC &FLIVO2 &\multicolumn{2}{c|}{RESPLE} &\multicolumn{3}{c}{RESPLE2}\\
        \cmidrule(lr){2-3} \cmidrule(lr){4-4} \cmidrule(lr){5-5} \cmidrule(lr){6-7} \cmidrule(lr){8-10}
        Variant &\ttt{ext} &\ttt{int} &\ttt{ext} &\ttt{ext} &\ttt{ext} &\ttt{int} &\ttt{ext} &\ttt{int} & \ttt{LO}\\
        \midrule
        \ttt{base1}  &$0.022$&$0.178$&$0.008$&$\underline{0.007}$&$0.010$&$0.016$&$\bf{0.005}$&$0.014$&$0.017$\\
        \rowcolor{gray!15} \ttt{high1}  &$0.022$&$0.556$&$0.020$&$0.019$&$\underline{0.011}$&$0.019$&$\bf{0.005}$&$0.017$&$0.021$\\
        \ttt{high2}  &$0.022$&$0.594$&$0.017$&$0.021$&$\underline{0.011}$&$0.023$&$\bf{0.005}$&$0.020$&$0.024$\\
        \rowcolor{gray!15} \ttt{side1}  &$0.019$&$0.178$&$\underline{0.008}$&$\underline{0.008}$&$\underline{0.008}$&$0.018$&$\bf{0.004}$&$0.016$&$0.017$\\
        \ttt{side2}  &$0.019$&$0.149$&$\underline{0.008}$&$\underline{0.008}$&$\underline{0.008}$&$0.017$&$\bf{0.004}$&$0.015$&$0.015$\\
        \rowcolor{gray!15} \ttt{flip1}  &$0.024$&$0.335$&$0.038$&\xmark&$\underline{0.014}$&$0.019$&$\bf{0.008}$&$0.018$&$0.022$\\
        \ttt{flip2}  &$0.026$&$0.314$&$0.015$&$\underline{0.012}$&$\underline{0.012}$&$0.018$&$\bf{0.008}$&$0.018$&$0.019$\\
        \rowcolor{gray!15} \ttt{flip3}  &$0.028$&$0.361$&$0.024$&$0.024$&$\underline{0.013}$&$0.020$&$\bf{0.011}$&$0.019$&$0.020$\\
        \ttt{flip4}  &$0.042$&$0.407$&$0.017$&$0.016$&$\underline{0.013}$&$0.019$&$\bf{0.009}$&$0.020$&$0.027$\\
        \rowcolor{gray!15} \ttt{long1}  &$0.025$&$0.102$&$\underline{0.013}$&$0.017$&$\underline{0.013}$&$0.019$&$\bf{0.010}$&$0.015$&$0.022$\\
        \ttt{long2}  &$0.033$&$0.087$&$\underline{0.014}$&$\bf{0.013}$&$0.016$&$0.020$&$\bf{0.013}$&$0.018$&$0.019$\\
        \rowcolor{gray!15} \ttt{long3}  &$0.034$&$0.133$&$0.017$&$\underline{0.016}$&$0.018$&$0.022$&$\bf{0.014}$&$0.020$&$0.021$\\
        \bottomrule
    \end{tabular}
    \vspace{-2mm}
    \label{tab:hopperpos}
\end{table}

We additionally evaluate velocity estimation recorded within enclosed spaces using the instantaneous estimates produced after each measurement update, reflecting their use in online control tasks~\cite{li2025high}. As shown in Tab.~\ref{tab:hoppervel}, all systems exhibit greater sensitivity to IMU quality in velocity estimation than in position estimation. With the external IMU, RESPLE2 achieves the lowest or the second lowest velocity estimation error on all sequences. With the internal IMU, it achieves higher accuracy than all other systems on all sequences under the same IMU configuration. Notably, due to its delayed knot determination, Coco-LIC introduces an additional latency of up to two optimization cycles (\SI{200}{\milli\second}), which may degrade online control performance~\cite{quenzel2025lio}.

\begin{table}[htbp]
	\vspace{-2mm}
    \scriptsize
    \centering
    \setlength{\tabcolsep}{2.3pt} 
    \caption{Velocity estimation error (RMSE, \SI{}{\meter/\second}) on \ttt{HOPPER}.}
        \begin{tabular}{r|cccccc|ccc}
            \toprule
            Method &\multicolumn{2}{c}{PLIO} &CoLIC &FLIVO2 &\multicolumn{2}{c|}{RESPLE} &\multicolumn{3}{c}{RESPLE2}\\
            \cmidrule(lr){2-3} \cmidrule(lr){4-4} \cmidrule(lr){5-5} \cmidrule(lr){6-7} \cmidrule(lr){8-10}
            IMU &\ttt{ext} &\ttt{int} &\ttt{ext} &\ttt{ext} &\ttt{ext} &\ttt{int} &\ttt{ext} &\ttt{int} & \ttt{LO}\\
            \midrule
            \ttt{base1}  &$0.245$ &$1.674$ &$\bf{0.081}$ &$0.254$ &$0.732$ &$0.798$ &$\underline{0.111}$ &$0.617$ &$1.019$\\
            \rowcolor{gray!15} \ttt{high1}  &$0.294$ &$2.098$ &$\underline{0.123}$ &$0.269$ &$0.683$ &$0.775$ &$\bf{0.097}$ &$0.678$ &$1.079$\\
            \ttt{high2}  &$0.261$ &$2.509$ &$\underline{0.125}$ &$0.353$ &$0.734$ &$0.924$ &$\bf{0.105}$ &$0.809$ &$1.264$\\
            \rowcolor{gray!15} \ttt{side1}  &$0.256$ &$1.836$ &$\bf{0.077}$ &$0.236$ &$0.582$ &$0.756$ &$\underline{0.107}$ &$0.591$ &$0.993$\\
            \ttt{side2}  &$\bf{0.123}$ &$1.757$ &$0.218$ &$0.290$ &$0.705$ &$0.802$ &$\underline{0.210}$ &$0.675$ &$1.068$\\
            \rowcolor{gray!15} \ttt{flip1}  &$\underline{0.252}$ &$2.013$ &$0.288$ &\xmark &$0.789$ &$0.843$ &$\bf{0.227}$ &$0.749$ &$1.176$\\
            \ttt{flip2}  &$0.300$ &$2.000$ &$\underline{0.258}$ &$0.317$ &$0.798$ &$0.881$ &$\bf{0.256}$ &$0.742$ &$1.155$\\
            \rowcolor{gray!15} \ttt{flip3}  &$\bf{0.231}$ &$1.837$ &$0.269$ &$0.332$ &$0.759$ &$0.859$ &$\underline{0.255}$ &$0.708$ &$1.139$\\
            \ttt{flip4}  &$0.316$ &$2.162$ &$\underline{0.204}$ &$0.266$ &$0.707$ &$0.752$ &$\bf{0.166}$ &$0.668$ &$1.245$\\
            \bottomrule
        \end{tabular}
	\vspace{-2mm}
    \label{tab:hoppervel}
\end{table}

\begin{figure}[t!]
    \centering
    \adjustbox{trim={0.03\width} {0.05\height} {0.04\width} {0.05\height},clip}{\includegraphics[width=1.05\linewidth]{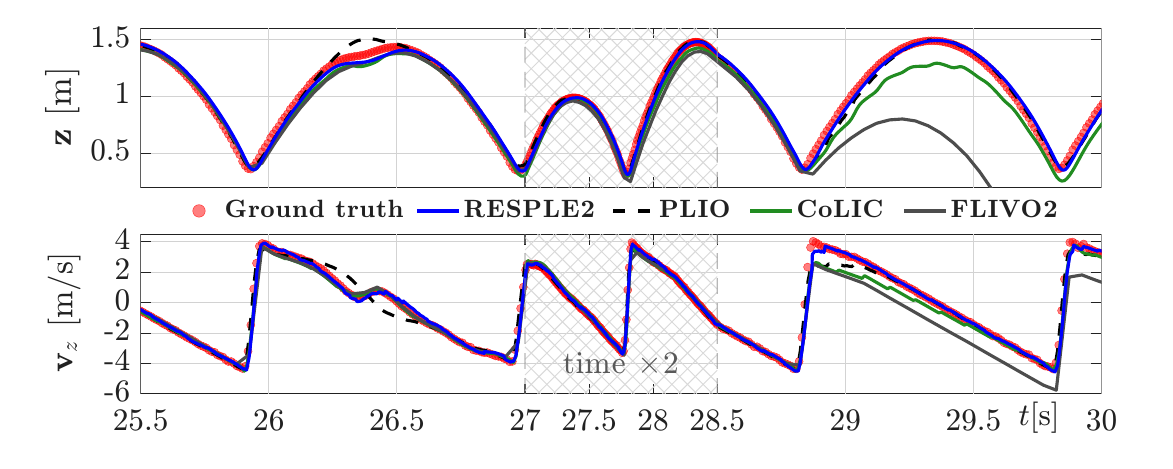}}\\
    \caption{Position (top) and velocity (bottom) along the $z$-axis on \ttt{flip1}.}
    \label{fig:hoppercmp}
    \vspace{-4mm}
\end{figure}

As depicted in \figref{fig:hoppercmp}, we further compare the position and velocity estimates of the systems against the ground truth on \ttt{flip1}, including a flip maneuver during \SI{26}{}-\SI{27}{\second}. After approximately \SI{29}{\second}, FAST-LIVO2 diverges, and Coco-LIC and Point-LIO deviate substantially from the ground truth, whereas RESPLE2 closely tracks it despite the dynamic maneuver.

\subsubsection{\ttt{HANDI}} 
The \ttt{HANDI} dataset comprises $13$ sequences recorded in diverse indoor and outdoor environments using different LVI configurations on a handheld platform. 
% based on an open-source design\footnote{Available at \url{https://asig-x.github.io/resple_demonstrator/}}
The \ttt{az} sequences use \ttt{Avia} and \ttt{ZedX1}, \ttt{ar-inout} uses \ttt{Avia} and \ttt{D455}, and \ttt{mz-inout} uses \ttt{Mid360} and \ttt{ZedX1}. The \ttt{az-dyna} sequences capture dynamic motion during jumping, running, rapid lateral rotations, and circular arm movements within a \SI{12}{\meter}$\times$\SI{12}{\meter} space. The \ttt{inout} and \ttt{az-corri} sequences involve long-distance travel, including scenarios with geometric degeneracy (e.g., in long corridors), as well as transitions between indoor and outdoor environments. The \ttt{az-out} sequences are recorded entirely outdoors, including in wilderness environments, and return to their starting locations, enabling end-to-end error evaluation. For all other sequences, a Qualisys camera tracking system provides ground truth for either the entire trajectory (\ttt{az-dyna}) or the initial and final portions of longer trajectories (rest). We enable the proposed global refinement module at loop closures for the long-distance sequences.  

\begin{figure}[t!]
    \centering
    \vspace{1mm}
    \begin{tabular}{@{}c@{\hspace{4pt}}c@{}}
        \framelabel[south west]{\adjustbox{trim={0.2\width} {0.1\height} {0.2\width} {0.18\height},clip}{\includegraphics[width=0.8\linewidth]{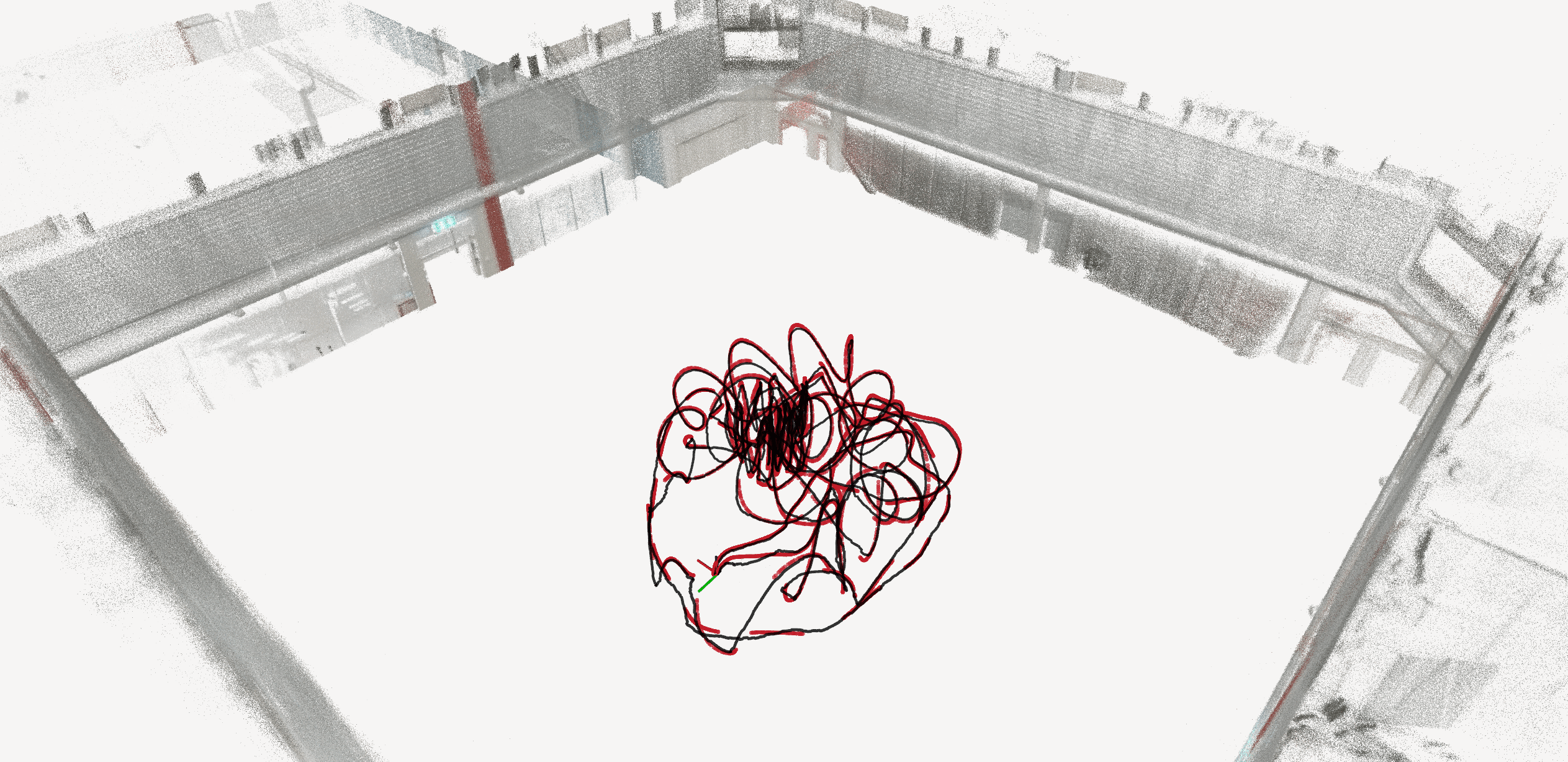}}}{{\ttt{az-dyna1}}}&
        \framelabel[south west]{\adjustbox{trim={0.2\width} {0.1\height} {0.2\width} {0.18\height},clip}{\includegraphics[width=0.8\linewidth]{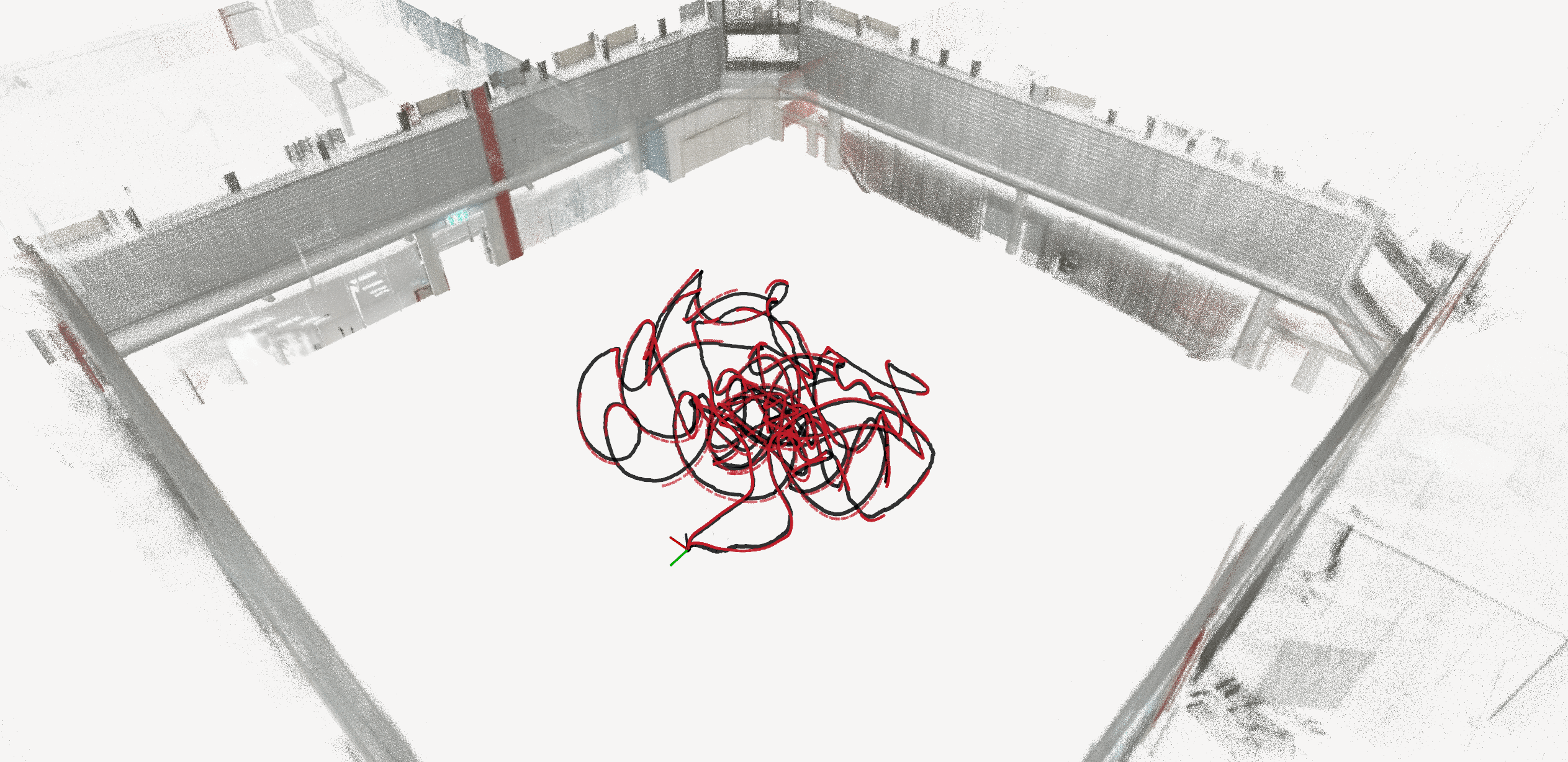}}}{{\ttt{az-dyna2}}}\\
        \framelabel[south west]{\adjustbox{trim={0.2\width} {0.1\height} {0.2\width} {0.18\height},clip}{\includegraphics[width=0.8\linewidth]{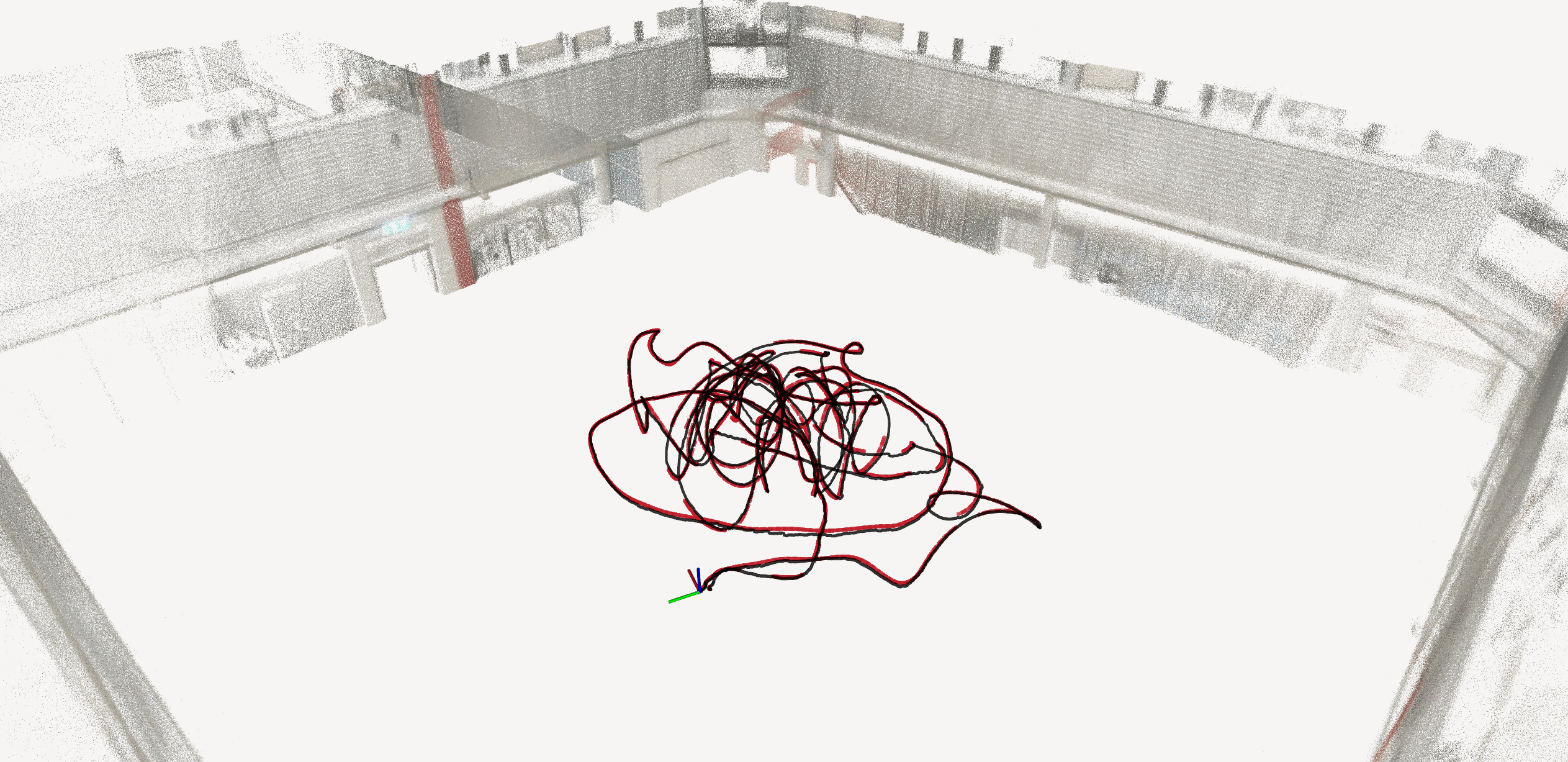}}}{{\ttt{az-dyna4}}}&
        \framelabel[south west]{\adjustbox{trim={0.2\width} {0.1\height} {0.2\width} {0.18\height},clip}{\includegraphics[width=0.8\linewidth]{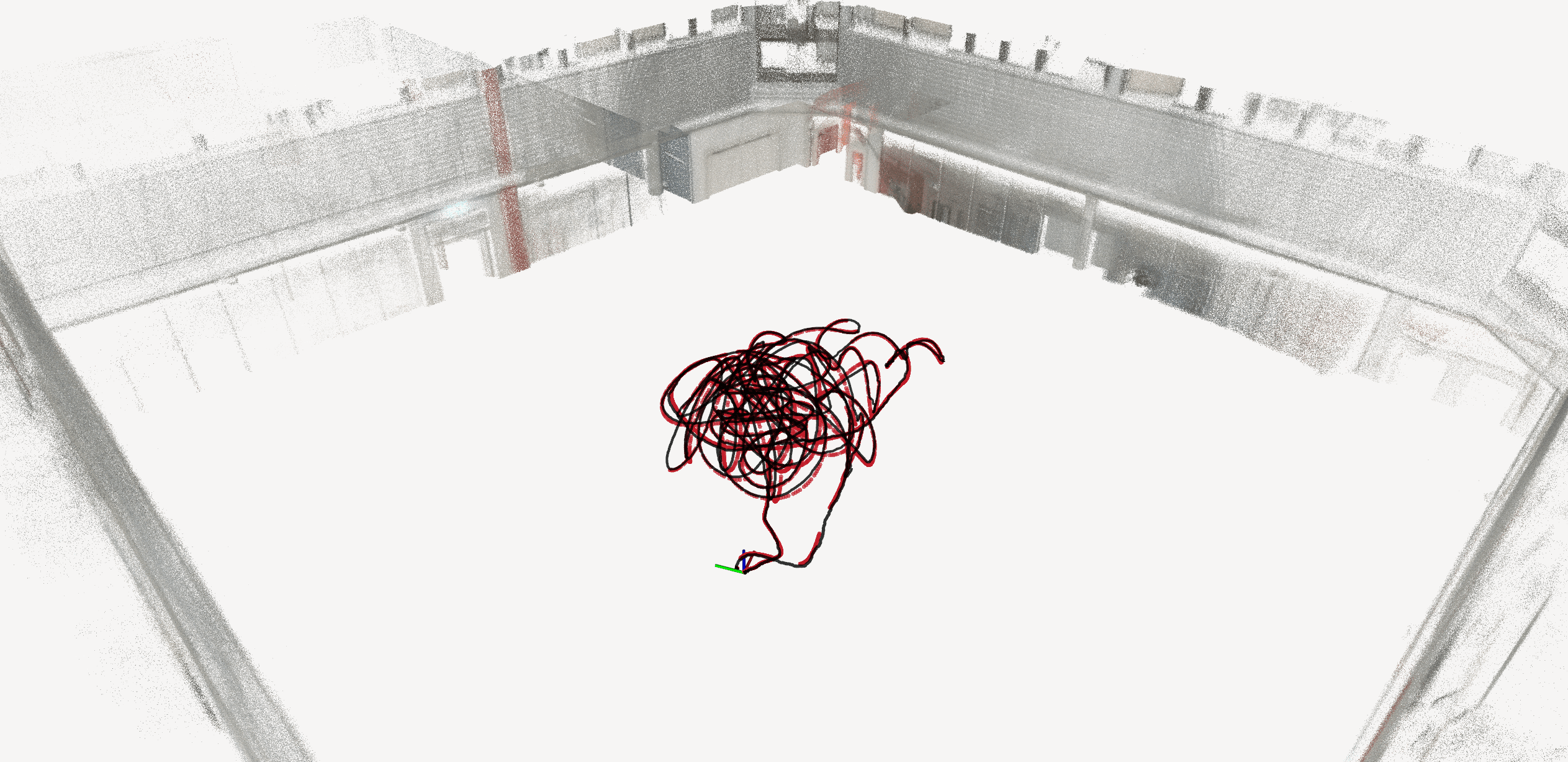}}}{{\ttt{az-dyna5}}}\\
    \end{tabular}
    \caption{RESPLE2 on \ttt{az-dyna} sequences. Black and red curves are estimated and ground-truth trajectories, respectively.}
    \label{fig:handi}
    \vspace{-2mm}
\end{figure}

As summarized in Tab.~\ref{tab:handi}, RESPLE2 completes all evaluated sequences for both LIO and LVIO, with at least one variant achieving the lowest odometry error on every sequence. RESPLE2-LVIO outperforms all competing systems in geometrically degenerate corridors and during indoor-outdoor transitions, and achieves the lowest end-to-end errors on the outdoor sequences. In general, incorporating visual measurements substantially improves the odometry accuracy in both dynamic and long-distance scenarios. Global refinement of the LVIO trajectories further reduces errors on long-distance sequences, with particularly pronounced gains in unstructured or geometrically degenerate environments, e.g., \ttt{az-corri} and \ttt{az-out2}. \figref{fig:handi} further illustrates representative runs of RESPLE2 on the \ttt{az-dyna} sequences. The hardware setups and mapping results of \ttt{ar-inout}, \ttt{az-inout}, and \ttt{mz-inout} are shown in \figref{fig:front}-(f), (g), and (h), respectively.

\begin{figure}[t!]
    \centering
    \framelabel[south west]{\adjustbox{trim={0.\width} {0.\height} {0.\width} {0.\height},clip}{\includegraphics[width=0.97\linewidth]{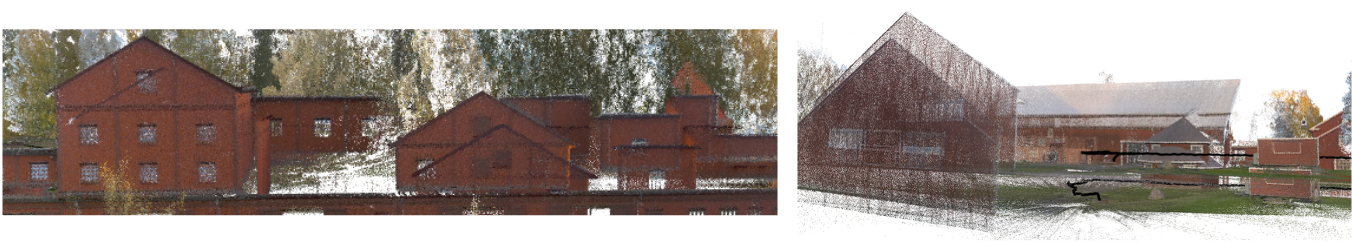}}}{Before}
    \framelabel[south west]{\adjustbox{trim={0.\width} {0.\height} {0.\width} {0.\height},clip}{\includegraphics[width=0.97\linewidth]{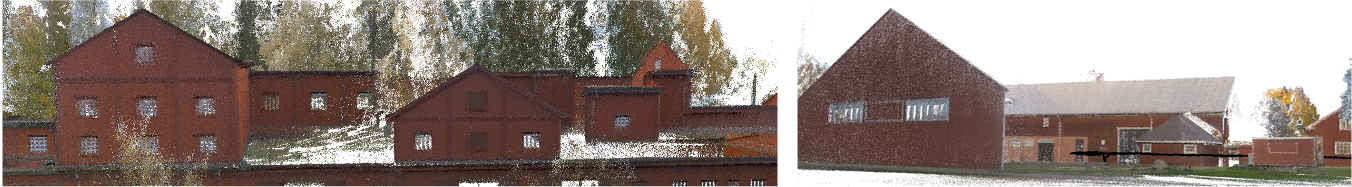}}}{After}
    \caption{Comparison before and after global refinement on \ttt{az-out2}.}
    \label{fig:lcs}
    \vspace{-2mm}
\end{figure}

\begin{figure}[t!]
    \vspace{1mm}
    \centering
    \adjustbox{trim={0.06\width} {0.1\height} {0.1\width} {0.1\height}}{\includegraphics[width=1\linewidth]{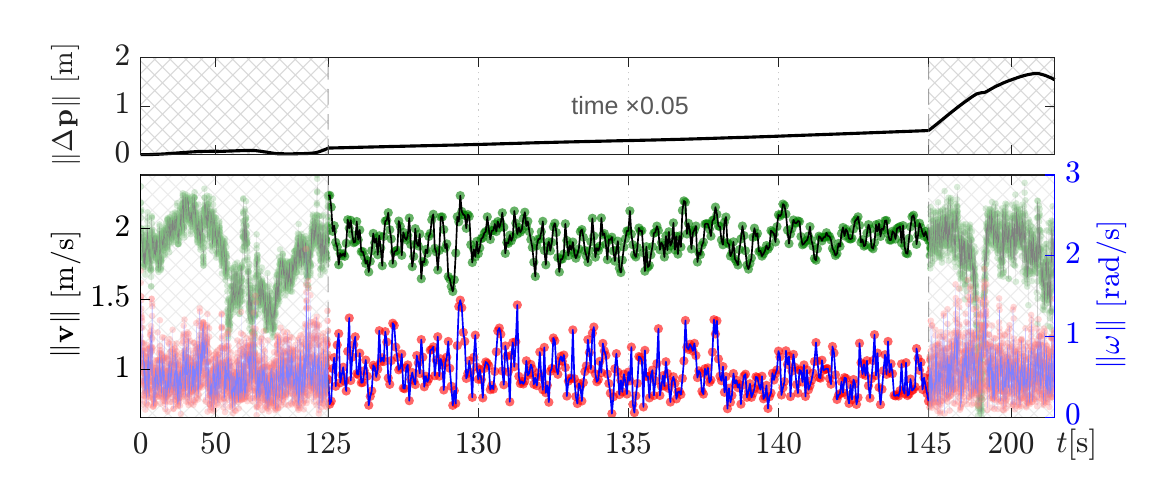}}    \caption{Position correction and preservation of body-frame angular (blue) and linear velocities (black) at the first global spline optimization.}
    \label{fig:globalVel}
    \vspace{-2mm}
\end{figure}

\figref{fig:lcs} further provides a qualitative comparison of mapping consistency before and after global refinement. \figref{fig:globalVel} shows the quantitative results of global refinement at the first detected loop closure in \ttt{ar-inout}. The proposed global spline optimization reduces position errors along the trajectory by up to nearly \SI{2}{\meter} while consistently preserving higher-order continuous-time states such as angular and linear velocities.    

\begin{table}[htbp]       
    \vspace{-2mm}
    \centering
    \scriptsize
    \setlength{\tabcolsep}{2.3pt} 
    \caption{APE (RMSE, meters) on \ttt{HANDI}.}
    \begin{tabular}{@{}r|ccc|cc@{}}
        \toprule
        Method & CoLIC & FLIVO2 & RESPLE &\multicolumn{2}{c}{RESPLE2} \\
        \cmidrule(lr){2-2} \cmidrule(lr){3-3} \cmidrule(lr){4-4} \cmidrule(lr){5-6} 
        Variant &LVIO &LVIO &LIO &LIO& LVIO (+LC) \\
        \midrule
        \ttt{az-base1} & $0.121$ & $0.059$ & $\underline{0.034}$ & ${0.036}$ & $\bf{0.028}$ (\ttt{N/A}) \\
        \rowcolor{gray!15} 
        \ttt{az-base2} & \xmark & \xmark & $0.085$ & $\underline{0.036}$ & $\bf{0.033}$ (\ttt{N/A}) \\ 
        \ttt{az-dyna1} & $0.105$ & ${0.080}$ & $0.157$ & $\bf{0.055}$ & $\underline{0.059}$ (\ttt{N/A}) \\
        \rowcolor{gray!15} 
        \ttt{az-dyna2} & $0.074$ & $0.074$ & $1.588$ & $\bf{0.053}$ & $\underline{0.054}$ (\ttt{N/A}) \\
        \ttt{az-dyna3} & $0.141$ & $\underline{0.095}$ & $4.489$ & $15.944$ & $\bf{0.059}$ (\ttt{N/A}) \\
        \rowcolor{gray!15} 
        \ttt{az-dyna4} & \xmark & $0.097$ & $0.045$ & $\underline{0.044}$ & $\bf{0.036}$ (\ttt{N/A}) \\ 
        \ttt{az-dyna5} & $0.082$ & $0.067$ & $0.069$ & $\underline{0.056}$ & $\bf{0.042}$ (\ttt{N/A}) \\
        \midrule
        \rowcolor{gray!15} 
        \ttt{az-corri} & $25.236$ & $46.476$ & $44.057$ & $6.166$ & $\underline{1.339}$ ($\bf{0.044}$) \\
        \ttt{az-inout} & $2.562$ & $5.122$ & $11.330$  & $1.334$ & $\underline{0.031}$ ($\bf{0.030}$) \\ 
        \rowcolor{gray!15} 
        \ttt{mz-inout} & $2.476$ & \xmark & $36.974$ & $1.491$ & $\underline{0.668}$ ($\bf{0.026}$) \\ 
        \ttt{ar-inout} & ${0.808}$ & \xmark & $1.122$ & $10.177$ & $\underline{0.394}$ ($\bf{0.076}$) \\ 
        \rowcolor{gray!15} 
        \ttt{az-out1}$^{*}$ & $1.686$ & $\underline{0.057}$ & $9.605$ & $4.780$ & $\bf{0.009}$ ($\bf{0.009}$) \\
        \ttt{az-out2}$^{*}$ & ${3.623}$ & $25.235$ & \xmark & $5.355$ & $\underline{2.767}$ ($\bf{0.009}$) \\
        \bottomrule
    \end{tabular}
    \parbox{0.97\columnwidth}{
    \vspace{4pt}
    $^*$Sequences start and end at the same location, for which end-to-end error is reported. For the long-distance sequences below the divider, RESPLE2 applies loop closure detection and global refinement, with resulting errors shown in parentheses.
    }
    \label{tab:handi}
\end{table}   

\section{Further Discussions} \label{sec:discussions}

\subsection{Ablations}
\subsubsection{Inertial-Informed Prediction}\label{subsubsec:ablatePred}
To assess the proposed inertial-informed control point prediction, we compare our method with a baseline in which each new RCP is predicted using a constant transition matrix $\fA$ that copies the most recent position and orientation-increment RCPs, with carefully tuned process noise. This baseline resembles the prediction scheme in RESPLE~\cite{cao2025resple} or the initialization scheme in Coco-LIC~\cite{lang2023coco}. Both settings apply the same knot-placement criteria based on \Cref{subsec:prediction}, ensuring identical knot intervals and measurement batches at each update step. This isolates the effect of the prediction strategy from that of adaptive knot placement. We use the \ttt{HOPPER} sequences recorded in enclosed spaces for testing under aggressive motion.

As shown in Tab.~\ref{tab:predrcp}, the proposed inertial-informed prediction reduces the position APE on all the sequences with relative reductions of approximately $39\%$-$76\%$. The largest reduction occurs on \ttt{high1}, which features extremely high accelerations. These consistent improvements support the benefit of incorporating current inertial measurements into control point prediction during highly dynamic motion.

\begin{figure}[t!]
    \centering
    \adjustbox{trim={0.04\width} {0.06\height} {0.07\width} {0.04\height},clip}{\includegraphics[width=1.01\linewidth]{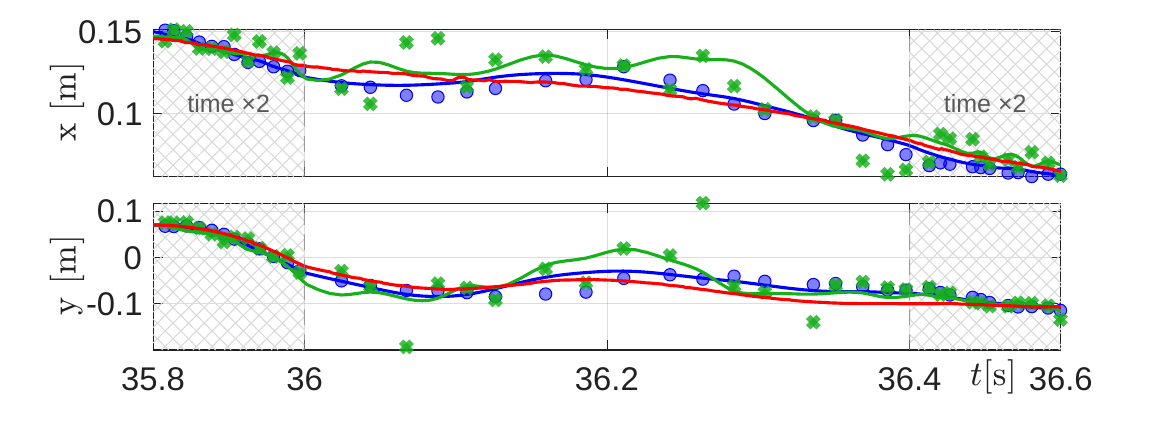}}
    \caption{Ablation of exact RCP prediction on \ttt{flip1} in $(x,y)$-coordinates w.r.t.\, ground truth (red). The IMU-predicted control points and those predicted using the constant transition matrix $\fA$ in \Cref{subsubsec:ablatePred} are shown in blue and green with the corresponding B-spline estimates, respectively.}
    \vspace{-2mm}
    \label{fig:predrcp}
\end{figure}

\begin{table}[htbp]       
    \vspace{-2mm}
    \centering
    \scriptsize
    \setlength{\tabcolsep}{0.9pt}
    \caption{Position APE (RMSE, meters) on \ttt{HOPPER} for ablation on inertial-informed prediction.}
    \begin{tabular}{@{}l*{9}{c}@{}}
        \toprule
        & \ttt{base1} & \ttt{high1} & \ttt{high2} & \ttt{side1} & \ttt{side2} & \ttt{flip1} & \ttt{flip2} & \ttt{flip3} & \ttt{flip4}\\
        \midrule
        Pred. w/o IMU  & $0.015$  & $0.017$ & $0.012$ & $0.010$ & $0.011$ & $0.015$ & $0.016$ & $0.018$ & $0.017$\\
        Pred. w/ IMU & $\bf{0.005}$ & $\bf{0.004}$ & $\bf{0.005}$ & $\bf{0.004}$ & $\bf{0.004}$ & $\bf{0.008}$ & $\bf{0.008}$ & $\bf{0.011}$ & $\bf{0.009}$\\
        \bottomrule
    \end{tabular}
    \label{tab:predrcp}
\end{table}

\begin{figure}[t!]
    \vspace{2mm}
    \begin{tabular}{@{}>{\raggedright\arraybackslash}p{\linewidth}@{}}
    \centering\arraybackslash\framelabel[north]{\adjustbox{trim={0.06\width} {0.13\height} {0.01\width} {0.\height},clip}{\includegraphics[width=1.05\linewidth]{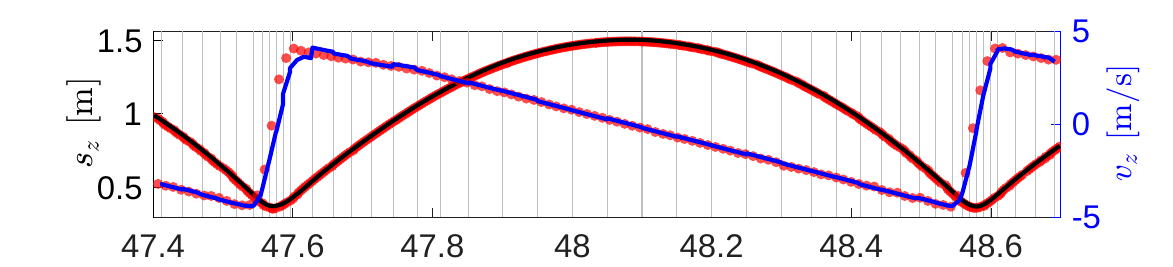}}}{RESPLE2}\\
    \centering\arraybackslash\framelabel[north]{\adjustbox{trim={0.06\width} {0.\height} {0.01\width} {0.\height},clip}{\includegraphics[width=1.05\linewidth]{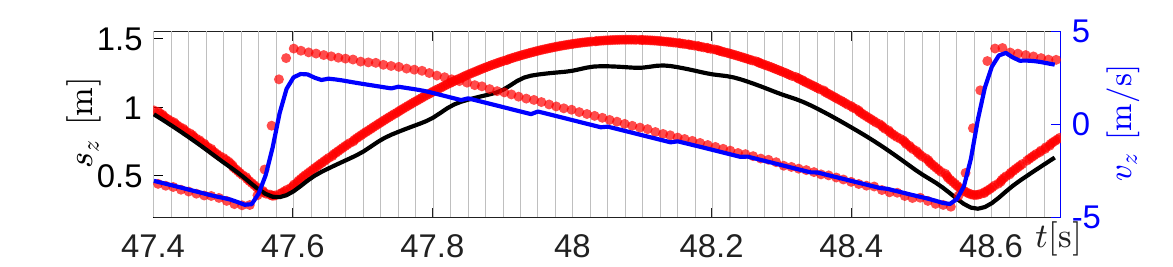}}}{Coco-LIC}\\
    \scriptsize{(a) Position (black) and velocity (blue) estimates over placed knots (gray) compared with ground truth (red). } \\ 
    
    \centering\arraybackslash\framelabel[north]{\adjustbox{trim={0.06\width} {0.13\height} {0.01\width} {0.\height},clip}{\includegraphics[width=1.05\linewidth]{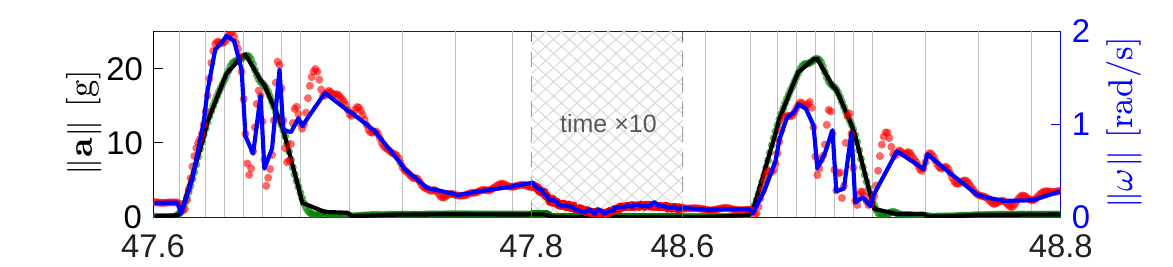}}}{RESPLE2}\\
    \centering\arraybackslash\framelabel[north]{\adjustbox{trim={0.06\width} {0.\height} {0.01\width} {0.\height},clip}{\includegraphics[width=1.05\linewidth]{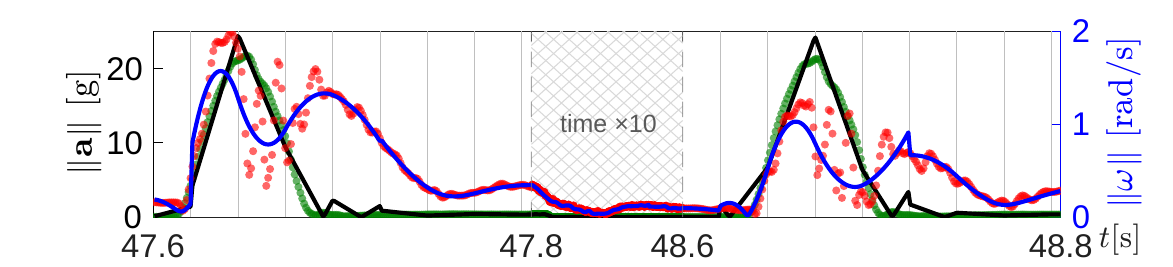}}}{Coco-LIC}\\
    \scriptsize{(b) Norms of acceleration (black) and angular velocity (blue) estimates over placed knots (gray) w.r.t.\ the accelerometer (green) and gyroscope (red) readings, respectively. } \\ 
    \end{tabular}
    \caption{Comparison of knot placement and motion estimates between RESPLE2 and Coco-LIC on a segment of \ttt{flip1}, with reference to ground truth (a) and IMU measurements (b). RESPLE2 achieves higher estimation accuracy and fidelity with a knot distribution that better adapts to motion.}
    \label{fig:abl_knot}
\end{figure}

\figref{fig:predrcp} further illustrates the corresponding position control point predictions on a segment of \ttt{flip1}. The IMU-predicted RCPs generally lie closer to the subsequently estimated trajectory, whereas the baseline predictions exhibit larger deviations. Under knot clamping, the newest control point coincides with the trajectory endpoint, thereby enabling improved control point initialization through direct IMU integration and providing a better endpoint prior for the subsequent iterated measurement update, including data association.

\subsubsection{Knot Placement}
To examine how knot placement adapts to dynamic motion, we compare RESPLE2 and Coco-LIC on a segment of \ttt{flip1} containing two landing impacts separated by a ballistic phase as shown in \figref{fig:abl_knot}-(a). RESPLE2 concentrates knots around the ground impacts, where velocity changes abruptly, and places sparser knots during the smoother ballistic motion, with the vertical position and velocity estimates closely following the ground truth throughout the segment. Coco-LIC exhibits a nearly uniform knot density over the same interval, with larger position and velocity deviations.

In Coco-LIC's design, the number of control points in each \SI{0.1}{\second} interval is determined by thresholding the acceleration and angular velocity in the global frame averaged over that interval. We observe that this averaging can reduce the contribution of brief, aggressive motion transitions, such as those during a landing impact lasting about \SI{50}{\milli\second}, limiting the sensitivity of knot placement to these transient events. Instead, RESPLE2 evaluates changes in position, orientation, and velocity at each IMU propagation substep, together with criteria on interval duration and measurement point counts. In particular, the velocity-change criterion allows rapid changes in translational motion to trigger knot extension without waiting for a fixed averaging interval.

As shown in \figref{fig:abl_knot}-(b), we further examine the consistency of inertial estimation within the same phase. RESPLE2 produces acceleration and angular-velocity estimates that are more closely aligned with the measurements, particularly during pronounced variations near the peaks, whereas Coco-LIC overshoots the acceleration peaks and underestimates the angular-velocity peaks around both ground impacts. These observations further support our motion-adaptive knot placement strategy for estimating dynamic motion.

Furthermore, over the entire \ttt{flip1} sequence, RESPLE2 uses $2087$ control points compared with $2362$ for Coco-LIC, a reduction of $11.6\%$, while achieving higher estimation accuracy. This also demonstrates the superior representation efficiency of our motion-adaptive knot placement.

\begin{figure}[t!]
    \vspace{2mm}
    \centering
    \adjustbox{trim={0.\width} {0.18\height} {0.025\width} {0.2\height},clip}{\includegraphics[width=1.05\linewidth]{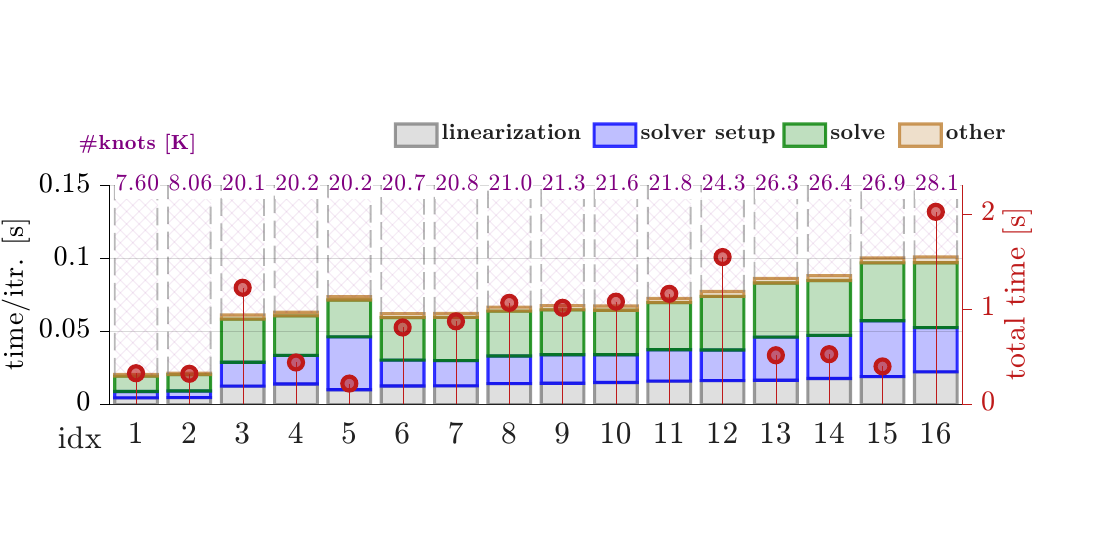}}
    \caption{Runtime profile of global spline optimization on \ttt{ar-inout}, including per-iteration cost, total optimization time, and knot counts.}
    \label{fig:handiglobalt}
    \vspace{-2mm}
\end{figure}  

\subsection{Runtime and Computational Analysis}	
Tab.~\ref{tab:runtime} reports the per-frame runtime breakdown on three sequences, one for each operating mode: \ttt{flip1} (LIO), \ttt{day\_open} (LVIO), and \ttt{Outdoor1} (LKIO). In all three modes, RESPLE2 runs well within real time, consuming only $15.4\%$, $18.3\%$, and $10.6\%$ of the available time budget per LiDAR frame, respectively. The IEKF update is the dominant stage throughout, accounting for roughly one third of the per-frame runtime in LIO, one quarter in LVIO, and about half in LKIO, followed by the image preprocessing, VoxelMap maintenance and LiDAR preprocessing. The proposed RCP prediction and reset are lightweight, together requiring at most \SI{0.51}{ms} per frame and less than $7\%$ of the total runtime on every sequence. The LVIO mode incurs the highest cost, as visual processing adds about \SI{4.1}{\milli\second} per frame and the visual measurements further enlarge the IEKF update. 

\begin{table}[htbp]
    \scriptsize
    \centering
    \caption{Runtime (\si{ms} per LiDAR frame) analysis.}
    \setlength{\tabcolsep}{2.3pt} 
    \begin{tabular}{lccc}
        \toprule
        Sequence & \ttt{flip1} & \ttt{day\_open} &\ttt{Outdoor1} \\
        \midrule
        LiDAR preprocessing & $1.570$ & $2.765$ & $0.810$ \\
        \rowcolor{gray!15} Image preprocessing & -- & $3.346$ & -- \\
        Visual feature generation & -- & $0.756$ & -- \\
        \rowcolor{gray!15} Knot extension & $0.372$ & $2.926$ & $0.380$ \\
        RCP prediction & $0.363$ & $0.184$ & $0.107$ \\
        \rowcolor{gray!15} RCP reset & $0.142$ & $0.163$ & $0.090$ \\
        Iterated update & $2.664$ & $4.883$ & $2.644$ \\
        \rowcolor{gray!15} VoxelMap maintenance & $2.291$ & $2.934$ & $1.013$ \\
        Others & $0.290$ & $0.388$ & $0.264$ \\
        \midrule
        \textbf{Total / Available} & $\bf7.7$/$\bf50$ & $\bf18.3$/$\bf100$ & $\bf5.3$/$\bf50$ \\
        \bottomrule
    \end{tabular}  
    \label{tab:runtime}
\end{table}

\figref{fig:handiglobalt} reports the runtime of the main components of the 16 global optimizations triggered on \ttt{ar-inout}, with optimization windows ranging from \SI{243}{\second} to \SI{853}{\second}. Each linearization evaluates the residuals and Jacobians and accumulates the linear system. Solver setup merges these contributions, assembles the sparse system matrix, and scales the system. The linear solve factorizes the matrix and computes the state increment. Each optimization typically converges within approximately $15$ iterations. The latest global spline optimization takes about \SI{2}{\second}, indicating a low overall computational cost.

\section{Conclusion} \label{sec:conclusion}
We presented RESPLE2, a recursive continuous-time $6$-DoF state estimation framework using non-uniform B-splines. By exploiting the knot-clamping boundary condition, the framework directly predicts control points from IMU-integrated kinematics on the trajectory and adapts knot placement to motion and measurement availability. A subsequent control point reset preserves local trajectory consistency as the spline is extended. In addition, a lightweight global optimization module is developed to directly refine the non-uniform B-spline control points by combining loop-closure constraints, reducing accumulated global drift and preserving local kinematic consistency. RESPLE2 supports flexible combinations of LiDAR, visual, inertial, and leg-kinematic measurements, as well as LiDAR-only operation. Extensive real-world evaluations across diverse platforms demonstrate accuracy and robustness comparable to or exceeding those of state-of-the-art methods while maintaining real-time operation, with particularly strong performance under aggressive motion and severe IMU saturation. These results highlight the potential of motion-adaptive, recursive B-spline estimation for robust multi-sensor egomotion estimation under challenging conditions. For future work, we plan to investigate adaptive sensor weighting and visual feature constraints that do not depend on LiDAR-derived depth to further improve estimation robustness under sensor degeneracy. Another promising direction is to integrate spline-based state estimation directly with optimization- or learning-based motion planning onboard mobile robots.

\section*{ACKNOWLEDGMENT}
ChatGPT is used for debugging code.

\appendix

\subsection{Derivation of the Knot-Clamping Boundary Property}\label{subsec:appClamp}
Using the general matrix representation of non-uniform cubic B-splines~\cite{qin1998general}, we derive the six coefficients in the basis matrix $\bom$ \eqref{eq:st} by imposing the clamping condition $t_n=t_{n+1}=t_{n+2}$ at the new knot $t_n$: 
\begin{equation*}
    \begin{aligned}
        m_{0,0}&=\textstyle\frac{(t_{n}-t_{n-1})^2}{(t_{n}-t_{n-2})(t_{n}-t_{n-3})}\,,\,m_{2,0}=\frac{(t_{n-1}-t_{n-2})^2}{(t_{n}-t_{n-2})^2}\,,\\
        m_{2,1}&=\textstyle \frac{3(t_{n}-t_{n-1})(t_{n-1}-t_{n-2})}{(t_{n}-t_{n-2})^2}\,,\,m_{2,2}=\frac{3(t_{n}-t_{n-1})^2}{(t_{n}-t_{n-2})^2}\,,\\
        m_{2,3}&=\textstyle -\frac{1}{3}m_{2,2}-m_{3,3}-\frac{t_{n}-t_{n-1}}{t_{n}-t_{n-2}}\,,\,m_{3,3}=1\,.
    \end{aligned}
\end{equation*}
 At the boundary $t=t_n$, the normalized time is $u=1$, giving $\uu=[\,1,1,1,1\,]^\top$, $\dot\uu=[\,0,1,2,3\,]^\top/\tau_{n-1}$, and $\ddot\uu=[\,0,0,2,6\,]^\top/\tau_{n-1}^2$. Substituting these vectors into the kinematic interpolation formula in \eqref{eq:stKino} yields the following matrix-vector products: 
\begin{equation*}
    \begin{aligned}
    \bom\uu&=
    \bigg[\sbbmat
    &0 \\
    &1-m_{2,0} -m_{2,1} -m_{2,2} -m_{2,3}-m_{3,3} \\
    &m_{2,0} + m_{2,1} + m_{2,2} + m_{2,3} \\
    &m_{3,3}
    \sebmat\bigg]=[\,0,0,0,1\,]^\top\,,\\
    \bom\dot\uu&=\frac{1}{\tau_{n-1}}\bigg[\sbbmat
    0 \\
    -m_{2,1} -2m_{2,2} -3m_{2,3}-3m_{3,3} \\
    m_{2,1} + 2m_{2,2} + 3m_{2,3} \\
        3m_{3,3}
    \sebmat\bigg]=\frac{[\,0,0,-3,3\,]^\top}{\tau_{n-1}}\,,\\
    \bom\ddot\uu&=\frac{1}{\tau_{n-1}^2}\bigg[\sbbmat
    0 \\
    -2m_{2,2} -6m_{2,3}-6m_{3,3} \\
    2m_{2,2} + 6m_{2,3} \\
        6m_{3,3}
    \sebmat\bigg]=\frac{[\,0,6\mu,-6-6\mu,6\,]^\top}{\tau_{n-1}^2}\,,
\end{aligned}
\end{equation*}
with $\mu=\textstyle\frac{t_{n}-t_{n-1}}{t_{n}-t_{n-2}}$. Since $\fS=[\us_{n-3}\,,\us_{n-2}\,,\us_{n-1}\,,\us_n]$, this then establishes \eqref{eq:sn}. For the orientation B-spline, evaluating the cumulative basis functions in \eqref{eq:lam} and their time derivatives at the clamped boundary $t=t_n$ gives     
\begin{equation*}
    \begin{aligned}
    \lambda_{n-3:n}
    &=\bigg[\sbbmat
    1 \\ 1 \\
    m_{2,0} + m_{2,1} + m_{2,2} + m_{2,3} + m_{3,3}\\ m_{3,3}
    \sebmat\bigg]=[\,1,1,1,1\,]^\top\,,\,\text{and}\\
    \dot\lambda_{n-3:n}
    &=\frac{1}{\tau_{n-1}}\bigg[\sbbmat
    0 \\ 0 \\
    m_{2,1} + 2m_{2,2} + 3m_{2,3} +3m_{3,3}\\ 3m_{3,3}
    \sebmat\bigg]=\frac{[\,0,0,0,3\,]^\top}{\tau_{n-1}}\,,
    \end{aligned}
\end{equation*}
with $\lambda_{n-3:n}=[\lambda_{n-3},\lambda_{n-2},\lambda_{n-1},\lambda_n]^\top$.
Substituting these basis function values into \eqref{eq:rt} yields the orientation at the boundary $t=t_n$:
\begin{equation*}
    \textstyle\spr(t_n)=\ur_{n-4}\bullet\prod_{i=n-3}^n\big(\ur_{i-1}^{-1}\bullet\ur_i\big)=\ur_n\,.
\end{equation*}
For angular-velocity interpolation at the clamped boundary $t=t_n=t_{n+1}=t_{n+2}$, we have $\dot\lambda_{n-3}=\dot\lambda_{n-2}=\dot\lambda_{n-1}=0$. The recursion in \eqref{eq:omrec} thus gives $\uom_1(t_n)=\uom_2(t_n)=\uzero_{3\times1}$, yielding
\begin{equation*}
    \uom(t_n)=\ue_n^{-1}\bullet\uzero_{3\times1}\bullet\ue_n+2\dot\lambda_n\,\ude_n=6\,\ude_n/\tau_{n-1}\,,
\end{equation*}
which establishes \eqref{eq:rn}.

\subsection{Derivation of the RCP Reset}\label{subsec:appReset}

\subsubsection{Position RCP Reset} \label{subsubsec:posReset}
In \eqref{eq:reset}, the pre-reset values are evaluated using \eqref{eq:sn} at the right endpoint of $[\,t_{n-2},t_{n-1}]$ with $u^-=1$. After unclamping, the same instant $t_{n-1}$ corresponds to the left endpoint of $[\,t_{n-1},t_n]$ with $u^+=0$. Hence,
\begin{equation*}
    \rg{\sps}(t_{n-1})^+=\big[\us_{n-3}^+,\us_{n-2}^+,\us_{n-1}^+,\us_n^+\big]\,\bom^+\rg{\uu}(t_{n-1})^+\,,
\end{equation*}
where $\us_{n-3:n}^+$ are the post-reset position control points and $\bom^+$ is the corresponding basis matrix. The post-reset normalized time vector and its derivatives at $t_{n-1}$ follow
\begin{equation}\label{eq:uup}
    \begin{aligned}
    &\uu(t_{n-1})^+=[\,1,0,0,0\,]^\top\,,\\
    &\dot\uu(t_{n-1})^+=[\,0,1,0,0\,]^\top/\tau_{n-1}\,,\\
    &\ddot\uu(t_{n-1})^+=[\,0,0,2,0\,]^\top/\tau_{n-1}^2\,.
    \end{aligned}
\end{equation}
The contributions of $\us_{n-4}^-$ and $\us_n^+$ to \eqref{eq:reset} vanish at $u^-=1$ and $u^+=0$, respectively. $\us_n$ is thus left unchanged during the reset, and \eqref{eq:reset} reduces to the following linear system:
\begin{equation*}
    \begin{aligned}
        &\fM^-\us_{n-3:n-1}^-=\fM^+\us_{n-3:n-1}^+\,,\eqwith\\
        &\fM^-=\diag(1,3/\tau_{n-2},6/\tau_{n-2}^2)\Big[\sbbmat
        0&0&1\\
        0&-1&1\\
        \mu^-&-1-\mu^-&1\sebmat\Big]\otimes\fI_3\eqand\\
        &\fM^+=\diag(1,1/\tau_{n-1},2/\tau_{n-1}^2)(\bom_{1:3,1:3}^+)^\top\otimes\fI_3\,.  
    \end{aligned}
\end{equation*}
with $\mu^-=(t_{n-1}-t_{n-2})/(t_{n-1}-t_{n-3})$. Here, $\bom_{1:3,1:3}^+$ extracts the $3\times3$ top-left corner of the post-reset basis matrix. Since $\fM^+$ is nonsingular, the post-reset RCPs can be obtained in closed form as
\begin{equation*}
    \us_{n-3:n-1}^+=(\fM^+)^{-1}\fM^-\,\us_{n-3:n-1}^-\,.
\end{equation*}
Since the latest RCP remains unchanged, this further yields
\begin{equation*}
    \us_{n-3:n}^+=\fJ_s\us_{n-3:n}^-\in\R^{12}\,,\text{with}\,\, \fJ_s=\diag((\fM^+)^{-1}\fM^-,\fI_3)\,,
\end{equation*}
which establishes the position RCP reset in \eqref{eq:pReset}.

\subsubsection{Orientation RCP Reset}\label{subsubsec:ortReset}
We now derive the orientation and angular velocity residuals in \eqref{eq:ortResetResidual}. According to \eqref{eq:lam}, the cumulative basis functions and their derivatives at $t_{n-1}$ follow  
\begin{equation*}
    \rg{\ula}(t_{n-1})^+=\bmp^+\rg{\uu}(t_{n-1})^+\,,
\end{equation*}
with $\rg{\uu}(t_{n-1})^+$ given by \eqref{eq:uup}, yielding
\begin{equation*}
    \begin{aligned}
        &\spr(t_{n-1})^+=\textstyle\ur_{n-4}\bullet\Exp(\ude_{n-3}^+)\bullet\prod_{i=n-2}^{n-1}\Exp(\lambda_i^+\ude_i^+)\,,\\      
        &\uom(t_{n-1})^+=\scR(\ue_{n-1}^+)^\top(2\dot{\lambda}_{n-2}^+\,\ude_{n-2}^+)+2\dot{\lambda}_{n-1}^+\,\ude_{n-1}^+\,.
    \end{aligned}
\end{equation*}
Here, the orientation component is given by the vector part of the relative quaternion
\begin{equation*}
    \begin{aligned}
    &\uz_r=[(\spr(t_{n-1})^-)^{-1}\bullet\spr(t_{n-1})^+]_{2:4}=\textstyle[\ur^-\bullet\prod_{i=n-2}^{n-1}\ue_{i}^+]_{2:4}\,,\\
    &\text{with} \,\, \ur^-=(\Exp(\ude_{n-2}^-)\bullet\Exp(\ude_{n-1}^-))^{-1} \,, \ue_{i}^+=\Exp(\lambda_i^+\ude_{i}^+)\,.
    \end{aligned}
\end{equation*}
The angular-velocity component is given by the difference between the post-reset and pre-reset angular velocities
\begin{equation*}
    \begin{aligned}
    &\uz_\omega=\uom(t_{n-1})^+-\uom(t_{n-1})^-\\
    &=\scR(\ue_{n-1}^+)^\top(2\dot\lambda_{n-2}^+\ude_{n-2}^+)+2\dot\lambda_{n-1}^+\ude_{n-1}^+-6\ude_{n-1}^-/\tau_{n-2}\,.\\
    \end{aligned}
\end{equation*}
For minimizing the residual in the sense of nonlinear least squares using Newton's method, we derive its Jacobian with respect to the increments \ $\ude^+=[(\ude_{n-2}^+)^\top,(\ude_{n-1}^+)^\top]^\top$ as
\begin{equation}\label{eq:hDelta}
    \begin{aligned}
    \fH_{\ude}
    &=\bbmat\fL_1\lrmat{\scQ}(\ue_{n-1}^+)\pad{\ue_{n-2}^+}{\ude_{n-2}^+}&
    \fL_1\llmat{\scQ}(\ue_{n-2}^+)\pad{\ue_{n-1}^+}{\ude_{n-1}^+}\\
    2\dot\lambda_{n-2}^+\scR(\ue_{n-1}^+)^\top&
    \fL_2\pad{\ue_{n-1}^+}{\ude_{n-1}^+}+2\dot\lambda_{n-1}^+\fI_3
    \ebmat\,,\lsft{-12mu}
    \end{aligned}
\end{equation}
with $\fL_1=[\uzero_{3\times1},\fI_3]\,\llmat{\scQ}(\ur^-)\in\R^{3\times4}$. The derivative $\fL_2=\pad{\scR(\ue_{n-1}^+)^\top(2\dot\lambda_{n-2}^+\ude_{n-2}^+)}{\ue_{n-1}^+}$ is evaluated using~\cite[Eq.~174]{sola2017quaternion}. This then enables the iterative solution in \eqref{eq:rReset}.

\subsubsection{Covariance Reset}\label{subsubsec:covReset}
The reset transformation for the position RCP is the Jacobian $\fJ_s$ in \eqref{eq:pReset}. The orientation reset is implicitly defined by the zero-residual condition $\uz_{\ude}(\ude^+,\ude^-)=\uzero_{6\times1}$. Linearizing this condition at the converged post-reset solution $\ude_*^+$ leads to
\begin{equation*}
    \mathrm{d}\uz_{\ude}=\fH_{\ude}^*\,\mathrm{d}\ude^++\fG_{\ude}^*\,\mathrm{d}\ude^-=\uzero_{6\times1}\,,
\end{equation*}
where $\fH_{\ude}^*$ \eqref{eq:hDelta} and $\fG_{\ude}^*$ are the Jacobians of the residual $\uz_{\ude}$ with respect to $\ude^+$ and $\ude^-$, respectively, evaluated at $\ude_*^+$ and $\ude^-$. The Jacobian $\fG_{\ude}^*$ w.r.t.\ the pre-reset $\ude^-$ follows
\begin{equation*}
\begin{aligned}
    &\fG_{\ude}^*=\bbmat
    \fL_3\lrmat{\fQ}_{n-1}\pad{\Exp(\ude_{n-2}^-)}{\ude_{n-2}^-}&\fL_3\llmat{\fQ}_{n-2}\pad{\Exp(\ude_{n-1}^-)}{\ude_{n-1}^-}\\
    \fzero_3&-\frac{6}{\tau_{n-2}}\fI_3
    \ebmat\,,\eqwith\\
    &\lrmat{\fQ}_{n-1}=\lrmat{\scQ}(\Exp(\ude_{n-1}^-)) \eqand \llmat{\fQ}_{n-2}=\llmat{\scQ}(\Exp(\ude_{n-2}^-))\,.
\end{aligned}
\end{equation*}
Here, $\fL_3=[\uzero_{3\times1},\fI_3]\,\lrmat{\scQ}(\ue_{n-2}^+\bullet\ue_{n-1}^+)\,\diag(1,-1,-1,-1)$, with all post-reset quantities evaluated at $\ude_*^+$. Therefore, the Jacobian of the reset map for the full set of orientation RCP increments $\ude_{n-3:n}$ is given by
\begin{equation*}
    \begin{aligned}
    &\fJ_r=\pad{\ude_{n-3:n}^+}{\ude_{n-3:n}^-}=\bbmat\fI_3&\fzero_{3\times6}&\fzero_3\\
    \fzero_{6\times3}&-(\fH_{\ude}^*)^{-1}\fG_{\ude}^*&\fzero_{6\times3}\\
    \fzero_3&\fJ_{n_1}&\fJ_{n_2}\ebmat\,,\eqwith\\
    &\fJ_{n_1}=\pad{\ude_n^+}{[(\ude_{n-2}^-)^\top,(\ude_{n-1}^-)^\top]^\top}\eqand\fJ_{n_2}=\pad{\ude_n^+}{\ude_n^-}\,.
    \end{aligned}
\end{equation*}
Here, the identity matrix indicates the unchanged increment $\ude_{n-3}$. $\fJ_{n_1}$ and $\fJ_{n_2}$ are the Jacobians of $\ude_n^+$ \eqref{eq:udenreset} w.r.t.\ $\ude_{n-2:n-1}^-$ and $\ude_n^-$ computed using the chain rule.

\bibliographystyle{IEEEtran.bst}
\bibliography{bibliography.bib}

\end{document}